\documentclass{article}
\pdfoutput=1

\PassOptionsToPackage{numbers, compress}{natbib}

    \usepackage[final]{neurips_2023}

\usepackage[utf8]{inputenc} 
\usepackage[T1]{fontenc}    
\usepackage{hyperref}       
\usepackage{url}            
\usepackage{booktabs}       
\usepackage{amsfonts}       
\usepackage{nicefrac}       
\usepackage{microtype}      
\usepackage{xcolor}         

\usepackage{graphicx}
\usepackage{amsmath}
\usepackage{amsthm}
\usepackage{amssymb}
\usepackage{booktabs}
\usepackage{multirow}
\usepackage{algorithm}
\usepackage{algpseudocode}
\usepackage{amsmath,amssymb,amsfonts}

\usepackage{pgfplots}
\usepackage{xscommand}

\title{Boosting Adversarial Transferability by\\ Achieving Flat Local Maxima}

\author{%
  Zhijin Ge\textsuperscript{1}\thanks{Equal Contribution.}\;, Hongying Liu\textsuperscript{2}\footnotemark[1]\;, Xiaosen Wang\textsuperscript{3}\footnotemark[1]\;, Fanhua Shang\textsuperscript{4}\thanks{Corresponding authors}\;, Yuanyuan Liu\textsuperscript{1}\footnotemark[2]\\
\textsuperscript{1}School of Artificial Intelligence, Xidian University\\
\textsuperscript{2}The Medical College, Tianjin University\\
\textsuperscript{3}Huawei Singular Security Lab\\
\textsuperscript{4}College of Intelligence and Computing, Tianjin University\\
\texttt{zhijinge@stu.xidian.edu.cn, hyliu2009@tju.edu.cn} \\
\texttt{xiaosen@hust.edu.cn, fhshang@tju.edu.cn, yyliu@xidian.edu.cn}
}

\begin{document}

\maketitle

\begin{abstract}
    Transfer-based attack adopts the adversarial examples generated on the surrogate model to attack various models, making it applicable in the physical world and attracting increasing interest. Recently, various adversarial attacks have emerged to boost adversarial transferability from different perspectives. In this work, inspired by the observation that flat local minima are correlated with good generalization, we assume and empirically validate that adversarial examples at a flat local region tend to have good transferability by introducing a penalized gradient norm to the original loss function. Since directly optimizing the gradient regularization norm is computationally expensive and intractable for generating adversarial examples, we propose an approximation optimization method to simplify the gradient update of the objective function. Specifically, we randomly sample an example and adopt a first-order procedure to approximate the curvature of Hessian/vector product, which makes computing more efficient by interpolating two neighboring gradients. Meanwhile, in order to obtain a more stable gradient direction, we randomly sample multiple examples and average the gradients of these examples to reduce the variance due to random sampling during the iterative process. Extensive experimental results on the ImageNet-compatible dataset show that the proposed method can generate adversarial examples at flat local regions, and significantly improve the adversarial transferability on either normally trained models or adversarially trained models than the state-of-the-art attacks. Our codes are available at: https://github.com/Trustworthy-AI-Group/PGN.
\end{abstract}

\section{Introduction}
A great number of works have shown that Deep Neural Networks (DNNs) are vulnerable to adversarial examples \citep{dong2018boosting, goodfellow2014explaining, moosavi2016deepfool, kurakin2018adversarial, wang2019atgan}, which are generated by applying human-imperceptible perturbations on clean input to result in misclassification. Furthermore, adversarial examples have an intriguing property of transferability \citep{dong2018boosting, dong2019evading, lin2019nesterov, xie2019improving, wu2021improving, wei2018transferable}, \textit{i.e.}, the adversarial example generated from the surrogate model can also misclass other models. The existence of transferability makes adversarial attacks practical to real-world applications because hackers do not need to know any information about the target model, which introduces a series of serious security problems in security-sensitive applications such as self-driving \citep{zhang2018camou, evtimov2017robust} and face-recognition \citep{sharif2016accessorize, lin2022real}.

Although several attack methods \citep{carlini2017towards, goodfellow2014explaining, kurakin2018adversarial,liang2020efficient} have exhibited great attack effectiveness in the white-box setting, they have low transferability when attacking black-box models, especially for some advanced defense models \citep{AleksanderMadry2018TowardsDL, tramer2018ensemble}. Previous works \citep{dong2018boosting, lin2019nesterov, xie2019improving} attribute that the reason for adversarial examples shows weak transferability due to dropping into poor local maxima or overfitting the surrogate model, which is not likely to transfer across models. To address this issue, many methods have been proposed from different perspectives. Gradient optimization attacks \citep{dong2018boosting, lin2019nesterov, wang2021enhancing,zhang2023improving} attempt to boost black-box performance by advanced gradient calculation. Input transformation attacks \citep{xie2019improving, dong2019evading, lin2019nesterov, wang2021admix, long2022frequency,ge2023improving,wang2023structure} aim to generate adversarial examples with higher transferability by applying various transformations to the inputs. Especially, the above methods are mainly proposed from the perspective of optimization and generalization, which regard the process of generating adversarial examples on the white-box model as a standard neural network training process, and treat adversarial transferability as equivalent to model generalization \citep{lin2019nesterov}. Although these methods can improve the transferability of adversarial examples, there are still some gaps between white-box attacks and transfer-based black-box attacks.

Inspired by the observation that flat minima often result in better model generalization~\citep{keskar2016large, zhao2022penalizing, neyshabur2017exploring, foret2020sharpness}, we assume and empirically validate that adversarial examples at a flat local region tend to have good transferability. Intuitively, we can achieve the flat local maxima for adversarial examples using a gradient regularization norm but it is computationally extensive to solve such a problem. To address this issue, we theoretically analyze the optimization process and propose a novel attack called Penalizing Gradient Norm (PGN). In particular, PGN approximates the Hessian/vector product by interpolating the first-order gradients of two samples. This approximation allows us to efficiently generate adversarial examples at flat local regions. To eliminate the error introduced by approximation, PGN incorporates the average gradient of several randomly sampled data points to update the adversarial perturbation.
Our main contributions can be summarized as follows:
\begin{itemize}[leftmargin=*,noitemsep,topsep=2pt]
    \item To the best of our knowledge, it is the first work that empirically validates that adversarial examples located in flat regions have good transferability.
    \item We propose a novel attack called Penalizing Gradient Norm (PGN), which can effectively generate adversarial examples at flat local regions with better transferability.
    \item Empirical evaluations show that PGN can significantly improve the attack transferability on both normally trained models and adversarially trained models, which can also be seamlessly combined with various previous attack methods for higher transferability.
\end{itemize}

\section{Related Work}
In this section, we provide a brief overview of the adversarial attack methods and introduce several studies on the flat local minima for model generalization.
\subsection{Adversarial Attacks}
In general, adversarial attacks can be divided into two categories, \ie, white-box attacks and black-box attacks. In the white-box setting, the target model is completely exposed to the attacker. For instance, Goodfellow \etal \citep{goodfellow2014explaining} proposed the Fast Gradient Sign Method (FGSM) to generate adversarial examples with one step of gradient update. Kurakin \etal \citep{kurakin2018adversarial} further extends FGSM to an iterative version with a smaller step size $\alpha$, denoted as I-FGSM. Madry \etal \citep{AleksanderMadry2018TowardsDL} extends I-FGSM with a random start to generate diverse adversarial examples. Existing white-box attacks have achieved superior performance with the knowledge of the target model. On the other hand, black-box attacks are more practical since they only access limited or no information about the target model. There are two types of black-box adversarial attacks \citep{huang2019black}: query-based and transfer-based attacks. Query-based attacks \citep{brendel2017decision, PinYuChen2017ZOOZO, wang2022triangle} often take hundreds or even thousands of quires to generate adversarial examples, making them inefficient in the physical world. In contrast, transfer-based attacks \citep{xie2019improving, wang2021enhancing, XiaosenWang2021BoostingAT, JunyoungByun2022ImprovingTT, wang2023rethinking, wang2023diversifying, wang2023boosting} generate adversarial examples on the surrogate model, which can also attack other models without accessing the target model, leading to great practical applicability and attracting increasing attention.

Unfortunately, adversarial examples crafted by white-box attacks generally exhibit limited transferability. To boost adversarial transferability, various methods are proposed from the perspective of optimization and generalization. MI-FGSM \citep{dong2018boosting} integrates momentum into I-FGSM to stabilize the update direction and escape from poor local maxima at each iteration. NI-FGSM \citep{lin2019nesterov} adopts Nesterov’s accelerated gradient \citep{nesterov1983method} to further enhance the transferability. Wang \etal \citep{wang2021enhancing} tunned the gradient using the gradient variance of previous iteration to find a more stable gradient direction. Wang \etal \citep{XiaosenWang2021BoostingAT} enhanced the momentum by accumulating the gradient of several data points in the direction of the previous gradient for better transferability.

Data augmentation, which has shown high effectiveness in improving model generalization \citep{mikolajczyk2018data, yu2021does, bayer2022survey, chlap2021review}, has been widely studied to boost adversarial transferability. Xie \etal \citep{xie2019improving} adopted diverse input patterns by randomly resizing and padding to generate transferable adversarial examples. Dong \etal \citep{dong2019evading} utilized several translated images to optimize the adversarial perturbations, and further calculated the gradients by convolving the gradient at untranslated images with a kernel matrix for high efficiency. SIM \citep{lin2019nesterov} optimizes the adversarial perturbations over several scaled copies of the input images. Admix \citep{wang2021admix} mixes up a set of images randomly sampled from other categories while maintaining the original label of the input. Spectrum simulation attack (SSA) \citep{long2022frequency} transforms the input image in the frequency domain to craft more transferable adversarial examples.

Besides, some methods improve adversarial transferability from different perspectives. For instance, Liu \etal \citep{liu2016delving} proposed an ensemble attack, which simultaneously attacks multiple surrogate models. Wu \etal \citep{wu2021improving} employed an adversarial transformation network that can capture the most harmful deformations to adversarial noises. Qin \etal \citep{qin2022boosting} injected the reverse adversarial perturbation at each step of the optimization procedure for better transferability.

\subsection{Flat Minima}
After Hochreiter \etal \citep{hochreiter1997flat} pointed out that well-generalized models may have flat minima, the connection between the flatness of minima and model generalization has been studied from both empirical and theoretical perspectives \citep{keskar2016large, neyshabur2017exploring, foret2020sharpness, zhao2022penalizing}. Li \etal \citep{li2018visualizing} observed that skip connections promote flat minima, which helps explain why skip connections are necessary for training extremely deep networks. Similarly, Santurkar \etal \citep{santurkar2018does} found that BatchNorm makes the optimization landscape significantly smooth in the training process. Sharpness-Aware Minimization (SAM) \citep{foret2020sharpness} improves model generalization by simultaneously minimizing the loss value and sharpness, which seeks the parameters in the neighborhoods with uniformly low loss values. Jiang \etal \citep{jiang2019fantastic} studied 40 complexity measures and showed that a sharpness-based measure has the highest correlation with generalization. Zhao \etal \citep{zhao2022penalizing} demonstrated that adding a gradient norm of the loss function can help the optimizer find flat local minima.

\section{Methodology}

\subsection{Preliminaries}
Given an input image $x$ with its corresponding ground-true label $y$, the deep model $f$ with parameter $\theta$ is expected to output the prediction $f(x;\theta)=y$ with high probability. Let $\mathcal{B}_\epsilon(x)=\{x': \|x' -x \|_p \le \epsilon\}$ be an $\epsilon$-ball of an input image $x$, where $\epsilon>0$ is a pre-defined perturbation magnitude, and $\|\cdot\|_p$ denotes the $L_p$-norm (e.g, the $L_1$-norm). The attacker aims to find an example $x^{adv} \in \mathcal{B}_\epsilon(x)$ that misleads the classifier $f(x^{adv};\theta)\neq y$. Let $J(x,y;\theta)$ be the loss function (\eg, cross-entropy loss) of the classifier $f$. Existing white-box attacks such as~\cite{goodfellow2014explaining, kurakin2018adversarial, dong2018boosting} usually generate adversarial examples by solving the following maximization problem:
\begin{equation}
    \label{eq:maxloss}
	\max_{x^{adv}\in \mathcal{B}_\epsilon(x)}J(x^{adv},y;\theta).
\end{equation}
These attack methods mainly generate adversarial examples through gradient iterations. For instance, I-FGSM~\citep{kurakin2018adversarial} iteratively updates the adversarial perturbation as follows:
\begin{equation}
    \label{eq:ifgsms}
    x_{t+1}^{adv} = \Pi_{\mathcal{B}_\epsilon(x)}\Big[{x_t^{adv}+\alpha \cdot \textrm{sign}(\nabla_{x_t^{adv}}J(x_t^{adv},y;\theta)})\Big], \ \;x_0^{adv} = x,
\end{equation}
where $\Pi_{\mathcal{B}_\epsilon(x)}(\cdot)$ projects an input into $\mathcal{B}_\epsilon(x)$, $\alpha=\epsilon/T$, and $T$ is the total number of iterations. For black-box attacks, the gradient is not accessible so that we cannot directly solve Problem \eqref{eq:maxloss} like I-FGSM. To address such a issue, transfer-based attacks generate adversarial examples on an accessible surrogate model, which can be transferred to fool the target models.

\subsection{Flat Local Maxima Tend to Improve Adversarial Transferability}
\label{sec:motivation}
In general, the adversarial transferability is equivalent to model generalization if we analogize the optimization of perturbation with the model training process~\citep{lin2019nesterov,wang2021enhancing}. Existing works~\citep{keskar2016large, zhao2022penalizing, neyshabur2017exploring, foret2020sharpness} have demonstrated that flat local minima tend to generalize better than their sharp counterparts. This inspires us to assume that adversarial examples at a flat local region w.r.t. the loss function tend to have better transferability across various models. A rigorous description is given as follows:

\begin{assumption}
    Given any small radius $\zeta > 0$ for the local region and two adversarial examples $x_1^{adv}$ and $x_2^{adv}$ for the same input image $x$, if $\max_{x' \in \mathcal{B}_{\zeta}(x_1^{adv})} \| \nabla_{x'}J(x', y;\theta) \|_2 < \max_{x' \in \mathcal{B}_{\zeta}(x_2^{adv})} \| \nabla_{x'}J(x', y;\theta) \|_2$, with high probability, $x_1^{adv}$ tends to be more transferable than $x_2^{adv}$ across various models.
    \label{assum:flat_local_region}
\end{assumption}

SAM \citep{foret2020sharpness} has indicated that a flat local minimum is an entire neighborhood having both low loss and low curvature. Zhao \etal \citep{zhao2022penalizing} also demonstrated that if the loss function has a smaller gradient value, this would indicate that the loss function landscape is flatter. Here we adopt the maximum gradient in the neighborhood to evaluate the flatness of the local region. Following Assumption~\ref{assum:flat_local_region}, we introduce a regularizer, which minimizes the maximum gradient in the $\zeta$-ball of the input, into Eq.~\eqref{eq:maxloss} as follows:
\begin{equation}
	\label{eq:reg_loss}
    \max_{x^{adv}\in \mathcal{B}_\epsilon(x)} \left[J(x^{adv},y;\theta)-\lambda \cdot \max_{x' \in \mathcal{B}_{\zeta}(x^{adv})} \| \nabla_{x'}J(x', y;\theta) \|_2\right],
\end{equation}
where $\lambda\geq0$ is the penalty coefficient. During the optimization process, we penalize the maximum gradient to perceive the sharper regions that can not be identified by using the averaged gradients. By optimizing Problem ~\eqref{eq:reg_loss}, we can find that the adversarial example with a small gradient of data points in its neighborhood should be at a flat local region. However, it is impractical to calculate the maximum gradient in the $\zeta$-ball. Hence, we approximately optimize Eq.~\eqref{eq:reg_loss} by randomly sampling an example $x' \in \mathcal{B}_{\zeta}(x^{adv})$ at each iteration using I-FGSM~\citep{kurakin2018adversarial} and MI-FGSM~\citep{dong2018boosting}, respectively. 

\begin{wrapfigure}{R}{0.5\textwidth}
    \vspace{-1em}
    \begin{tikzpicture}
    \begin{axis}[
    	xlabel=Targeted Models,
    	ylabel=Attack success rates (\%),
    	grid=both,
    	minor grid style={gray!25},
    	major grid style={gray!25},
        xtick={1,2,3,4,5,6,7},
        xticklabels={Inc-v4, IncRes-v2, Res-101, Res-152, Inc-v3$_{ens3}$, Inc-v3$_{ens4}$, IncRes-v2$_{ens}$},
        xticklabel style={rotate=20},
    	width=1.1\linewidth,
        height=0.68\linewidth,
        ylabel style={font=\small, yshift=-5pt},
        xlabel style={font=\small, yshift=8pt},
        tick label style={font=\tiny},
        legend style={font=\tiny, fill opacity=0.8}
    ]
        \addplot[line width=1pt,dashed,mark=o,mark options=solid,color=purple1] %
        	table[x=Model,y=I-FGSM-wor,col sep=comma]{images/data/results_with_regularizer.csv};
        \addlegendentry{I-FGSM w/o Reg.};

        \addplot[line width=1pt,solid,mark=*,color=purple1] %
        	table[x=Model,y=I-FGSM-wr,col sep=comma]{images/data/results_with_regularizer.csv};
        \addlegendentry{I-FGSM w/ Reg.};

        \addplot[line width=1pt,dashed,mark=square,mark options=solid,color=red] %
        	table[x=Model,y=MI-FGSM-wor,col sep=comma]{images/data/results_with_regularizer.csv};
        \addlegendentry{MI-FGSM w/o Reg.};

        \addplot[line width=1pt,solid,mark=square*,color=red] %
        	table[x=Model,y=MI-FGSM-wr,col sep=comma]{images/data/results_with_regularizer.csv};
        \addlegendentry{MI-FGSM w/ Reg.};
    \end{axis}
    \end{tikzpicture}
    \vspace{-1.5em}
    \caption{The average attack success rates (\%) of I-FGSM and MI-FGSM w/wo the gradient regularization on seven black-box models. The adversarial examples are generated on Inc-v3.}
    \label{fig:motivation}
    \vspace{-0.5em}
\end{wrapfigure}
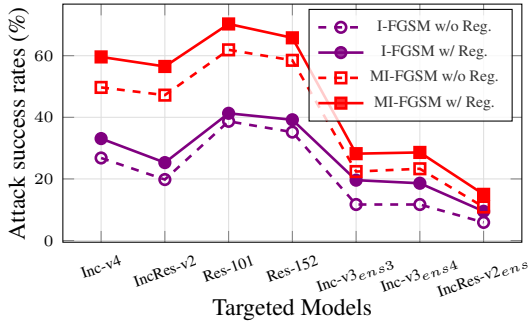

As shown in Fig.~\ref{fig:motivation}, the regularizer of gradients can significantly boost the adversarial transferability of either I-FGSM or MI-FGSM. In general, the average attack success rate is improved by a clear margin of $5.3\%$ and $7.2\%$ for I-FGSM and MI-FGSM, respectively. Such remarkable performance improvement strongly validates Assumption~\ref{assum:flat_local_region}. RAP~\citep{qin2022boosting} also advocates that the adversarial example should be located at a flat local region. However, to the best of our knowledge, it is the first work that empirically validates that flat local maxima can result in good adversarial transferability. Based on this finding, we aim to improve the adversarial transferability by locating the adversarial examples in a flat local region, and a detailed approach for finding flat local maxima will be provided in Sec.~\ref{sec:grm}.

\subsection{Finding Flat Local Maxima}
\label{sec:grm}
Although we have verified that adversarial examples at a flat local region tend to have better transferability, the optimization process involved in Eq.~\eqref{eq:reg_loss} is computationally expensive and intractable for generating transferable adversarial examples. The detailed ablation study is summarized in Sec.~\ref{sec:ablation}. To address this challenge, we propose an approximate optimization approach that efficiently optimizes  Eq.~\eqref{eq:reg_loss} to generate more transferable adversarial examples.

Since $x'$ in Eq.~\eqref{eq:reg_loss} is close to $x^{adv}$ with a small $\zeta$, we assume that the effects of maximizing $J(x^{adv}, y; \theta)$ and maximizing $J(x', y; \theta)$ are expected to be equivalent. Then we can approximately simplify Eq.~\eqref{eq:reg_loss} as follows:
\begin{equation}
	\label{eq:appro_reg_loss}
    \max_{x^{adv}\in \mathcal{B}_\epsilon(x)} \mathcal{L}(x^{adv},y;\theta) \approx J(x',y;\theta)-\lambda \cdot \| \nabla_{x'}J(x', y;\theta) \|_2, \quad \mathrm{s.t.} \quad x'\in \mathcal{B}_\zeta(x^{adv}),
\end{equation}
where $\mathcal{L}$ is a loss function. Here a penalized gradient norm is introduced into the original loss function $J$ for achieving flat local maxima.

In practice, it is computationally expensive to directly optimize Eq.~\eqref{eq:appro_reg_loss}, since we need to calculate its Hessian matrix. Note that existing adversarial attacks typically rely on the sign of the gradient, rather than requiring an exact gradient value. Thus, we approximate the second-order Hessian matrix by using the finite difference method to accelerate the attack process.

\begin{theorem}[Finite Difference Method~\citep{andrei2009accelerated}]
	\label{def:finite_difference}
	Given a finite difference step size $\alpha$ and one normalized gradient direction vector $v=-\frac{\nabla_{x} J(x,y;\theta)}{\left \| \nabla_{x}J(x,y;\theta) \right \|_2 }$, the Hessian/vector product can be approximated by the first-order gradient as follows:
	\begin{equation}
		\label{eq:gradientapprox}
		\nabla_{x}^2 J(x,y;\theta) v  \approx  \frac{\nabla _{x}J(x,y;\theta)|_{x=x+\alpha \cdot v}-\nabla_{x}J(x,y;\theta)}{\alpha}.
	\end{equation}
\end{theorem}

With the finite difference method, we can solve Eq.~\eqref{eq:appro_reg_loss} approximately by using the Hessian/vector product for high efficiency. In particular, we can calculate the gradient at each iteration as follows.

\begin{corollary}
	\label{corollary:appro}
	The gradient of the objective function \eqref{eq:appro_reg_loss} at the $t$-th iteration can be approximated as:
	\begin{equation}
		\label{eq:precorre}
		\nabla _{x_t^{adv}}\mathcal{L}(x_t^{adv},y;\theta) \approx (1-\delta)\cdot \nabla_{x_t'}J(x_t',y;\theta) + \delta \cdot \nabla_{x_t'}J({x_t'},y;\theta)|_{x_t'=x_t'+\alpha \cdot v},
	\end{equation}
	where $x_t'$ is a point randomly sampled in $\mathcal{B}_\zeta(x_t^{adv})$, $\alpha$ is the iteration step size, and $\delta=\frac{\lambda}{\alpha}$ is a balanced coefficient.
\end{corollary}

The detailed proof of Corollary~\ref{corollary:appro} is provided in the Appendix. From Corollary~\ref{corollary:appro}, we can approximate the gradient of Eq.~\eqref{eq:appro_reg_loss} by interpolating two neighboring gradients. This approximation technique significantly enhances the efficiency of the attack process.

In Eq.~\eqref{eq:appro_reg_loss}, we approximate the loss at the input image $x^{adv}$ and the maximum gradient in $\mathcal{B}_{\zeta}(x^{adv})$ by randomly sampling an example $x'\in \mathcal{B}_{\zeta}(x^{adv})$. However, this approach introduces variance due to the random sampling process. To address this issue, we randomly sample multiple examples and average the gradients of these examples to obtain a more stable gradient. Such averaging helps reduce the error introduced by the approximation and randomness, resulting in a more accurate gradient estimation. This, in turn, allows us to effectively explore flat local maxima and achieve better transferability.

\begin{algorithm}[tb]
\caption{Penalizing Gradient Norm (PGN) attack method}
\label{alg:PGN}
\textbf{Input}: A clean image $x$ with ground-truth label $y$, and the loss function $J$ with parameters $\theta$.

\textbf{Parameters}: The magnitude of perturbation $\epsilon$; the maximum number of iterations, $T$; the decay factor $\mu$; the balanced coefficient $\delta$; the upper bound (i.e., $\zeta$) of random sampling in $\zeta$-ball; the number of randomly sampled examples, $N$.
\begin{algorithmic}[1]
    \State $g_0 = 0$, \;$x_0^{adv}=x$, \; $\alpha = \epsilon/T$;
    \For {$t = 0, 1, \cdots, T-1$}
        \State Set $\bar{g}$ = 0;
            \For {$i = 0, 1, \cdots, N-1$}
                \State Randomly sample an example $x' \in \mathcal{B}_{\zeta}(x_t^{adv})$;
                \State Calculate the gradient at the sample $x'$, $g' = \nabla_{x'}J(x',y;\theta)$;
                \State Compute the predicted point by $x^{\ast} = x' - \alpha \cdot \frac{g'}{\left \|g' \right \|_1}$;
                \State Calculate the gradient of the predicted point, $g^{\ast} = \nabla_{x^{\ast}}J(x^{\ast},y;\theta)$;
                \State Accumulate the updated gradient by $\bar{g} = \bar{g} + \frac{1}{N} \cdot \left[ (1-\delta) \cdot g' + \delta \cdot g^{\ast} \right]$;
            \EndFor
	  \State $g_{t+1} =\mu \cdot g_t +\frac{\bar{g}}{\left \| \bar{g} \right \|_1 }$;
        \State Update $x_{t+1}^{adv}$ via $x_{t+1}^{adv} = \Pi_{\mathcal{B}_\epsilon(x)}\left[{x_t^{adv}+\alpha \cdot \textrm{sign}(g_{t+1})}\right]$;
        \EndFor
        \State \textbf{return} $x^{adv}=x_T^{adv}$.
    \end{algorithmic}
\textbf{Output}: An adversarial example $x^{adv}$.
\end{algorithm}

In short, we introduce a novel attack method, called Penalizing Gradient Norm (PGN). The PGN attack aims to guide adversarial examples towards flatter regions by constraining the norm or magnitude of the gradient. The details of the PGN attack are outlined in Algorithm \ref{alg:PGN}. Since PGN is derived by optimizing Eq.~\eqref{eq:reg_loss} to achieve a flat local maximum, it can be seamlessly integrated with existing gradient-based attack methods and input transformation-based attack methods, leveraging their strengths to further improve adversarial transferability.

\section{Experiments}
In this section, we conduct extensive experiments on the ImageNet-compatible dataset. We first provide the experimental setup. Then we compare the results of the proposed methods with existing methods on both normally trained models and adversarially trained models. Finally, we conduct ablation studies to study the effectiveness of key parameters in our PGN. The experimental results were performed multiple times and averaged to ensure the experimental results were reliable. 
\subsection{Experimental Settings}
\label{sec:setting}
\textbf{Dataset.} We conduct our experiments on the ImageNet-compatible dataset,  which is widely used in previous works \citep{JunyoungByun2022ImprovingTT, long2022frequency, qin2022boosting}. It contains 1,000 images with the size of $299 \times 299 \times 3$, ground-truth labels, and target labels for targeted attacks.

\textbf{Models.} To validate the effectiveness of our methods, we test attack performance in five popular pre-trained models, including Inception-v3 (Inc-v3) \citep{szegedy2016rethinking}, Inception-v4 (Inc-v4), InceptionResnet-v2 (IncRes-v2) \citep{szegedy2017inception}, ResNet-101 (Res-101), and ResNet-152 (Res-152) \cite{KaimingHe2015DeepRL}. We also consider adversarially trained models including, Inc-v3$_{ens3}$, Inc-v4$_{ens4}$, and IncRes-v2$_{ens}$ \cite{tramer2018ensemble}.

\textbf{Baselines.} We take five popular gradient-based iterative adversarial attacks as our baselines, including, MI-FGSM \citep{dong2018boosting}, NI-FGSM \citep{lin2019nesterov}, VMI-FGSM \citep{wang2021enhancing}, EMI-FGSM \citep{XiaosenWang2021BoostingAT}, RAP \citep{qin2022boosting}. We also integrate the proposed method with various input transformations to validate the generality of our PGN, such as DIM \citep{dong2019evading}, TIM \citep{xie2019improving}, SIM \cite{lin2019nesterov}, Admix \citep{wang2021admix}, and SSA \citep{long2022frequency}.

\textbf{Hyper-parameters.} We set the maximum perturbation of the parameter $\epsilon = 16.0/255$, the number of iterations $T = 10$, and the step size $\alpha = \epsilon/T$. For MI-FGSM and NI-FGSM, we set the decay factor $\mu = 1.0$. For VMI-FGSM, we set the number of sampled examples $N = 20$ and the upper bound of neighborhood size $\beta = 1.5 \times \epsilon$. For EMI-FGSM, we set the number of examples $N = 11$, the sampling interval bound $\eta = 7$, and adopt the linear sampling. For the attack method, RAP, we set the step size $\alpha=2.0/255$, the number of iterations $K = 400$, the inner iteration number $T = 10$, the late-start $K_{LS} = 100$, the size of neighborhoods $\epsilon_n = 16.0/255$. For our proposed PGN, we set the number of examples $N = 20$, the balanced coefficient $\delta = 0.5$, and the upper bound of $\zeta = 3.0 \times \epsilon$.

\subsection{Visualization of Loss Surfaces for Adversarial Example}
\begin{figure*}
    \centering
    \begin{subfigure}{.15\textwidth}
        \centering
        \includegraphics[width=\linewidth]{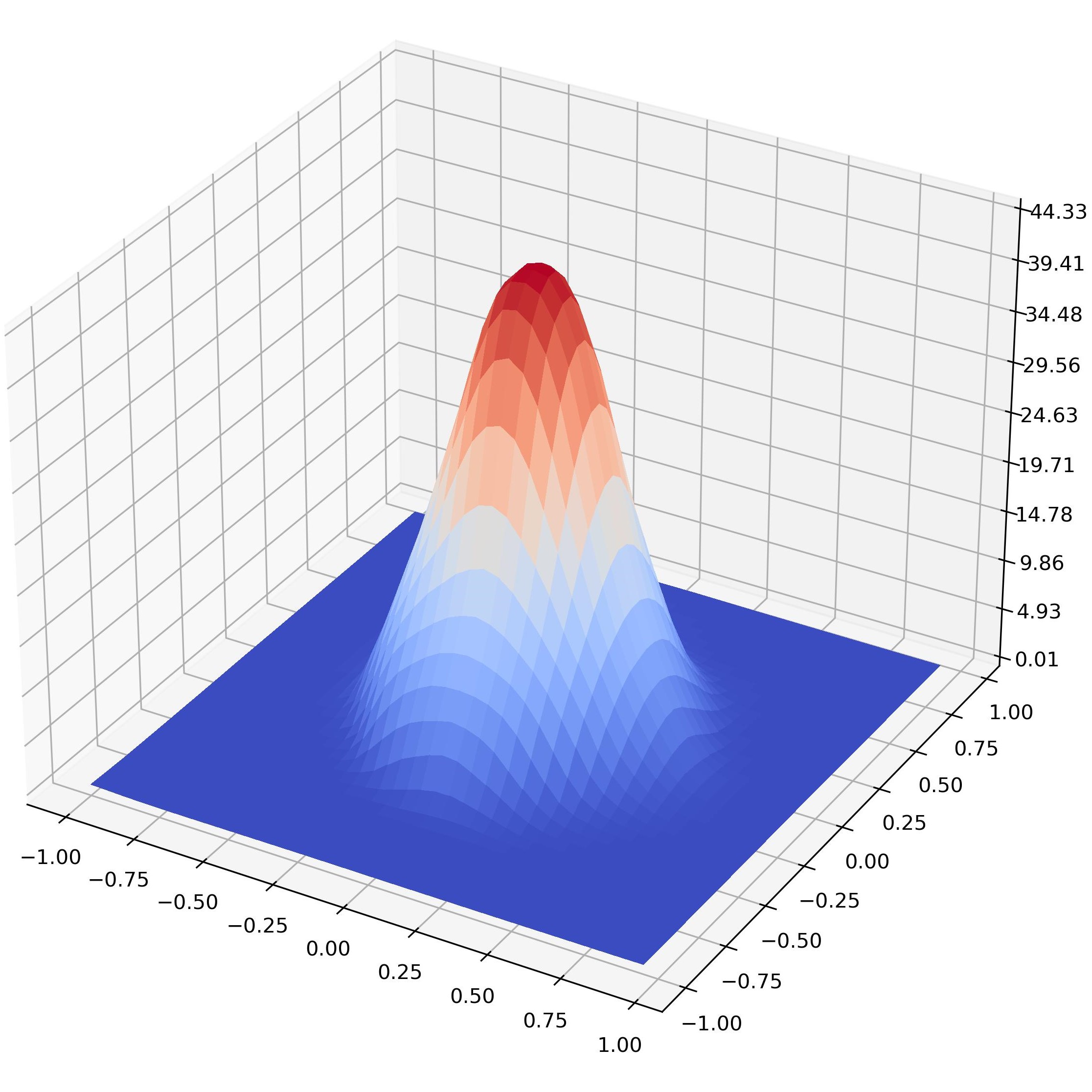}
        \vspace{-0.5cm}
    \end{subfigure}%
    \hspace{1mm}
    \begin{subfigure}{.15\textwidth}
        \centering
        \includegraphics[width=\linewidth]{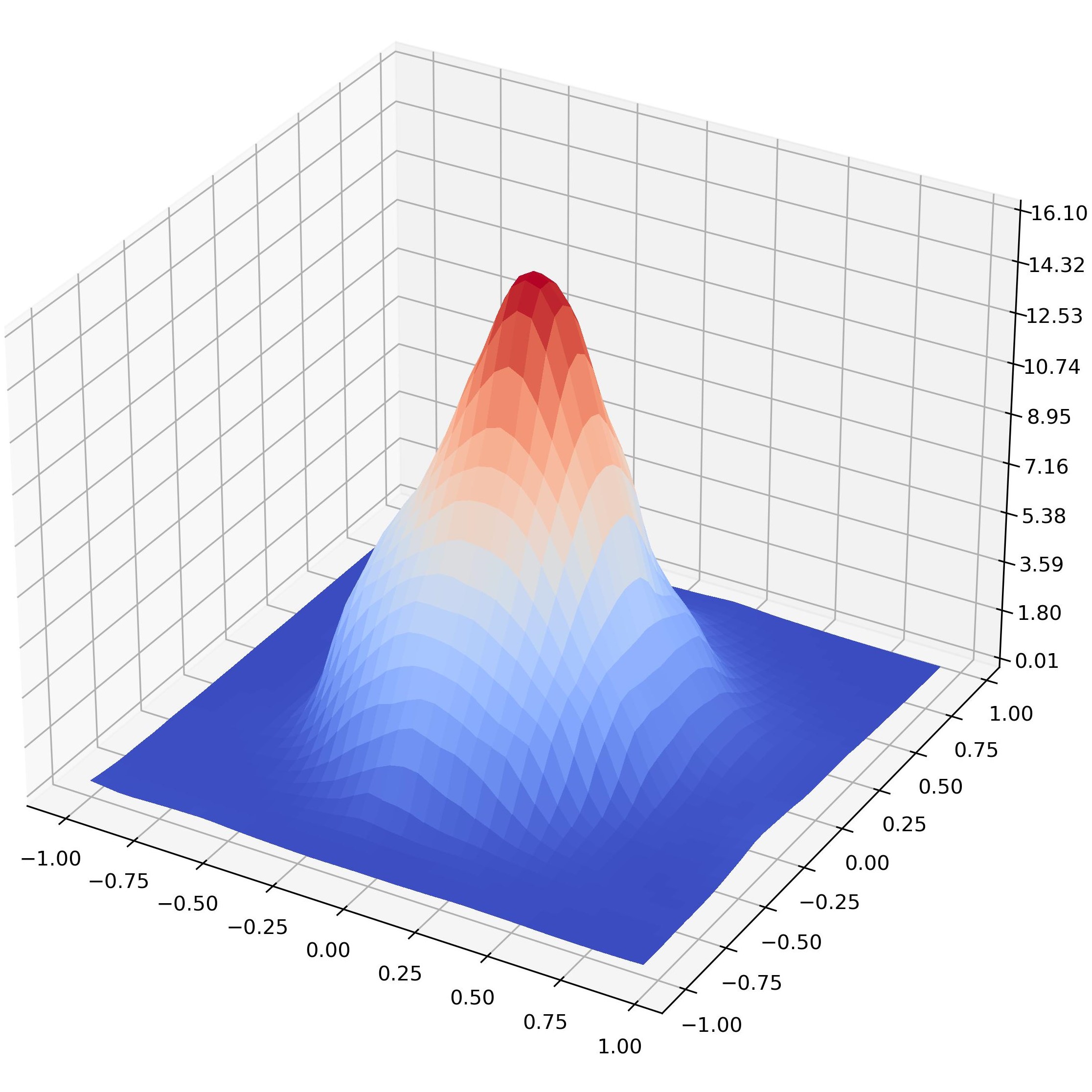}
        \vspace{-0.5cm}
    \end{subfigure}%
    \hspace{1mm}
    \begin{subfigure}{.15\textwidth}
        \centering
        \includegraphics[width=\linewidth]{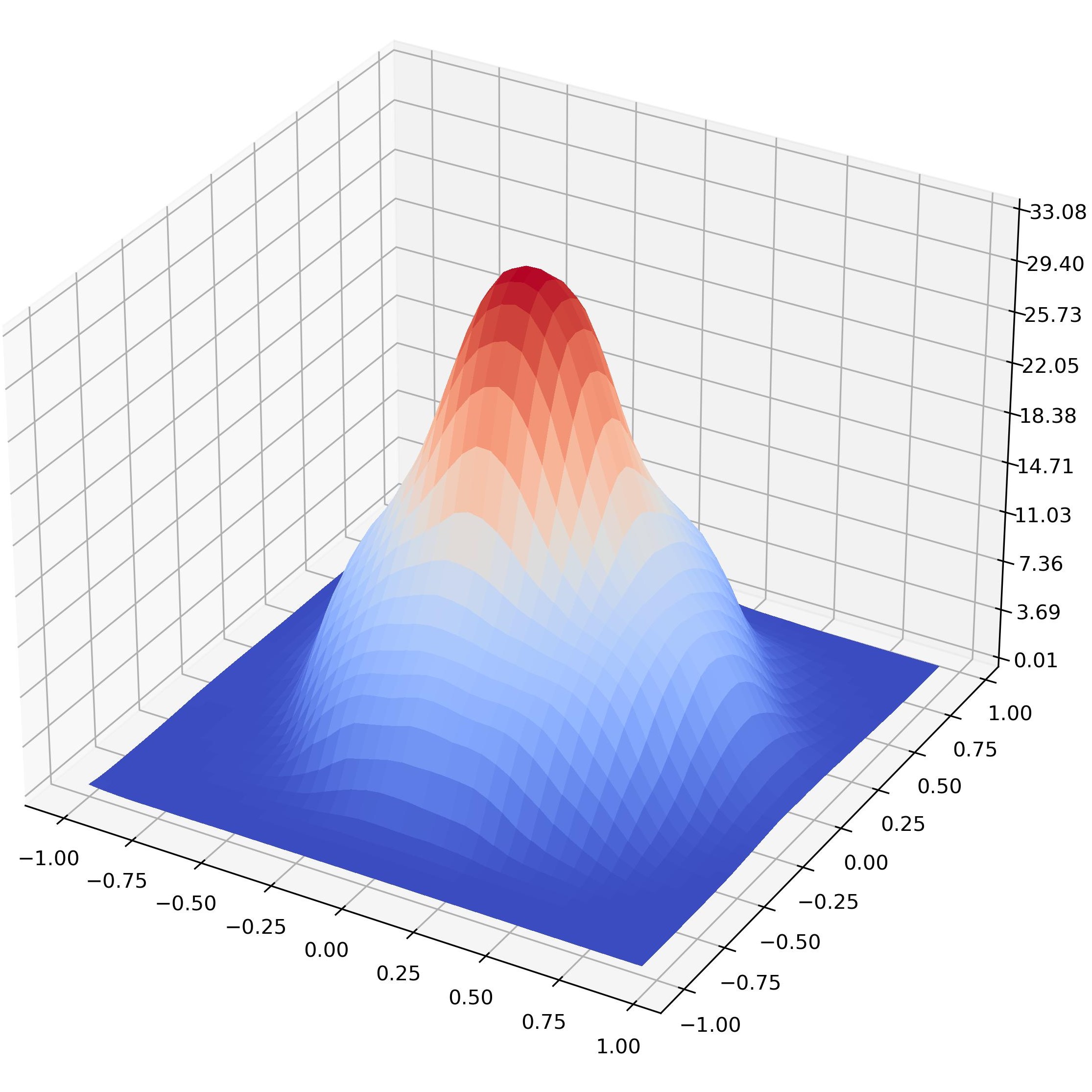}
        \vspace{-0.5cm}
    \end{subfigure}
    \hspace{1mm}
    \centering
    \begin{subfigure}{.15\textwidth}
        \centering
        \includegraphics[width=\linewidth]{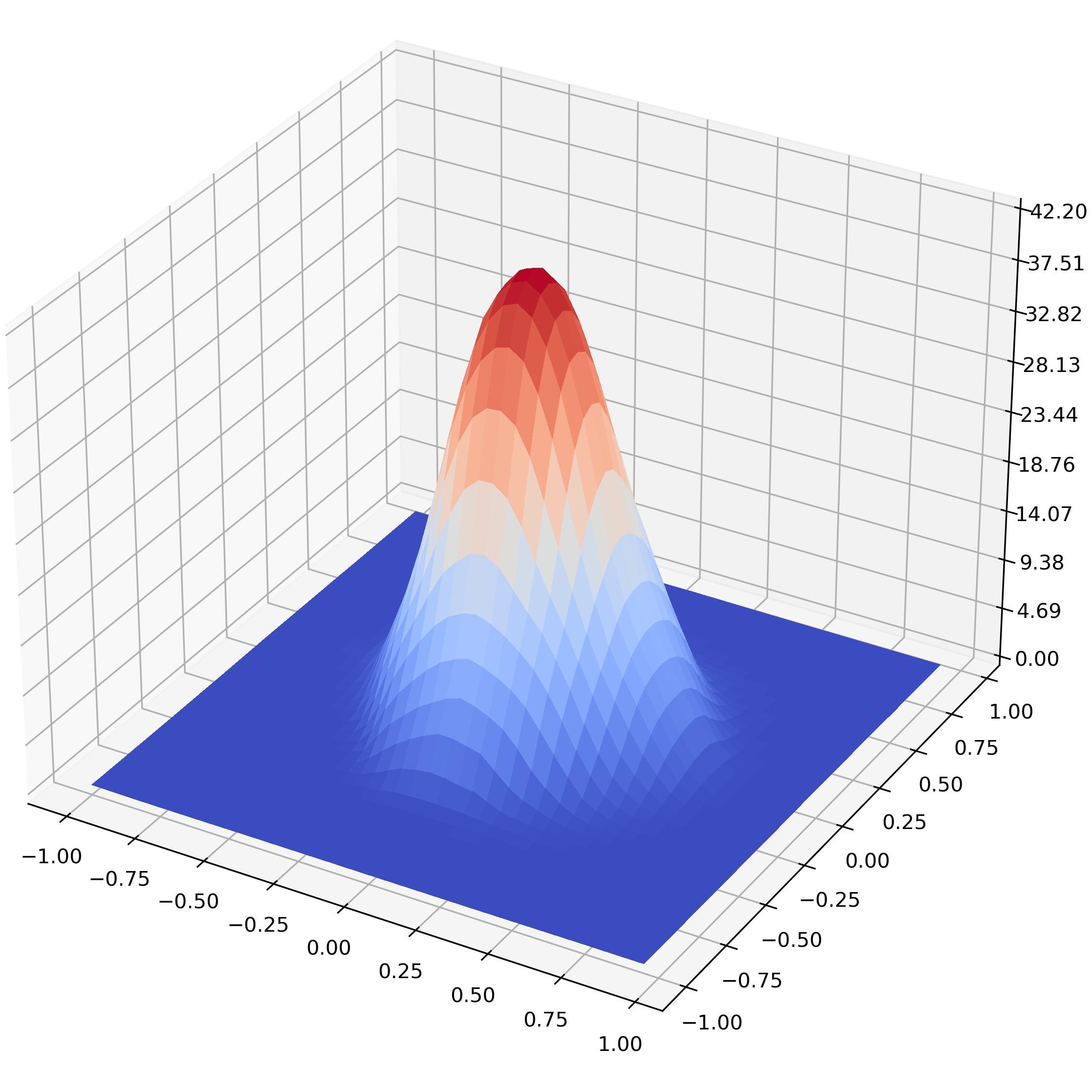}
        \vspace{-0.5cm}
    \end{subfigure}%
    \hspace{1mm}
    \begin{subfigure}{.15\textwidth}
        \centering
        \includegraphics[width=\linewidth]{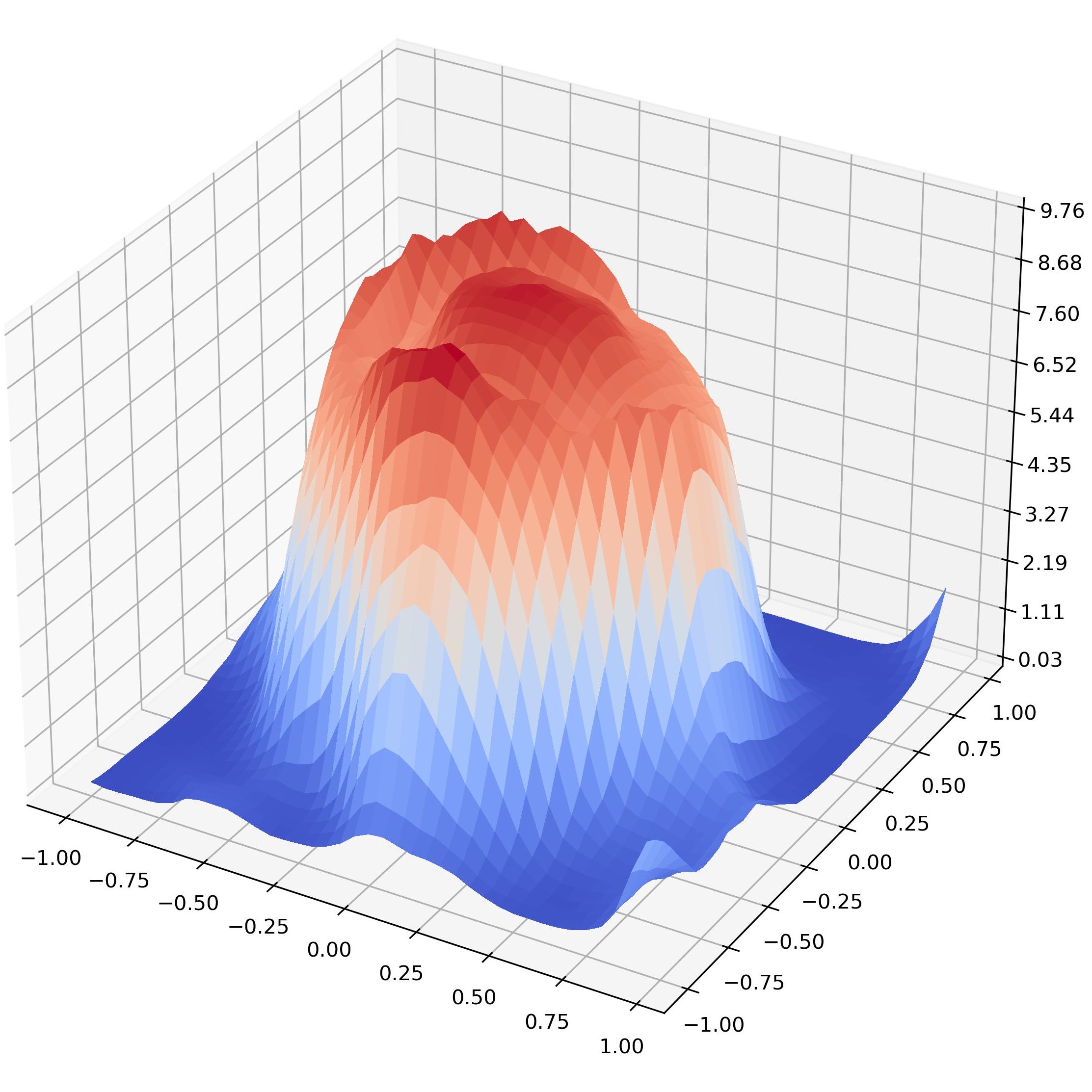}
        \vspace{-0.5cm}
    \end{subfigure}%
    \hspace{1mm}
    \begin{subfigure}{.15\textwidth}
        \centering
        \includegraphics[width=\linewidth]{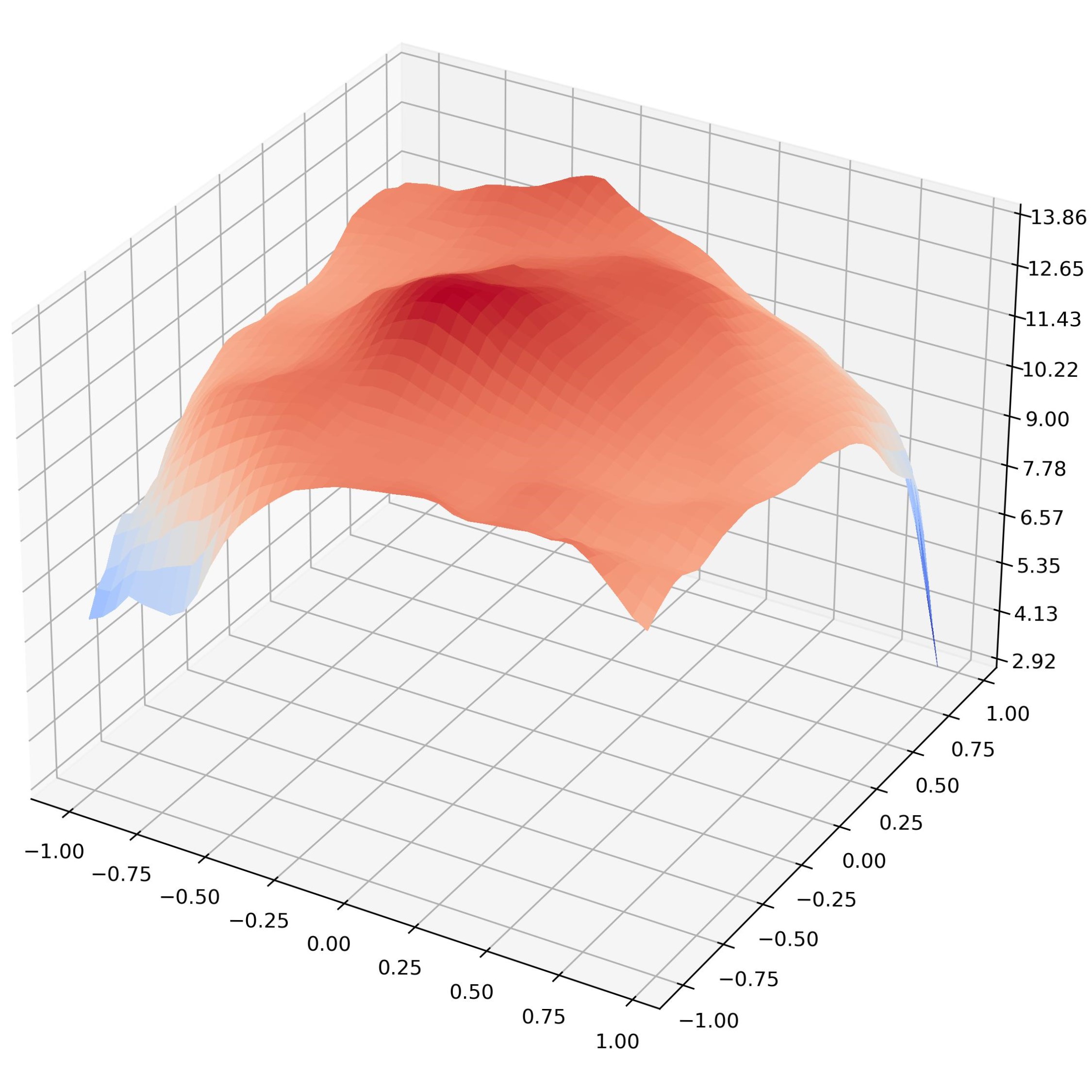}
        \vspace{-0.5cm}
    \end{subfigure}

    \vspace{4mm}

    \begin{subfigure}{.15\textwidth}
        \centering
        \includegraphics[width=\linewidth]{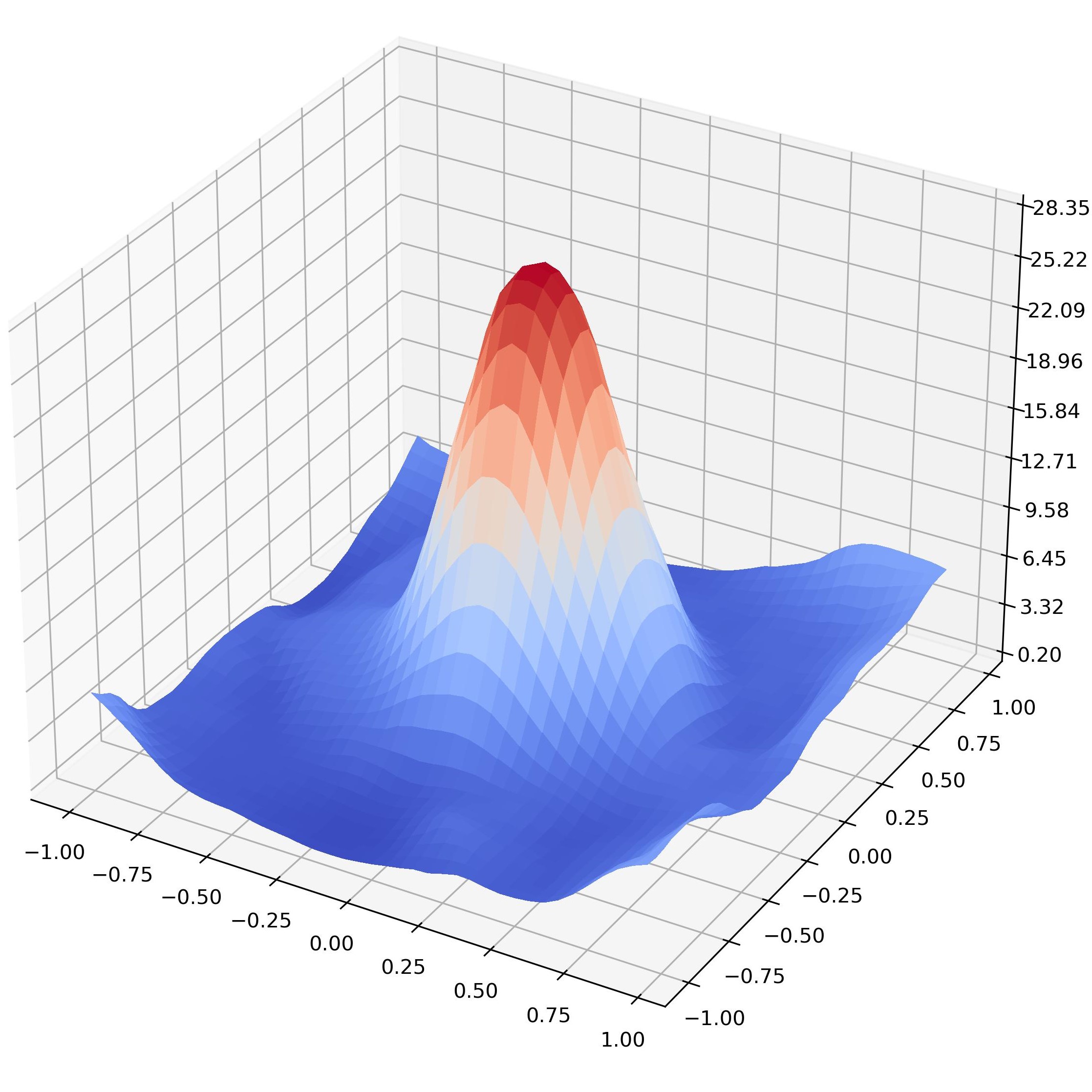}
        \vspace{-0.5cm}
        \caption{MI-FGSM}
    \end{subfigure}%
    \hspace{1mm}
    \begin{subfigure}{.15\textwidth}
        \centering
        \includegraphics[width=\linewidth]{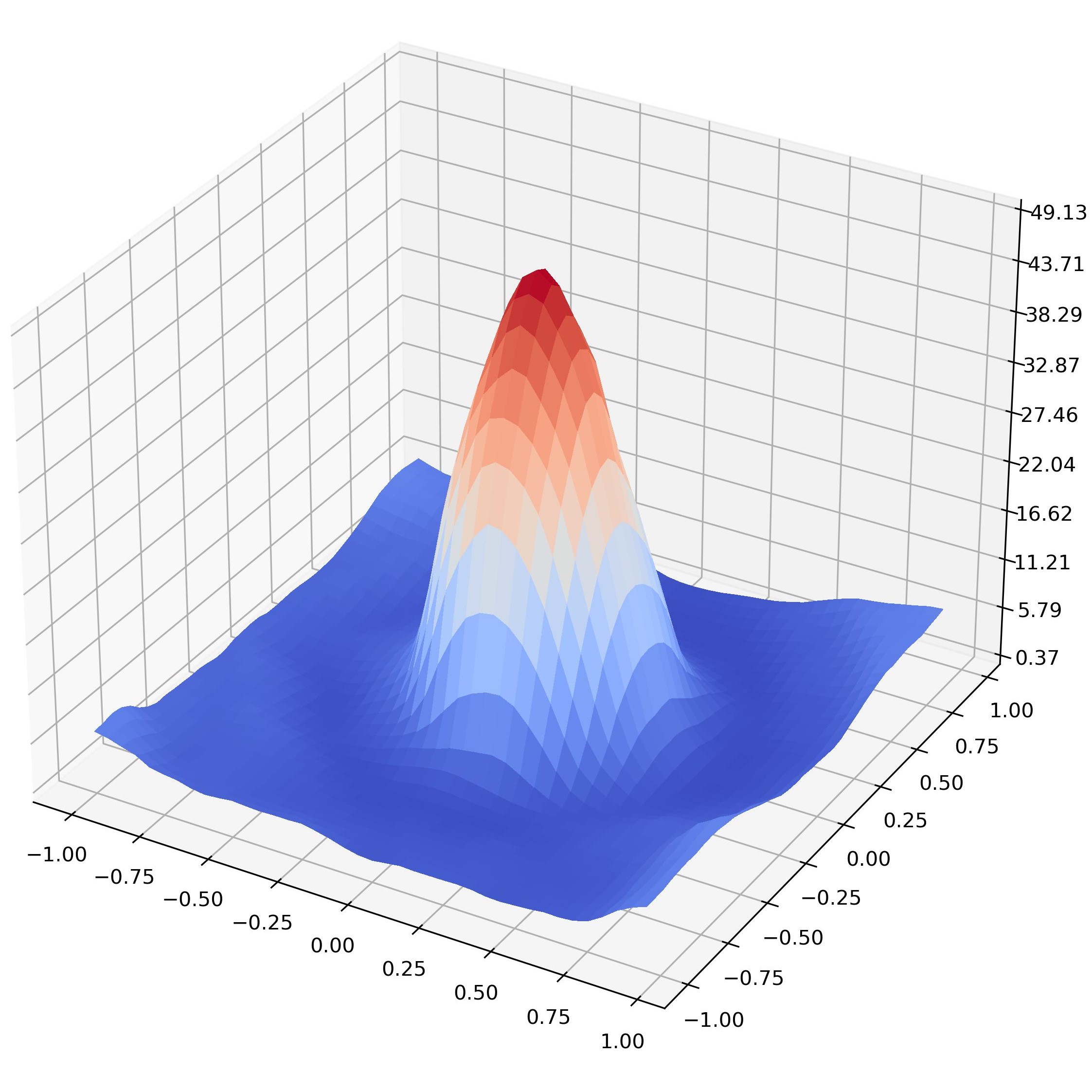}
        \vspace{-0.5cm}
        \caption{NI-FGSM}
    \end{subfigure}%
    \hspace{1mm}
    \begin{subfigure}{.15\textwidth}
        \centering
        \includegraphics[width=\linewidth]{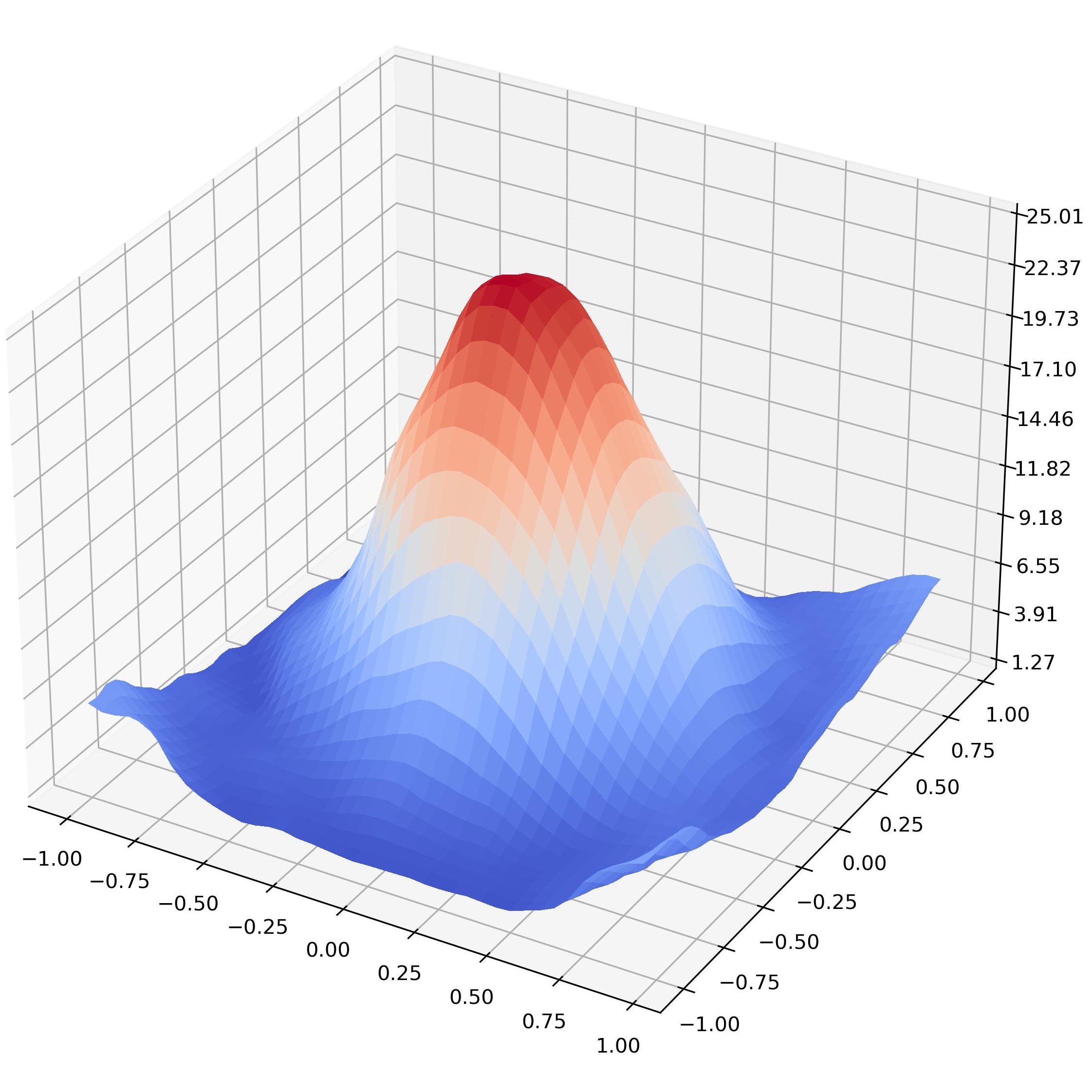}
        \vspace{-0.5cm}
        \caption{VMI-FGSM}
    \end{subfigure}
    \hspace{1mm}
    \centering
    \begin{subfigure}{.15\textwidth}
        \centering
        \includegraphics[width=\linewidth]{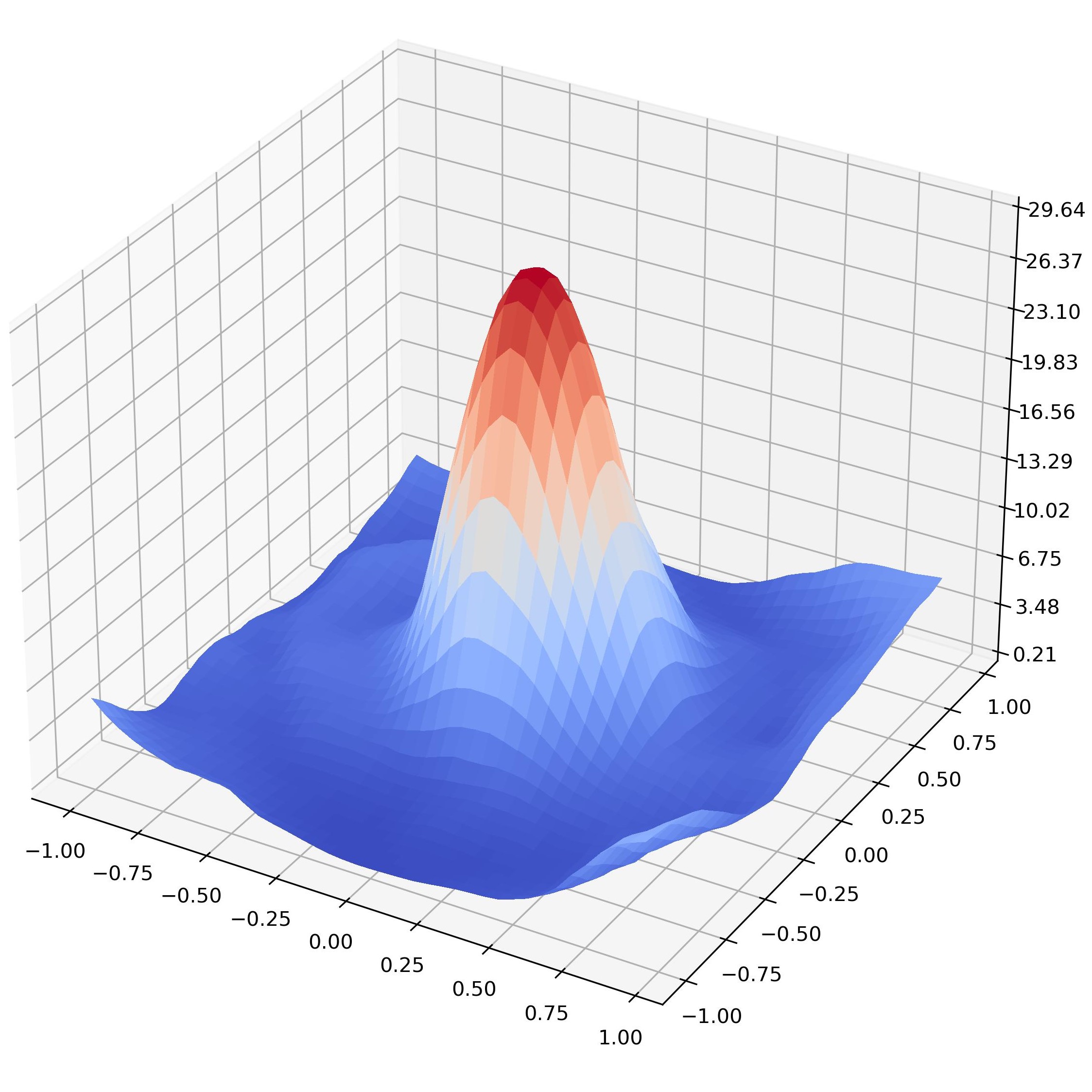}
        \vspace{-0.5cm}
        \caption{EMI-FGSM}
    \end{subfigure}%
    \hspace{1mm}
    \begin{subfigure}{.15\textwidth}
        \centering
        \includegraphics[width=\linewidth]{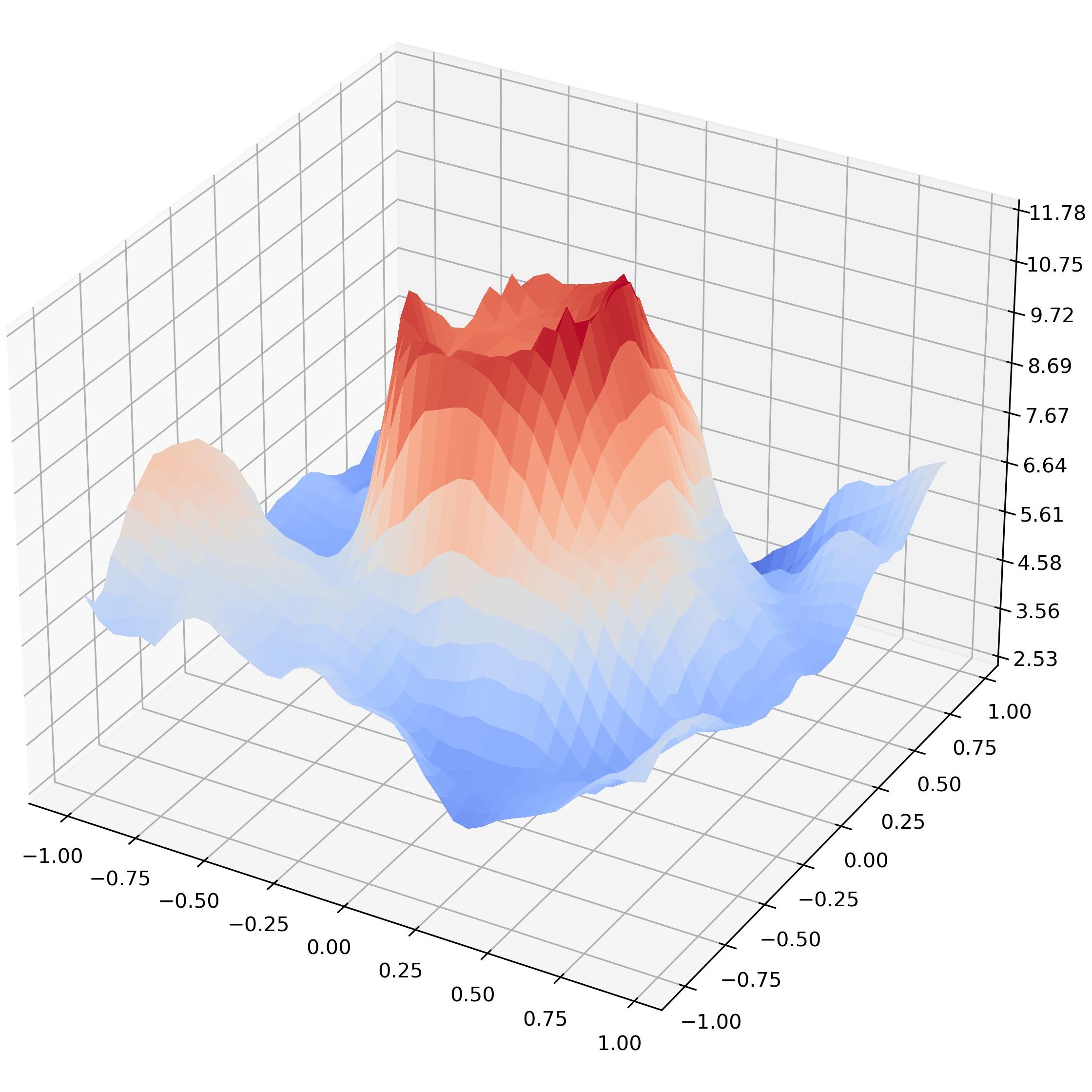}
        \vspace{-0.5cm}
        \caption{RAP}
    \end{subfigure}%
    \hspace{1mm}
    \begin{subfigure}{.15\textwidth}
        \centering
        \includegraphics[width=\linewidth]{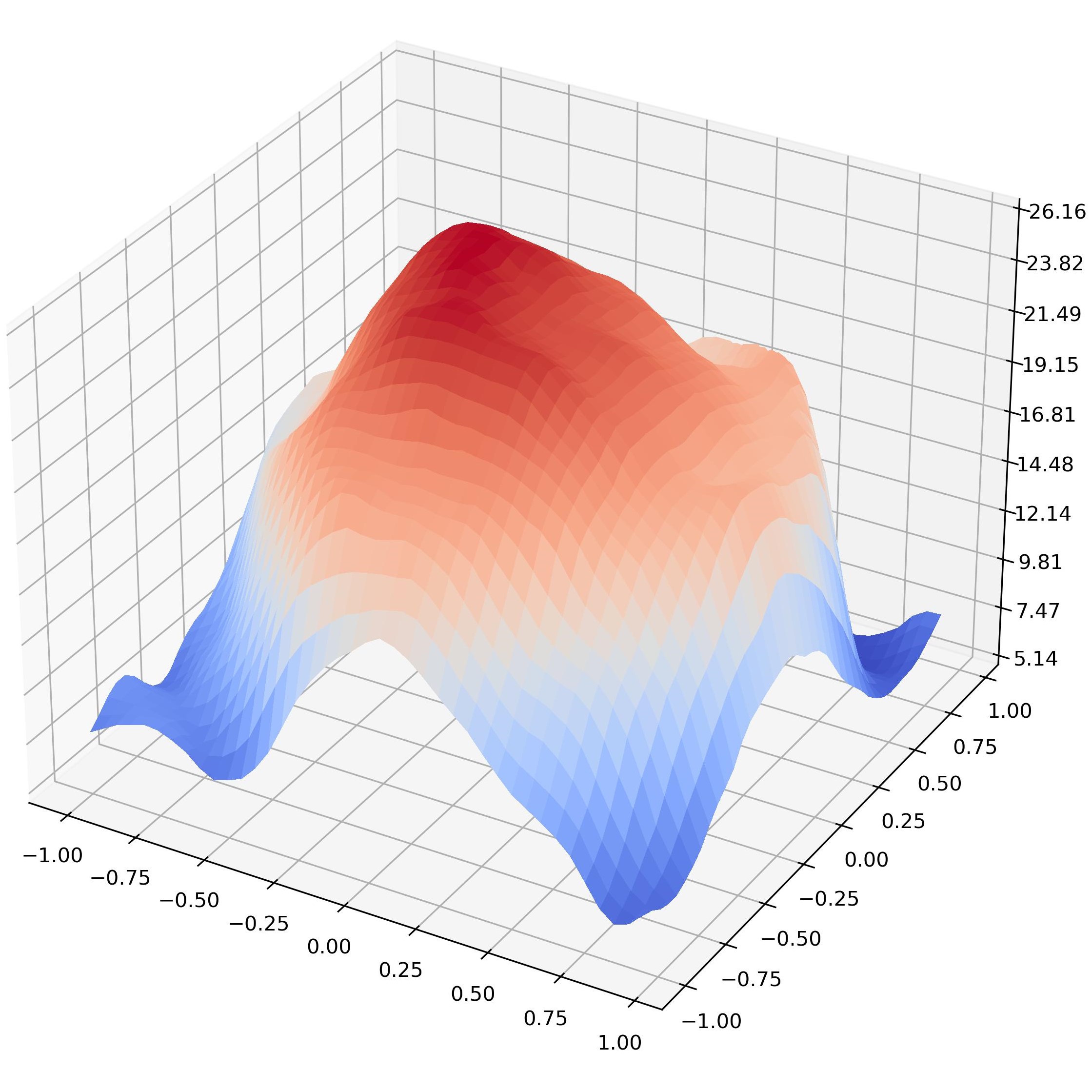}
        \vspace{-0.5cm}
        \caption{PGN}
    \end{subfigure}

    \caption{Visualization of loss surfaces along two random directions for two randomly sampled adversarial examples on the surrogate model (\ie, Inc-v3). The center of each 2D graph corresponds to the adversarial example generated by different attack methods (see more examples in the Appendix).}
    \label{fig:transformation}
\end{figure*}

To validate that our proposed PGN method can help the adversarial examples find a flat maxima region, we compare the loss surface maps of the adversarial examples generated by different attack methods on the surrogate model (\ie, Inc-v3). Each 2D graph corresponds to an adversarial example, with the adversarial example shown at the center. We randomly select two images from the dataset and compare the loss surfaces in Fig.~\ref{fig:transformation}, each row represents the visualization of one image. From the comparison, we observe that our PGN method can help adversarial examples achieve flatter maxima regions compared to the baselines. The adversarial examples generated by our method are located in larger and smoother flat local maxima. This confirms that our method can generate adversarial examples located in the flat maximum. With the adversarial examples located in more flat local maxima, PGN exhibits much better transferability than existing attacks as shown in Sec.~\ref{sec:exp:single}-~\ref{sec:exp:combine}, which further supports our motivation.

\subsection{Attack a Single Model}
\label{sec:exp:single}
\begin{table}[!t]
\small
  \centering
  \caption{The untargeted attack success rates (\%$\pm$std, over 10 random runs) of various gradient-based attacks in the single model setting. The adversarial examples are crafted on Inc-v3, Inc-v4, IncRes-v2, and Res-101 by MI-FGSM (MI), NI-FGSM (NI), VMI-FGSM (VMI), EMI-FGSM (EMI), RAP, and our PGN attack methods, respectively. Here * indicates the white-box model.}
  \vspace{1mm}
  \setlength{\tabcolsep}{2.00pt}
    \begin{tabular}{|c|c|ccccccc|}
    \hline
    Model & Attack & Inc-v3 & Inc-v4 & IncRes-v2 & Res-101 & Inc-v3$_{ens3}$ & Inc-v3$_{ens4}$ & IncRes-v2$_{ens}$  \\
    \hline
    \hline
    \multirow{6}[2]{*}{Inc-v3} & MI & \textbf{100.0$\pm$0.0}*  & 51.0$\pm$0.47 & 45.8$\pm$0.60  & 49.0$\pm$0.24  & 22.6$\pm$0.52  & 22.0$\pm$0.35  & 10.9$\pm$0.24  \\
          & NI & \textbf{100.0$\pm$0.0}*   & 61.4$\pm$0.42  & 59.6$\pm$0.54   & 57.2$\pm$0.18   & 22.5$\pm$0.37  & 22.7$\pm$0.35  & 11.5$\pm$0.26  \\
          & VMI & \textbf{100.0$\pm$0.0}*  & 74.8$\pm$0.58  & 69.9$\pm$0.92  & 65.5$\pm$0.67   & 41.6$\pm$0.54  & 41.6$\pm$0.54  & 25.0$\pm$0.34 \\
          & EMI & \textbf{100.0$\pm$0.0}*   & 80.7$\pm$0.58  & 77.1$\pm$0.37  & 72.4$\pm$0.83  & 33.0$\pm$0.61  & 31.9$\pm$0.49  & 17.0$\pm$0.48 \\
          & RAP   & 99.9$\pm$0.10* & 84.5$\pm$0.69  & 79.3$\pm$0.47 & 76.5$\pm$0.65 & 56.9$\pm$0.84 & 51.3$\pm$0.62 & 31.9$\pm$0.35 \\
          & \cellcolor[rgb]{0.9, 0.9, 0.9}PGN & \cellcolor[rgb]{0.9, 0.9, 0.9}\textbf{100.0$\pm$0.0}* & \cellcolor[rgb]{0.9, 0.9, 0.9}\textbf{90.6$\pm$0.67} & \cellcolor[rgb]{0.9, 0.9, 0.9}\textbf{89.5$\pm$0.75} & \cellcolor[rgb]{0.9, 0.9, 0.9}\textbf{81.2$\pm$0.68} & \cellcolor[rgb]{0.9, 0.9, 0.9}\textbf{64.6$\pm$0.75} & \cellcolor[rgb]{0.9, 0.9, 0.9}\textbf{65.6$\pm$0.94} & \cellcolor[rgb]{0.9, 0.9, 0.9}\textbf{45.3$\pm$0.77} \\
    \hline
    \multirow{6}[2]{*}{Inc-v4} & MI & 57.2$\pm$0.36  & \textbf{100.0$\pm$0.0}*  & 46.1$\pm$0.14  & 51.5$\pm$0.33  & 19.1$\pm$0.46  & 18.4$\pm$0.23  & 10.2$\pm$0.36 \\
          & NI & 62.8$\pm$0.43  & \textbf{100.0$\pm$0.0}*  & 52.7$\pm$0.34    & 56.7$\pm$0.19    & 19.2$\pm$0.25  & 18.3$\pm$0.37  & 11.7$\pm$0.29 \\
          & VMI & 77.6$\pm$0.65  & 99.8$\pm$0.10*  & 69.8$\pm$0.41   & 66.7$\pm$0.33  & 41.1$\pm$0.87  & 41.2$\pm$0.54  & 27.0$\pm$0.24 \\
          & EMI & 84.2$\pm$0.62  & 99.7$\pm$0.10*  & 75.0$\pm$0.70  & 74.4$\pm$0.64  & 31.5$\pm$0.44  & 28.0$\pm$0.65  & 16.2$\pm$0.36 \\
          & RAP   & 85.3$\pm$0.74 & 99.5$\pm$0.21* & 79.5$\pm$0.62 &  77.2$\pm$0.42 & 45.2$\pm$0.69 & 46.8$\pm$0.48  & 29.3$\pm$0.51\\
          & \cellcolor[rgb]{0.9, 0.9, 0.9}PGN & \cellcolor[rgb]{0.9, 0.9, 0.9}\textbf{91.2$\pm$0.58} & \cellcolor[rgb]{0.9, 0.9, 0.9}99.6$\pm$0.15* & \cellcolor[rgb]{0.9, 0.9, 0.9}\textbf{87.6$\pm$0.74} & \cellcolor[rgb]{0.9, 0.9, 0.9}\textbf{83.5$\pm$0.53} & \cellcolor[rgb]{0.9, 0.9, 0.9}\textbf{67.0$\pm$0.68} & \cellcolor[rgb]{0.9, 0.9, 0.9}\textbf{64.2$\pm$0.63} & \cellcolor[rgb]{0.9, 0.9, 0.9}\textbf{49.1$\pm$0.82}  \\
    \hline
    \multirow{6}[2]{*}{IncRes-v2} & MI & 58.2$\pm$0.21  & 52.4$\pm$0.41  & 99.3$\pm$0.21*  & 50.7$\pm$0.26 & 22.0$\pm$0.37  & 22.0$\pm$0.31  & 13.8$\pm$0.43 \\
          & NI & 60.3$\pm$0.35 & 57.1$\pm$0.17  & 99.5$\pm$0.17*  & 55.3$\pm$0.35  & 18.3$\pm$0.18 & 19.3$\pm$0.29  & 12.1$\pm$0.16 \\
          & VMI & 78.2$\pm$0.64  & 77.0$\pm$0.57  & 99.1$\pm$0.36*  & 66.0$\pm$0.48  & 47.6$\pm$0.69  & 43.3$\pm$0.36  & 37.7$\pm$0.37 \\
          & EMI & 85.2$\pm$0.78  & 83.3$\pm$0.29  & 99.7$\pm$0.18*  & 74.0$\pm$0.56  & 38.4$\pm$0.48  & 33.8$\pm$0.53  & 24.1$\pm$0.48 \\
          & RAP & 87.1$\pm$0.75 & 84.2$\pm$0.45 & 99.4$\pm$0.28* & 79.4$\pm$0.64 & 50.3$\pm$0.47  & 49.8$\pm$0.89 & 40.2$\pm$0.54 \\
          & \cellcolor[rgb]{0.9, 0.9, 0.9}PGN & \cellcolor[rgb]{0.9, 0.9, 0.9}\textbf{92.0$\pm$0.69} & \cellcolor[rgb]{0.9, 0.9, 0.9}\textbf{92.3$\pm$0.63} & \cellcolor[rgb]{0.9, 0.9, 0.9}\textbf{99.8$\pm$0.10}* & \cellcolor[rgb]{0.9, 0.9, 0.9}\textbf{83.5$\pm$0.41} & \cellcolor[rgb]{0.9, 0.9, 0.9}\textbf{74.6$\pm$0.75} & \cellcolor[rgb]{0.9, 0.9, 0.9}\textbf{71.5$\pm$0.64} & \cellcolor[rgb]{0.9, 0.9, 0.9}\textbf{66.62$\pm$0.58} \\
    \hline
    \multirow{6}[2]{*}{Res-101} & MI & 51.5$\pm$0.26  & 42.2$\pm$0.35  & 36.3$\pm$0.24  & \textbf{100.0$\pm$0.0}*  & 18.7$\pm$0.32  & 16.6$\pm$0.14  & 9.0$\pm$0.22 \\
          & NI & 55.6$\pm$0.35 & 46.9$\pm$0.41  & 40.8$\pm$0.28  & \textbf{100.0$\pm$0.0}*   & 17.5$\pm$0.57  & 17.6$\pm$0.42  & 9.2$\pm$0.24 \\
          & VMI & 75.0$\pm$0.40  & 69.2$\pm$0.59  & 63.0$\pm$0.84  & \textbf{100.0$\pm$0.0}*  & 35.9$\pm$0.41  & 35.7$\pm$0.87  & 24.1$\pm$0.57 \\
          & EMI & 74.3$\pm$0.65  & 71.7$\pm$0.47  & 62.6$\pm$0.29  & \textbf{100.0$\pm$0.0}*   & 25.7$\pm$0.74  & 24.6$\pm$0.98  & 13.3$\pm$0.68 \\
          & RAP &  80.4$\pm$0.75  & 75.5$\pm$0.56 & 68.0$\pm$0.84  & 99.9$\pm$0.10* & 40.3$\pm$0.47  & 39.9$\pm$0.73  &  30.4$\pm$1.03 \\
          & \cellcolor[rgb]{0.9, 0.9, 0.9}PGN & \cellcolor[rgb]{0.9, 0.9, 0.9}\textbf{86.2$\pm$0.84} & \cellcolor[rgb]{0.9, 0.9, 0.9}\textbf{83.3$\pm$0.66} & \cellcolor[rgb]{0.9, 0.9, 0.9}\textbf{77.8$\pm$0.69} & \cellcolor[rgb]{0.9, 0.9, 0.9}\textbf{100.0$\pm$0.0}* & \cellcolor[rgb]{0.9, 0.9, 0.9}\textbf{63.1$\pm$1.32} & \cellcolor[rgb]{0.9, 0.9, 0.9}\textbf{62.9$\pm$0.74} & \cellcolor[rgb]{0.9, 0.9, 0.9}\textbf{50.8$\pm$0.88} \\
    \hline
    \end{tabular}%
  \label{tab:attack_single_model}%
\end{table}%

We conduct a series of gradient-based attacks under a single-model setting and report the attack success rates, which indicate the misclassification rates of the target models when using adversarial examples as inputs. The adversarial examples are generated on four different models: Inc-v3, Inc-v4, IncRes-v2, and Res-101, respectively. The results are summarized in Table \ref{tab:attack_single_model}.

From the results, we observe that our PGN method not only maintains a high attack success rate on white-box models but also significantly improves the attack success rate on black-box models. For example, when generating adversarial examples on Inc-v3,  VMI-FGSM, EMI-FGSM, RAP achieve the attack success rates of $74.8\%$, $80.7\%$ and $84.5\%$, respectively on Inc-v4. In comparison, our PGN method achieves an impressive attack success rate of $90.6\%$, surpassing RAP (the best baseline) by a margin of $6.1\%$. Moreover, when targeting adversarially trained models, our PGN attack method consistently outperforms other gradient-based attacks, improving the attack success rate by at least $11.8\%$ compared to state-of-the-art methods on average. This convincingly validates the high effectiveness of our proposed method against both normally trained and adversarially trained models. Such outstanding results highlight the effectiveness of locating adversarial examples in flat regions for improved transferability, which is consistent with our motivation.

\subsection{Attack an Ensemble of Models}
\label{sec:exp:ens}
In addition to attacking a single model, we also evaluate the performance of our PGN method in an ensemble-model setting to further validate its effectiveness. In this subsection, we adopt the ensemble attack method in \citep{dong2018boosting}, which creates an ensemble by averaging the logit outputs of different models. Specifically, the adversaries are generated by integrating three normally trained models, including Inc-v3, Inc-v4, and IncRes-v2. All the ensemble models are assigned equal weights and we test the performance of transferability on both normally trained models and adversarially trained models.

The results, presented in Table \ref{tab:attack_an_ensemble_model}, demonstrate that our PGN method consistently achieves the highest attack success rates in the black-box setting. Compared to previous gradient-based attack methods, PGN achieves an average success rate of $92.78\%$, outperforming VMI-FGSM, EMI-FGSM, and RAP by $12.42\%$, $11.88\%$, and $7.37\%$, respectively. Notably, our method exhibits even greater improvements against adversarially trained models, surpassing the best attack method RAP by over $18.4\%$ on average. These results validate that incorporating penalized gradient norms into the loss function effectively enhances the transferability of adversarial attacks, which also confirms the superiority of our proposed method in adversarial attacks.

\begin{table}[t]
  \centering
  \caption{The untargeted attack success rates (\%) of various gradient-based attacks on eight models in the multi-model setting. The adversarial examples are generated on the ensemble models, \textit{i.e.} Inc-v3, Inc-v4, and IncRes-v2. Here * indicates the white-box model.}
  \vspace{1mm}
  \setlength{\tabcolsep}{2.00pt}
    \begin{tabular}{|c|cccccccc|c|}
    \hline
    Attack & Inc-v3 & Inc-v4 & IncRes-v2 & Res-101 & Res-152 & Inc-v3$_{ens3}$ & Inc-v3$_{ens4}$ & IncRes-v2$_{ens}$ & Avg.  \\
    \hline
    \hline
    MI & 99.8*  & 99.5*  & 97.8*  & 66.8  & 68.4  & 36.7  & 36.9  & 22.6  & 66.06 \\
    NI & \textbf{100.0}* & \textbf{99.9}* & 99.6* & 74.6  & 74.0  & 37.7  & 37.1  & 22.2  & 68.14 \\
    VMI & 99.9*  & 99.5*  & 98.0*  & 82.2  & 81.6  & 66.8  & 64.1  & 50.8  & 80.36 \\
    EMI & \textbf{100.0}* & \textbf{99.9}*  & \textbf{99.7}* & 92.3  & 93.0  & 62.8  & 59.5  & 40.0    & 80.90 \\
    RAP & \textbf{100.0}* & 99.4* & 98.2* & 93.5 & 93.4 & 70.1 & 69.8 & 58.9 & 85.41\\
    \cellcolor[rgb]{0.9, 0.9, 0.9}PGN & \cellcolor[rgb]{0.9, 0.9, 0.9}\textbf{100.0}* & \cellcolor[rgb]{0.9, 0.9, 0.9}\textbf{99.9}* & \cellcolor[rgb]{0.9, 0.9, 0.9}99.6*  & \cellcolor[rgb]{0.9, 0.9, 0.9}\textbf{94.2} & \cellcolor[rgb]{0.9, 0.9, 0.9}\textbf{94.6} & \cellcolor[rgb]{0.9, 0.9, 0.9}\textbf{88.2} & \cellcolor[rgb]{0.9, 0.9, 0.9}\textbf{86.6} & \cellcolor[rgb]{0.9, 0.9, 0.9}\textbf{79.2} & \cellcolor[rgb]{0.9, 0.9, 0.9}\textbf{92.78} \\
    \hline
    \end{tabular}%
  \label{tab:attack_an_ensemble_model}%
\end{table}%

\subsection{Combined with Input Transformation Attacks}
\label{sec:exp:combine}
\begin{table}[t]
  \centering
  \caption{The untargeted attack success rates (\%) of our PGN method, when it is integrated with DIM, TIM, SIM, Admix, and SSA, respectively. The adversarial examples are generated on Inc-v3. Here * indicates the white-box model.}
  \vspace{1mm}
  \setlength{\tabcolsep}{2.00pt}
    \begin{tabular}{|c|ccccccc|c|}
    \hline
    Attack & Inc-v3 & Inc-v4 & IncRes-v2 & Res-101 & Inc-v3$_{ens3}$ & Inc-v3$_{ens4}$ & IncRes-v2$_{ens}$ & Avg.  \\
    \hline
    \hline
    DIM   & 99.7*  & 72.2  & 67.3  & 62.8  & 32.8    & 30.7  & 16.4  &  54.56\\
    \cellcolor[rgb]{0.9, 0.9, 0.9}PGN-DIM & \cellcolor[rgb]{0.9, 0.9, 0.9}\textbf{100.0}*  & \cellcolor[rgb]{0.9, 0.9, 0.9}\textbf{93.6} & \cellcolor[rgb]{0.9, 0.9, 0.9}\textbf{91.9}  & \cellcolor[rgb]{0.9, 0.9, 0.9}\textbf{87.3} & \cellcolor[rgb]{0.9, 0.9, 0.9}\textbf{78.3}  & \cellcolor[rgb]{0.9, 0.9, 0.9}\textbf{77.5}  & \cellcolor[rgb]{0.9, 0.9, 0.9}\textbf{59.8} & \cellcolor[rgb]{0.9, 0.9, 0.9}\textbf{84.06} \\
    \hline
    TIM   & 99.9*   & 51.6  & 47.2  & 47.8  & 29.6  & 30.7  & 20.5  & 46.76 \\
    \cellcolor[rgb]{0.9, 0.9, 0.9}PGN-TIM & \cellcolor[rgb]{0.9, 0.9, 0.9}\textbf{100.0}*  & \cellcolor[rgb]{0.9, 0.9, 0.9}\textbf{87.6}  & \cellcolor[rgb]{0.9, 0.9, 0.9}\textbf{84.1} & \cellcolor[rgb]{0.9, 0.9, 0.9}\textbf{75.0} &  \cellcolor[rgb]{0.9, 0.9, 0.9}\textbf{78.1} & \cellcolor[rgb]{0.9, 0.9, 0.9}\textbf{77.6} & \cellcolor[rgb]{0.9, 0.9, 0.9}\textbf{65.5} &  \cellcolor[rgb]{0.9, 0.9, 0.9}\textbf{81.13}\\
    \hline
    SIM   & \textbf{100.0}* & 70.5  & 68.2  & 63.8  & 37.5  & 37.8  & 22.0  & 57.11 \\
    \cellcolor[rgb]{0.9, 0.9, 0.9}PGN-SIM & \cellcolor[rgb]{0.9, 0.9, 0.9}\textbf{100.0}* & \cellcolor[rgb]{0.9, 0.9, 0.9}\textbf{92.5} & \cellcolor[rgb]{0.9, 0.9, 0.9}\textbf{91.2} & \cellcolor[rgb]{0.9, 0.9, 0.9}\textbf{84.0} & \cellcolor[rgb]{0.9, 0.9, 0.9}\textbf{76.1}  & \cellcolor[rgb]{0.9, 0.9, 0.9}\textbf{75.7} & \cellcolor[rgb]{0.9, 0.9, 0.9}\textbf{59.0} & \cellcolor[rgb]{0.9, 0.9, 0.9}\textbf{82.64} \\
    \hline
    Admix & \textbf{100.0}*   & 78.6  & 77.3  & 69.5   & 41.6  & 40.3  & 24.1  & 61.63 \\
    \cellcolor[rgb]{0.9, 0.9, 0.9}PGN-Admix & \cellcolor[rgb]{0.9, 0.9, 0.9}\textbf{100.0}* & \cellcolor[rgb]{0.9, 0.9, 0.9}\textbf{93.1} & \cellcolor[rgb]{0.9, 0.9, 0.9}\textbf{92.2} & \cellcolor[rgb]{0.9, 0.9, 0.9}\textbf{85.5} & \cellcolor[rgb]{0.9, 0.9, 0.9}\textbf{76.9}  & \cellcolor[rgb]{0.9, 0.9, 0.9}\textbf{77.2} & \cellcolor[rgb]{0.9, 0.9, 0.9}\textbf{60.2} & \cellcolor[rgb]{0.9, 0.9, 0.9}\textbf{83.57} \\
    \hline
    SSA   & 99.7*  & 88.3  & 86.8  & 77.7  & 56.7  & 55.3  & 35.2  & 71.39 \\
    \cellcolor[rgb]{0.9, 0.9, 0.9}PGN-SSA & \cellcolor[rgb]{0.9, 0.9, 0.9}\textbf{99.8}* & \cellcolor[rgb]{0.9, 0.9, 0.9}\textbf{89.9} &\cellcolor[rgb]{0.9, 0.9, 0.9}\textbf{89.7} &\cellcolor[rgb]{0.9, 0.9, 0.9}\textbf{82.9} & \cellcolor[rgb]{0.9, 0.9, 0.9}\textbf{69.2} &\cellcolor[rgb]{0.9, 0.9, 0.9}\textbf{67.8} &\cellcolor[rgb]{0.9, 0.9, 0.9}\textbf{47.1} & \cellcolor[rgb]{0.9, 0.9, 0.9}\textbf{78.03} \\
    \hline
    \end{tabular}%
  \label{tab:attack_with_input_transformation}%
\end{table}%

Existing input transformation-based attacks have shown great compatibility with each other. Similarly, due to the simple and efficient gradient update process, our proposed PGN method can also be combined with these input transformation-based methods to improve the transferability of adversarial examples. To further demonstrate the efficacy of the proposed PGN method, we integrate our method into these input transformations \ie, DIM, TIM, SIM, Admix, and SSA. We generate adversarial examples on the Inc-v3 model and test the transferability of adversarial examples on six black-box models.

The experimental results are shown in Table \ref{tab:attack_with_input_transformation}. When combined with our gradient update strategy, it can significantly improve the adversarial transferability of these input transformation-based attack methods in the black-box setting. At the same time, our method also has great improvement after combining these methods. For example, DIM only achieves an average success rate of $54.56\%$ on the seven models, while when combined with our PGN method it can achieve an average rate of $84.06\%$, which is $29.5\%$ higher than before. Especially, after combining these input transformation-based methods, our PGN tends to achieve much better results on the ensemble adversarially trained models, compared with the results in Table \ref{tab:attack_single_model}. It is a significant improvement and shows that our method has good scalability and can be combined with existing methods to further improve the success rate of transfer-based black-box attacks. In addition, our PGN method can also be combined with various gradient-based attack methods to enhance the transferability of previous works. The more experimental results are shown in the Appendix.

\subsection{Ablation Study on Finite Difference Method}
\label{sec:ablation}
In this subsection, we will analyze and experimentally verify the effectiveness of the finite difference method in accelerating the approximation of the second-order Hessian matrix. We first theoretically analyze the acceleration effect of the finite difference method and substantiate our theoretical analysis with comparative experiments.

\textbf{Theoretical analysis.} For the baseline attack method, I-FGSM \citep{kurakin2018adversarial}, the gradient is computed only once per iteration. Thus, its computational complexity is $O(n)$, where $n$ represents the image size. However, when we introduce the penalty gradient term, the need arises to compute the second-order Hessian matrix, leading to a theoretical computational complexity of $O(n^2)$.  To address this, we propose the finite difference method as an approximation to the Hessian matrix, which requires the computation of the gradient twice in each iteration, effectively yielding a computational complexity of $O(2n)$. This theoretically promises significant improvements in computational efficiency.

\textbf{Experimental comparison.} To validate the effectiveness of the finite difference method in accelerating computation, we compare the runtime and computational memory before and after using the finite-difference method. These experiments were conducted using codes executed on an RTX 2080 Ti with a CUDA environment. We employed I-FGSM and evaluated the total running time on 1,000 images (excluding data loading time) and the attack success rate on black-box models. The outcomes are presented in Table \ref{tab:ablation_fdm}. Directly optimizing Eq.~\eqref{eq:appro_reg_loss} results in better attack performance with high computational resources. With the finite difference method (FDM), we can better approximate the performance of direct optimization of the second-order Hessian matrix, which significantly reduces the running time and the computational memory. Furthermore, owing to the relatively modest image size and the comparatively small number of parameters compared to the model, the accelerated computing capabilities of CUDA enable the actual running time to be less than the theoretical estimates.
\begin{table}[t]
  \centering
  \caption{Comparison of the approximation effect between directly optimizing the Hessian matrix ($H_m$) and using the Finite Difference Method (FDM) to approximate. "Time" represents the total running time on 1,000 images, and "Memory" represents the computing memory size.}
    \setlength{\tabcolsep}{2.50pt}
    \begin{tabular}{|c|cc|ccccc|c|c|}
    \hline
    Attack & $H_m$ & FDM   & Inc-v3 & Inc-v4 & IncRes-v2 & Res-101 & Res-152 & Time (s) & Memory (MiB) \\
    \hline
    \hline
    \multirow{3}[2]{*}{I-FGSM} & \ding{55} & \ding{55} & \textbf{100.0}*  & 27.8  & 19.1  & 38.1  & 35.2  & 52.31  & 1631 \\
          & \ding{51} & \ding{55} & \textbf{100.0}*   & \textbf{39.2}  & \textbf{30.2}  & \textbf{47.0}  & \textbf{45.5}  & \textbf{469.54}  & \textbf{7887} \\
          & \ding{51} & \ding{51} & \textbf{100.0}*   & 37.9  & 28.6  & 45.7  & 44.6  & 96.42   & 1631 \\
    \hline
    \end{tabular}%
  \label{tab:ablation_fdm}%
\end{table}%

\subsection{Ablation Study on Hyper-parameters}
\label{sec:ablation_hy}
In this subsection, we conduct a series of ablation experiments to study the impact of different parameters, including the balanced coefficient $\delta$ and the upper bound of $\zeta$-ball, and the study of the number of sampled examples $N$ will be illustrated in the Appendix. To simplify our analysis, we only consider the transferability of adversarial examples crafted on the Inc-v3 model.

\begin{figure}
    \centering
    \begin{subfigure}{0.45\linewidth}
        \centering
        \begin{tikzpicture}[clip]
        \begin{axis}[
        	xlabel={\tiny The balanced coefficient $\delta$}, 
        	ylabel={\tiny Attack success rates (\%)},,
        	grid=both,
        	minor grid style={gray!25, dashed},
        	major grid style={gray!25, dashed},
            scale only axis,
        	width=0.8\linewidth,
            ylabel style={font=\tiny, yshift=-5pt},
            xlabel style={font=\tiny, yshift=5pt},
            xtick=data,
            tick label style={font=\tiny},
            legend style={font=\tiny, fill opacity=0.8,legend columns=2},
            legend pos=south west
        ]
            \addplot[line width=1pt,solid,mark=*,color=cyan, mark options={mark size=1pt}] %
            	table[x=delta,y=Incv4,col sep=comma]{images/Parameters/search_delta.csv};
             \addlegendentry{Inc-v4};

            \addplot[line width=1pt,solid,mark=pentagon,color=purple, mark options={mark size=1pt}] %
            	table[x=delta,y=IncRes_v2,col sep=comma]{images/Parameters/search_delta.csv};
             \addlegendentry{IncRes-v2};

            \addplot[line width=1pt,solid,mark=triangle,color=teal, mark options={mark size=1pt}] %
            	table[x=delta,y=Res101,col sep=comma]{images/Parameters/search_delta.csv};
             \addlegendentry{Res-101};

            \addplot[line width=1pt,solid,mark=square,color=orange, mark options={mark size=1pt}] %
            	table[x=delta,y=Res152,col sep=comma]{images/Parameters/search_delta.csv};
             \addlegendentry{Res-152};


            \addplot[line width=1pt,solid,mark=oplus,color=red, mark options={mark size=1pt}] %
            	table[x=delta,y=Incv3_ens4,col sep=comma]{images/Parameters/search_delta.csv};
             \addlegendentry{Inc-v3$_{ens4}$};

            \addplot[line width=1pt,solid,mark=diamond,color=violet, mark options={mark size=1pt}] %
            	table[x=delta,y=IncRes_v2_ens,col sep=comma]{images/Parameters/search_delta.csv};
             \addlegendentry{IncRes-v2$_{ens}$};
        \end{axis}
        \end{tikzpicture}
        \caption{\scriptsize The hyper-parameter $\delta$}
        \label{fig:parameters_studies:delta}
    \end{subfigure}
    \begin{subfigure}{0.45\linewidth}
        \centering
        \begin{tikzpicture}[clip]
        \begin{axis}[
        	xlabel={\tiny The upper bound of neighborhood $\zeta$}, 
        	ylabel={\tiny Attack success rates (\%)},,
        	grid=both,
        	minor grid style={gray!25},
        	major grid style={gray!25},
            scale only axis,
        	width=0.8\linewidth,
            ylabel style={font=\tiny, yshift=-5pt},
            xlabel style={font=\tiny, yshift=5pt},
            xtick=data,
            xticklabels={0, $0.5 \epsilon$, $1 \epsilon$, $1.5 \epsilon$, $2 \epsilon$, $2.5 \epsilon$, $3 \epsilon$, $3.5 \epsilon$, $4 \epsilon$, $4.5 \epsilon$, $5 \epsilon$},
            tick label style={font=\tiny},
            legend style={font=\tiny, fill opacity=0.8,legend columns=2},
            legend pos=south west
        ]
            \addplot[line width=1pt,solid,mark=*,color=cyan, mark options={mark size=1pt}] %
            	table[x=zeta,y=Incv4,col sep=comma]{images/Parameters/search_zeta.csv};
             \addlegendentry{Inc-v4};

            \addplot[line width=1pt,solid,mark=pentagon,color=purple, mark options={mark size=1pt}] %
            	table[x=zeta,y=IncRes_v2,col sep=comma]{images/Parameters/search_zeta.csv};
             \addlegendentry{IncRes-v2};

            \addplot[line width=1pt,solid,mark=triangle,color=teal, mark options={mark size=1pt}] %
            	table[x=zeta,y=Res101,col sep=comma]{images/Parameters/search_zeta.csv};
             \addlegendentry{Res-101};

            \addplot[line width=1pt,solid,mark=square,color=orange, mark options={mark size=1pt}] %
            	table[x=zeta,y=Res152,col sep=comma]{images/Parameters/search_zeta.csv};
             \addlegendentry{Res-152};

            \addplot[line width=1pt,solid,mark=oplus,color=red, mark options={mark size=1pt}] %
            	table[x=zeta,y=Incv3_ens4,col sep=comma]{images/Parameters/search_zeta.csv};
             \addlegendentry{Inc-v3$_{ens4}$};

            \addplot[line width=1pt,solid,mark=diamond,color=violet, mark options={mark size=1pt}] %
            	table[x=zeta,y=IncRes_v2_ens,col sep=comma]{images/Parameters/search_zeta.csv};
             \addlegendentry{IncRes-v2$_{ens}$};
        \end{axis}
        \end{tikzpicture}
        \caption{\scriptsize The hyper-parameter $\zeta$}
        \label{fig:parameters_studies:zeta}
    \end{subfigure}
    \caption{The untargeted attack success rates (\%) on six black-box models with various hyper-parameters $\delta$ or $\zeta$. The adversarial examples are generated by PGN on Inc-v3.}
    \label{fig:parameters_studies}
\end{figure}
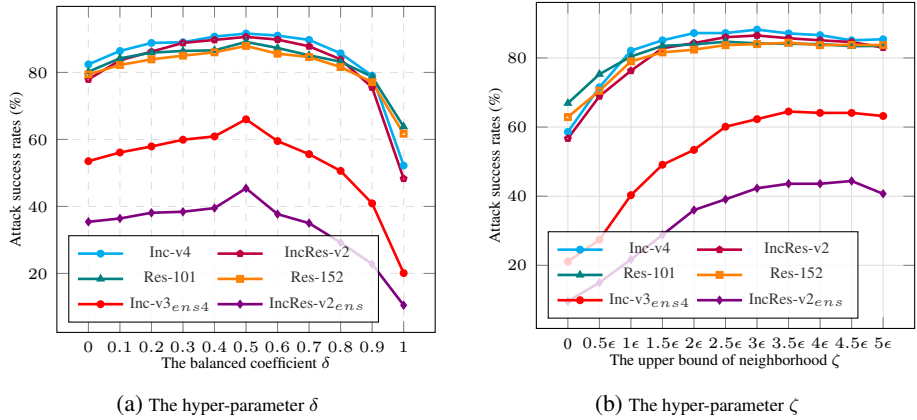

\textbf{The balanced coefficient $\delta$.} In Sec.~\ref{sec:grm}, we introduce a balanced coefficient $\delta$ to represent the penalty coefficient $\lambda$ (see the Appendix for details). Compared to studying the parameter of $\lambda$, studying $\delta$ will be much easier because it divides the scope of the parameter learning. As shown in Fig.~\ref{fig:parameters_studies:delta}, we study the influence of the $\delta$ in the black-box setting where $\zeta$ is fixed to $3.0 \times \epsilon$. As we increase $\delta$, the transferability increases and achieves the peak for these black-box models when $\delta=0.5$. This indicates that when averaging these two gradients the performance is best. Therefore, we set $\delta=0.5$ in our experiments.

\textbf{The upper bound of $\zeta$-ball.} In Fig.~\ref{fig:parameters_studies:zeta}, we study the influence of the upper bound neighborhood size for random sampling, determined by parameter $\zeta$, on the success rates in the black-box setting. In our experiments, we use uniform sampling, which reduces the bias due to uneven sample distribution. As we increase $\zeta$, the transferability increases and achieves the peak for normally trained models when $\zeta=3.0 \times \epsilon$, but it is still increasing against the adversarially trained models. When $\zeta > 4.5 \times \epsilon$, the performance of adversarial transferability will decrease on seven black-box models. To achieve the trade-off for the transferability on normally trained models and adversarially trained models, we set $\zeta=3.0 \times \epsilon$ in our experiments.

\section{Conclusion}
Inspired by the observation that flat local minima often result in better generalization, we hypothesize and empirically validate that adversarial examples at a flat local region tend to have better adversarial transferability. Intuitively, we can optimize the perturbation with a gradient regularize in the neighborhood of the input sample to generate an adversarial example in a flat local region but it is non-trivial to solve such an objective function. To address such an issue, we propose a novel attack method called Penalizing Gradient Norm (PGN). Specifically, PGN approximates the Hessian/vector product by interpolating the first-order gradients of two samples. To better explore its neighborhood, PGN adopts the average gradient of several randomly sampled data points to update the adversarial perturbation. Extensive experiments on the ImageNet-compatible dataset demonstrate that PGN can generate adversarial examples at more flat local regions and achieve much better transferability than existing transfer-based attacks. Our PGN can be seamlessly integrated with other gradient-based and input transformation-based attacks to further improve adversarial transferability, demonstrating its versatility and ability to improve adversarial transferability across different attack scenarios.

\section{Limitation}
Although we have experimentally verified that flat local minima can improve the transferability of adversarial attacks, there is still a lack of theoretical analysis regarding the relationship between flatness and transferability. An existing perspective is that transferability may be related to the generalization ability of flatness. We will also keep studying the theoretical connection between transferability and flat local minima in our future work. We hope our work sheds light on the potential of flat local maxima in generating transferable adversarial examples and provides valuable insights for further exploration in the field of adversarial attacks.

\section*{Acknowledgments}
We want to thank the anonymous reviewers for their valuable suggestions and comments. This work was supported by the National Natural Science Foundation of China (Nos.\ 62072334, 6227071567, and 61976164), and the National Science Basic Research Plan in Shaanxi Province of China (No.\ 2022GY-061).

\bibliographystyle{plainnat}
\bibliography{abbr,reference}

\newcommand{\arxiv}{arXiv preprint arXiv:}
\begin{thebibliography}{66}
\providecommand{\natexlab}[1]{#1}
\providecommand{\url}[1]{\texttt{#1}}
\expandafter\ifx\csname urlstyle\endcsname\relax
  \providecommand{\doi}[1]{doi: #1}\else
  \providecommand{\doi}{doi: \begingroup \urlstyle{rm}\Url}\fi

\bibitem[Andrei(2009)]{andrei2009accelerated}
Neculai Andrei.
\newblock Accelerated conjugate gradient algorithm with finite difference
  hessian/vector product approximation for unconstrained optimization.
\newblock \emph{Journal of Computational and Applied Mathematics}, 230\penalty0
  (2):\penalty0 570--582, 2009.

\bibitem[Bayer et~al.(2022)Bayer, Kaufhold, and Reuter]{bayer2022survey}
Markus Bayer, Marc-Andr{\'e} Kaufhold, and Christian Reuter.
\newblock A survey on data augmentation for text classification.
\newblock \emph{ACM Computing Surveys}, 55\penalty0 (7):\penalty0 1--39, 2022.

\bibitem[Brendel et~al.(2017)Brendel, Rauber, and Bethge]{brendel2017decision}
Wieland Brendel, Jonas Rauber, and Matthias Bethge.
\newblock Decision-based adversarial attacks: Reliable attacks against
  black-box machine learning models.
\newblock \emph{arXiv preprint arXiv:1712.04248}, 2017.

\bibitem[Byun et~al.(2022)Byun, Cho, Kwon, Kim, and
  Kim]{JunyoungByun2022ImprovingTT}
Junyoung Byun, Seungju Cho, Myung-Joon Kwon, Hee-Seon Kim, and Changick Kim.
\newblock Improving the transferability of targeted adversarial examples
  through object-based diverse input.
\newblock In \emph{Proceedings of the IEEE/CVF Conference on Computer Vision
  and Pattern Recognition}, pages 15244--15253, 2022.

\bibitem[Carlini and Wagner(2017)]{carlini2017towards}
Nicholas Carlini and David Wagner.
\newblock Towards evaluating the robustness of neural networks.
\newblock In \emph{2017 IEEE Symposium on Security and Privacy}, pages 39--57,
  2017.

\bibitem[Chen et~al.(2017)Chen, Zhang, Sharma, Yi, and
  Hsieh]{PinYuChen2017ZOOZO}
Pin-Yu Chen, Huan Zhang, Yash Sharma, Jinfeng Yi, and Cho-Jui Hsieh.
\newblock Zoo: Zeroth order optimization based black-box attacks to deep neural
  networks without training substitute models.
\newblock \emph{In Proceedings of the 10th ACM Workshop on Artificial
  Intelligence and Security}, 2017.

\bibitem[Chen et~al.(2021)Chen, Xie, Niu, Liu, Wei, and
  Tian]{chen2021visformer}
Zhengsu Chen, Lingxi Xie, Jianwei Niu, Xuefeng Liu, Longhui Wei, and Qi~Tian.
\newblock Visformer: The vision-friendly transformer.
\newblock In \emph{Proceedings of the IEEE/CVF International Conference on
  Computer Vision}, pages 589--598, 2021.

\bibitem[Chlap et~al.(2021)Chlap, Min, Vandenberg, Dowling, Holloway, and
  Haworth]{chlap2021review}
Phillip Chlap, Hang Min, Nym Vandenberg, Jason Dowling, Lois Holloway, and
  Annette Haworth.
\newblock A review of medical image data augmentation techniques for deep
  learning applications.
\newblock \emph{Journal of Medical Imaging and Radiation Oncology}, 65\penalty0
  (5):\penalty0 545--563, 2021.

\bibitem[Cohen et~al.(2019)Cohen, Rosenfeld, and Kolter]{cohen2019certified}
Jeremy Cohen, Elan Rosenfeld, and Zico Kolter.
\newblock Certified adversarial robustness via randomized smoothing.
\newblock In \emph{Proceedings of International Conference on Machine
  Learning}, pages 1310--1320. PMLR, 2019.

\bibitem[Dong et~al.(2018)Dong, Liao, Pang, Su, Zhu, Hu, and
  Li]{dong2018boosting}
Yinpeng Dong, Fangzhou Liao, Tianyu Pang, Hang Su, Jun Zhu, Xiaolin Hu, and
  Jianguo Li.
\newblock Boosting adversarial attacks with momentum.
\newblock In \emph{Proceedings of the IEEE/CVF Conference on Computer Vision
  and Pattern Recognition}, pages 9185--9193, 2018.

\bibitem[Dong et~al.(2019)Dong, Pang, Su, and Zhu]{dong2019evading}
Yinpeng Dong, Tianyu Pang, Hang Su, and Jun Zhu.
\newblock Evading defenses to transferable adversarial examples by
  translation-invariant attacks.
\newblock In \emph{Proceedings of the IEEE/CVF Conference on Computer Vision
  and Pattern Recognition}, pages 4312--4321, 2019.

\bibitem[Dosovitskiy et~al.(2021)Dosovitskiy, Beyer, Kolesnikov, Weissenborn,
  Zhai, Unterthiner, Dehghani, Minderer, Heigold, Gelly, Uszkoreit, and
  Houlsby]{dosovitskiy2020vit}
Alexey Dosovitskiy, Lucas Beyer, Alexander Kolesnikov, Dirk Weissenborn,
  Xiaohua Zhai, Thomas Unterthiner, Mostafa Dehghani, Matthias Minderer, Georg
  Heigold, Sylvain Gelly, Jakob Uszkoreit, and Neil Houlsby.
\newblock An image is worth 16x16 words: Transformers for image recognition at
  scale.
\newblock \emph{In Proceedings of International Conference on Learning
  Representations}, 2021.

\bibitem[Evtimov et~al.(2017)Evtimov, Eykholt, Fernandes, Kohno, Li, Prakash,
  Rahmati, and Song]{evtimov2017robust}
Ivan Evtimov, Kevin Eykholt, Earlence Fernandes, Tadayoshi Kohno, Bo~Li, Atul
  Prakash, Amir Rahmati, and Dawn Song.
\newblock Robust physical-world attacks on machine learning models.
\newblock \emph{arXiv preprint arXiv:1707.08945}, 2\penalty0 (3):\penalty0 4,
  2017.

\bibitem[Foret et~al.(2021)Foret, Kleiner, Mobahi, and
  Neyshabur]{foret2020sharpness}
Pierre Foret, Ariel Kleiner, Hossein Mobahi, and Behnam Neyshabur.
\newblock Sharpness-aware minimization for efficiently improving
  generalization.
\newblock \emph{In Proceedings of International Conference on Learning
  Representations}, 2021.

\bibitem[Ge et~al.(2023)Ge, Shang, Liu, Liu, Wan, Feng, and
  Wang]{ge2023improving}
Zhijin Ge, Fanhua Shang, Hongying Liu, Yuanyuan Liu, Liang Wan, Wei Feng, and
  Xiaosen Wang.
\newblock {Improving the Transferability of Adversarial Examples with Arbitrary
  Style Transfer}.
\newblock In \emph{Proceedings of the ACM International Conference on
  Multimedia}, 2023.

\bibitem[Goodfellow et~al.(2015)Goodfellow, Shlens, and
  Szegedy]{goodfellow2014explaining}
Ian~J Goodfellow, Jonathon Shlens, and Christian Szegedy.
\newblock Explaining and harnessing adversarial examples.
\newblock In \emph{Proceedings of International Conference on Learning
  Representations}, 2015.

\bibitem[Guo et~al.(2018)Guo, Rana, Cisse, and van~der
  Maaten]{guo2018countering}
Chuan Guo, Mayank Rana, Moustapha Cisse, and Laurens van~der Maaten.
\newblock Countering adversarial images using input transformations.
\newblock In \emph{Proceedings of International Conference on Learning
  Representations}, 2018.

\bibitem[He et~al.(2016)He, Zhang, Ren, and Sun]{KaimingHe2015DeepRL}
Kaiming He, Xiangyu Zhang, Shaoqing Ren, and Jian Sun.
\newblock Deep residual learning for image recognition.
\newblock In \emph{Proceedings of the IEEE/CVF Conference on Computer Vision
  and Pattern Recognition}, pages 770--778, 2016.

\bibitem[Heo et~al.(2021)Heo, Yun, Han, Chun, Choe, and Oh]{heo2021rethinking}
Byeongho Heo, Sangdoo Yun, Dongyoon Han, Sanghyuk Chun, Junsuk Choe, and
  Seong~Joon Oh.
\newblock Rethinking spatial dimensions of vision transformers.
\newblock In \emph{Proceedings of the IEEE/CVF International Conference on
  Computer Vision}, pages 11936--11945, 2021.

\bibitem[Hochreiter and Schmidhuber(1997)]{hochreiter1997flat}
Sepp Hochreiter and J{\"u}rgen Schmidhuber.
\newblock Flat minima.
\newblock \emph{Neural Computation}, 9\penalty0 (1):\penalty0 1--42, 1997.

\bibitem[Huang and Zhang(2019)]{huang2019black}
Zhichao Huang and Tong Zhang.
\newblock Black-box adversarial attack with transferable model-based embedding.
\newblock \emph{arXiv preprint arXiv:1911.07140}, 2019.

\bibitem[Jia et~al.(2019)Jia, Wei, Cao, and Foroosh]{jia2019comdefend}
Xiaojun Jia, Xingxing Wei, Xiaochun Cao, and Hassan Foroosh.
\newblock Comdefend: An efficient image compression model to defend adversarial
  examples.
\newblock In \emph{Proceedings of the IEEE/CVF Conference on Computer Vision
  and Pattern Recognition}, pages 6084--6092, 2019.

\bibitem[Jiang et~al.(2019)Jiang, Neyshabur, Mobahi, Krishnan, and
  Bengio]{jiang2019fantastic}
Yiding Jiang, Behnam Neyshabur, Hossein Mobahi, Dilip Krishnan, and Samy
  Bengio.
\newblock Fantastic generalization measures and where to find them.
\newblock \emph{arXiv preprint arXiv:1912.02178}, 2019.

\bibitem[Keskar et~al.(2016)Keskar, Mudigere, Nocedal, Smelyanskiy, and
  Tang]{keskar2016large}
Nitish~Shirish Keskar, Dheevatsa Mudigere, Jorge Nocedal, Mikhail Smelyanskiy,
  and Ping Tak~Peter Tang.
\newblock On large-batch training for deep learning: Generalization gap and
  sharp minima.
\newblock \emph{arXiv preprint arXiv:1609.04836}, 2016.

\bibitem[Krizhevsky et~al.(2009)Krizhevsky, Hinton,
  et~al.]{krizhevsky2009learning}
Alex Krizhevsky, Geoffrey Hinton, et~al.
\newblock Learning multiple layers of features from tiny images.
\newblock \emph{Technical Report}, 2009.

\bibitem[Kurakin et~al.(2018)Kurakin, Goodfellow, and
  Bengio]{kurakin2018adversarial}
Alexey Kurakin, Ian~J Goodfellow, and Samy Bengio.
\newblock Adversarial examples in the physical world.
\newblock In \emph{Artificial Intelligence Safety and Security}, pages 99--112.
  2018.

\bibitem[Li et~al.(2018)Li, Xu, Taylor, Studer, and
  Goldstein]{li2018visualizing}
Hao Li, Zheng Xu, Gavin Taylor, Christoph Studer, and Tom Goldstein.
\newblock Visualizing the loss landscape of neural nets.
\newblock \emph{In Proceedings of Advances in Neural Information Processing
  Systems}, 31, 2018.

\bibitem[Liang et~al.(2020)Liang, Wei, Yao, and Cao]{liang2020efficient}
Siyuan Liang, Xingxing Wei, Siyuan Yao, and Xiaochun Cao.
\newblock Efficient adversarial attacks for visual object tracking.
\newblock In \emph{European Conference on Computer Vision}, pages 34--50, 2020.

\bibitem[Liao et~al.(2018)Liao, Liang, Dong, Pang, Hu, and
  Zhu]{liao2018defense}
Fangzhou Liao, Ming Liang, Yinpeng Dong, Tianyu Pang, Xiaolin Hu, and Jun Zhu.
\newblock Defense against adversarial attacks using high-level representation
  guided denoiser.
\newblock In \emph{Proceedings of the IEEE Conference on Computer Vision and
  Pattern Recognition}, pages 1778--1787, 2018.

\bibitem[Lin et~al.(2022)Lin, Hsu, Chen, and Yu]{lin2022real}
Chang-Sheng Lin, Chia-Yi Hsu, Pin-Yu Chen, and Chia-Mu Yu.
\newblock Real-world adversarial examples via makeup.
\newblock In \emph{Proceedings of the IEEE International Conference on
  Acoustics, Speech and Signal Processing (ICASSP)}, pages 2854--2858, 2022.

\bibitem[Lin et~al.(2019)Lin, Song, He, Wang, and Hopcroft]{lin2019nesterov}
Jiadong Lin, Chuanbiao Song, Kun He, Liwei Wang, and John~E Hopcroft.
\newblock Nesterov accelerated gradient and scale invariance for adversarial
  attacks.
\newblock In \emph{Proceedings of International Conference on Learning
  Representations}, 2019.

\bibitem[Liu et~al.(2017)Liu, Chen, Liu, and Song]{liu2016delving}
Yanpei Liu, Xinyun Chen, Chang Liu, and Dawn Song.
\newblock Delving into transferable adversarial examples and black-box attacks.
\newblock In \emph{Proceedings of International Conference on Learning
  Representations}, 2017.

\bibitem[Liu et~al.(2021)Liu, Lin, Cao, Hu, Wei, Zhang, Lin, and
  Guo]{liu2021swin}
Ze~Liu, Yutong Lin, Yue Cao, Han Hu, Yixuan Wei, Zheng Zhang, Stephen Lin, and
  Baining Guo.
\newblock Swin transformer: Hierarchical vision transformer using shifted
  windows.
\newblock In \emph{Proceedings of the IEEE/CVF International Conference on
  Computer Vision}, pages 10012--10022, 2021.

\bibitem[Liu et~al.(2019)Liu, Liu, Liu, Xu, Lin, Wang, and Wen]{liu2019feature}
Zihao Liu, Qi~Liu, Tao Liu, Nuo Xu, Xue Lin, Yanzhi Wang, and Wujie Wen.
\newblock Feature distillation: Dnn-oriented jpeg compression against
  adversarial examples.
\newblock In \emph{Proceedings of the IEEE Conference on Computer Vision and
  Pattern Recognition}, pages 860--868, 2019.

\bibitem[Long et~al.(2022)Long, Zhang, Zeng, Gao, Liu, Zhang, and
  Song]{long2022frequency}
Yuyang Long, Qilong Zhang, Boheng Zeng, Lianli Gao, Xianglong Liu, Jian Zhang,
  and Jingkuan Song.
\newblock Frequency domain model augmentation for adversarial attack.
\newblock In \emph{Proceedings of the European Conference on Computer Vision},
  pages 549--566. Springer, 2022.

\bibitem[Madry et~al.(2017)Madry, Makelov, Schmidt, Tsipras, and
  Vladu]{AleksanderMadry2018TowardsDL}
Aleksander Madry, Aleksandar Makelov, Ludwig Schmidt, Dimitris Tsipras, and
  Adrian Vladu.
\newblock Towards deep learning models resistant to adversarial attacks.
\newblock \emph{arXiv preprint arXiv:1706.06083}, 2017.

\bibitem[Miko{\l}ajczyk and Grochowski(2018)]{mikolajczyk2018data}
Agnieszka Miko{\l}ajczyk and Micha{\l} Grochowski.
\newblock Data augmentation for improving deep learning in image classification
  problem.
\newblock In \emph{2018 International Interdisciplinary PhD Workshop (IIPhDW)},
  pages 117--122. IEEE, 2018.

\bibitem[Moosavi-Dezfooli et~al.(2016)Moosavi-Dezfooli, Fawzi, and
  Frossard]{moosavi2016deepfool}
Seyed-Mohsen Moosavi-Dezfooli, Alhussein Fawzi, and Pascal Frossard.
\newblock Deepfool: a simple and accurate method to fool deep neural networks.
\newblock In \emph{Proceedings of the IEEE/CVF Conference on Computer Vision
  and Pattern Recognition}, pages 2574--2582, 2016.

\bibitem[Naseer et~al.(2020)Naseer, Khan, Hayat, Khan, and
  Porikli]{naseer2020self}
Muzammal Naseer, Salman Khan, Munawar Hayat, Fahad~Shahbaz Khan, and Fatih
  Porikli.
\newblock A self-supervised approach for adversarial robustness.
\newblock In \emph{Proceedings of the IEEE/CVF Conference on Computer Vision
  and Pattern Recognition}, pages 262--271, 2020.

\bibitem[Nesterov(1983)]{nesterov1983method}
Yurii Nesterov.
\newblock A method for unconstrained convex minimization problem with the rate
  of convergence o (1/k\^{} 2).
\newblock In \emph{Doklady An Ussr}, volume 269, pages 543--547, 1983.

\bibitem[Neyshabur et~al.(2017)Neyshabur, Bhojanapalli, McAllester, and
  Srebro]{neyshabur2017exploring}
Behnam Neyshabur, Srinadh Bhojanapalli, David McAllester, and Nati Srebro.
\newblock Exploring generalization in deep learning.
\newblock \emph{In Proceedings of Advances in Neural Information Processing
  Systems}, 30, 2017.

\bibitem[Qin et~al.(2022)Qin, Fan, Liu, Shen, Zhang, Wang, and
  Wu]{qin2022boosting}
Zeyu Qin, Yanbo Fan, Yi~Liu, Li~Shen, Yong Zhang, Jue Wang, and Baoyuan Wu.
\newblock Boosting the transferability of adversarial attacks with reverse
  adversarial perturbation.
\newblock \emph{arXiv preprint arXiv:2210.05968}, 2022.

\bibitem[Santurkar et~al.(2018)Santurkar, Tsipras, Ilyas, and
  Madry]{santurkar2018does}
Shibani Santurkar, Dimitris Tsipras, Andrew Ilyas, and Aleksander Madry.
\newblock How does batch normalization help optimization?
\newblock \emph{In Proceedings of Advances in Neural Information Processing
  Systems}, 31, 2018.

\bibitem[Sharif et~al.(2016)Sharif, Bhagavatula, Bauer, and
  Reiter]{sharif2016accessorize}
Mahmood Sharif, Sruti Bhagavatula, Lujo Bauer, and Michael~K Reiter.
\newblock Accessorize to a crime: Real and stealthy attacks on state-of-the-art
  face recognition.
\newblock In \emph{Proceedings of the 2016 ACM on Asia Conference on Computer
  and Communications Security}, pages 1528--1540, 2016.

\bibitem[Szegedy et~al.(2016)Szegedy, Vanhoucke, Ioffe, Shlens, and
  Wojna]{szegedy2016rethinking}
Christian Szegedy, Vincent Vanhoucke, Sergey Ioffe, Jon Shlens, and Zbigniew
  Wojna.
\newblock Rethinking the inception architecture for computer vision.
\newblock In \emph{Proceedings of the IEEE Conference on Computer Vision and
  Pattern Recognition}, pages 2818--2826, 2016.

\bibitem[Szegedy et~al.(2017)Szegedy, Ioffe, Vanhoucke, and
  Alemi]{szegedy2017inception}
Christian Szegedy, Sergey Ioffe, Vincent Vanhoucke, and Alexander~A Alemi.
\newblock Inception-v4, inception-resnet and the impact of residual connections
  on learning.
\newblock In \emph{Thirty-first AAAI Conference on Artificial Intelligence},
  2017.

\bibitem[Tramèr et~al.(2018)Tramèr, Kurakin, Papernot, Goodfellow, Boneh, and
  Mcdaniel]{tramer2018ensemble}
Florian Tramèr, Alexey Kurakin, Nicolas Papernot, Ian Goodfellow, Dan Boneh,
  and Patrick Mcdaniel.
\newblock Ensemble adversarial training: Attacks and defenses.
\newblock In \emph{Proceedings of International Conference on Learning
  Representations}, 2018.

\bibitem[Wang et~al.(2023{\natexlab{a}})Wang, He, Wang, and
  Wang]{wang2023boosting}
Kunyu Wang, Xuanran He, Wenxuan Wang, and Xiaosen Wang.
\newblock {Boosting Adversarial Transferability by Block Shuffle and Rotation}.
\newblock \emph{arXiv preprint arXiv:2308.10299}, 2023{\natexlab{a}}.

\bibitem[Wang and He(2021)]{wang2021enhancing}
Xiaosen Wang and Kun He.
\newblock Enhancing the transferability of adversarial attacks through variance
  tuning.
\newblock In \emph{Proceedings of the IEEE/CVF Conference on Computer Vision
  and Pattern Recognition}, pages 1924--1933, 2021.

\bibitem[Wang et~al.(2019)Wang, He, Song, Wang, and Hopcroft]{wang2019atgan}
Xiaosen Wang, Kun He, Chuanbiao Song, Liwei Wang, and John~E. Hopcroft.
\newblock {AT-GAN: A Generative Attack Model for Adversarial Transferring on
  Generative Adversarial Nets}.
\newblock \emph{arXiv preprint arXiv:1904.07793}, 2019.

\bibitem[Wang et~al.(2021{\natexlab{a}})Wang, He, Wang, and He]{wang2021admix}
Xiaosen Wang, Xuanran He, Jingdong Wang, and Kun He.
\newblock Admix: Enhancing the transferability of adversarial attacks.
\newblock In \emph{Proceedings of the IEEE/CVF International Conference on
  Computer Vision}, pages 16158--16167, 2021{\natexlab{a}}.

\bibitem[Wang et~al.(2021{\natexlab{b}})Wang, Lin, Hu, Wang, and
  He]{XiaosenWang2021BoostingAT}
Xiaosen Wang, Jiadong Lin, Han Hu, Jingdong Wang, and Kun He.
\newblock Boosting adversarial transferability through enhanced momentum.
\newblock \emph{In Proceedings of the British Machine Vision Conference},
  2021{\natexlab{b}}.

\bibitem[Wang et~al.(2022)Wang, Zhang, Tong, Gong, He, Li, and
  Liu]{wang2022triangle}
Xiaosen Wang, Zeliang Zhang, Kangheng Tong, Dihong Gong, Kun He, Zhifeng Li,
  and Wei Liu.
\newblock Triangle attack: A query-efficient decision-based adversarial attack.
\newblock In \emph{Proceedings of The European Conference on Computer Vision
  (ECCV)}, pages 156--174. Springer, 2022.

\bibitem[Wang et~al.(2023{\natexlab{b}})Wang, Tong, and He]{wang2023rethinking}
Xiaosen Wang, Kangheng Tong, and Kun He.
\newblock {Rethinking the Backward Propagation for Adversarial
  Transferability}.
\newblock In \emph{Proceedings of the Advances in Neural Information Processing
  Systems}, 2023{\natexlab{b}}.

\bibitem[Wang et~al.(2023{\natexlab{c}})Wang, Zhang, and
  Zhang]{wang2023structure}
Xiaosen Wang, Zeliang Zhang, and Jianping Zhang.
\newblock {Structure Invariant Transformation for better Adversarial
  Transferability}.
\newblock In \emph{Proceedings of the IEEE/CVF International Conference on
  Computer Vision}, pages 4607--4619, 2023{\natexlab{c}}.

\bibitem[Wang et~al.(2023{\natexlab{d}})Wang, Zhang, Liang, and
  Wang]{wang2023diversifying}
Zhiyuan Wang, Zeliang Zhang, Siyuan Liang, and Xiaosen Wang.
\newblock {Diversifying the High-level Features for better Adversarial
  Transferability}.
\newblock In \emph{Proceedings of the British Machine Vision Conference},
  2023{\natexlab{d}}.

\bibitem[Wei et~al.(2018)Wei, Liang, Chen, and Cao]{wei2018transferable}
Xingxing Wei, Siyuan Liang, Ning Chen, and Xiaochun Cao.
\newblock Transferable adversarial attacks for image and video object
  detection.
\newblock \emph{arXiv preprint arXiv:1811.12641}, 2018.

\bibitem[Wu et~al.(2020)Wu, Wang, Xia, Bailey, and Ma]{wu2020skip}
Dongxian Wu, Yisen Wang, Shu-Tao Xia, James Bailey, and Xingjun Ma.
\newblock Skip connections matter: On the transferability of adversarial
  examples generated with resnets.
\newblock In \emph{Proceedings of International Conference on Learning
  Representations}, 2020.

\bibitem[Wu et~al.(2021)Wu, Su, Lyu, and King]{wu2021improving}
Weibin Wu, Yuxin Su, Michael~R Lyu, and Irwin King.
\newblock Improving the transferability of adversarial samples with adversarial
  transformations.
\newblock In \emph{Proceedings of the IEEE/CVF Conference on Computer Vision
  and Pattern Recognition}, pages 9024--9033, 2021.

\bibitem[Xie et~al.(2018)Xie, Wang, Zhang, Ren, and Yuille]{xie2018mitigating}
Cihang Xie, Jianyu Wang, Zhishuai Zhang, Zhou Ren, and Alan Yuille.
\newblock Mitigating adversarial effects through randomization.
\newblock In \emph{Proceedings of International Conference on Learning
  Representations}, 2018.

\bibitem[Xie et~al.(2019)Xie, Zhang, Zhou, Bai, Wang, Ren, and
  Yuille]{xie2019improving}
Cihang Xie, Zhishuai Zhang, Yuyin Zhou, Song Bai, Jianyu Wang, Zhou Ren, and
  Alan~L Yuille.
\newblock Improving transferability of adversarial examples with input
  diversity.
\newblock In \emph{Proceedings of the IEEE/CVF Conference on Computer Vision
  and Pattern Recognition}, pages 2730--2739, 2019.

\bibitem[Xu et~al.(2018)Xu, Evans, and Qi]{xu2018feature}
Weilin Xu, David Evans, and Yanjun Qi.
\newblock Feature squeezing: Detecting adversarial examples in deep neural
  networks.
\newblock \emph{In Proceedings of Network and Distributed System Security
  Symposium}, 2018.

\bibitem[Yu et~al.(2021)Yu, Zhang, Chen, Yin, and Liu]{yu2021does}
Da~Yu, Huishuai Zhang, Wei Chen, Jian Yin, and Tie-Yan Liu.
\newblock How does data augmentation affect privacy in machine learning?
\newblock In \emph{Proceedings of the AAAI Conference on Artificial
  Intelligence}, volume~35, pages 10746--10753, 2021.

\bibitem[Zhang et~al.(2023)Zhang, tse Huang, Wang, Li, Wu, Wang, Su, and
  Lyu]{zhang2023improving}
Jianping Zhang, Jen tse Huang, Wenxuan Wang, Yichen Li, Weibin Wu, Xiaosen
  Wang, Yuxin Su, and Michael~R. Lyu.
\newblock {Improving the Transferability of Adversarial Samples by
  Path-Augmented Method}.
\newblock In \emph{Proceedings of the IEEE/CVF Conference on Computer Vision
  and Pattern Recognition}, pages 8173--8182, 2023.

\bibitem[Zhang et~al.(2018)Zhang, Foroosh, David, and Gong]{zhang2018camou}
Yang Zhang, Hassan Foroosh, Philip David, and Boqing Gong.
\newblock Camou: Learning physical vehicle camouflages to adversarially attack
  detectors in the wild.
\newblock In \emph{Proceedings of International Conference on Learning
  Representations}, 2018.

\bibitem[Zhao et~al.(2022)Zhao, Zhang, and Hu]{zhao2022penalizing}
Yang Zhao, Hao Zhang, and Xiuyuan Hu.
\newblock Penalizing gradient norm for efficiently improving generalization in
  deep learning.
\newblock In \emph{Proceedings of International Conference on Machine
  Learning}, pages 26982--26992. PMLR, 2022.

\end{thebibliography}

\newpage
\appendix
\section{Proof of Corollary 1}
In this work, we introduce a penalized gradient norm to the original loss function, which helps the adversarial examples to achieve a flat maximum. Then we randomly sample an example $x'$ in the neighborhood of the adversarial example $x^{adv}$ and simplify the objective function as follows:
\begin{equation}
\label{eq:objective_loss}
     \max_{x^{adv}\in \mathcal{B}_\epsilon(x)} \mathcal{L}(x^{adv},y;\theta) \approx \left[J(x',y;\theta)-\lambda \cdot \| \nabla_{x'}J(x', y;\theta) \|_2\right], \quad \mathrm{s.t.} \quad x'\in \mathcal{B}_\zeta(x^{adv}).
\end{equation}
Gradient-based attacks require calculating the gradient of the objective function during practical optimization, thus the gradient of the current loss function \eqref{eq:objective_loss} can be expressed as follows:
\begin{equation}
\begin{split}
\label{eq:gradient_objective_loss}
     \nabla_{x^{adv}}\mathcal{L}(x^{adv},y;\theta) &\approx \nabla_{x'}J(x',y;\theta)- \lambda \cdot \nabla_{x'}(\| \nabla_{x'}J(x', y;\theta) \|_2) \\
     &\approx\nabla_{x'}J(x',y;\theta)- \lambda \cdot \nabla_{x'}^2J(x',y;\theta) \cdot \frac{\nabla_{x'}J(x',y;\theta)}{\left \| \nabla_{x'}J(x',y;\theta) \right \|_2 } .
\end{split}
\end{equation}
In practice, it is computationally expensive to directly optimize Eq.~\eqref{eq:gradient_objective_loss}, since we need to calculate the Hessian matrix. In this work, we approximate the second-order Hessian matrix using the finite difference method to accelerate the attack process. Specifically, local Taylor expansion would be employed to approximate the operation results between the Hessian matrix and the gradient vector.
\subsection{Proof of Theorem 1}
\begin{proof}
    According to the Taylor expansion, we have
    \begin{equation}
        \label{eq:taylor_expansion}
        \nabla_{x}J(x+\Delta x,y;\theta)= \nabla_{x}J(x,y;\theta) + \nabla_{x}^2J(x,y;\theta) \Delta x + O({\left \| \Delta x\right \|^2 }) ,
    \end{equation}
    where $\Delta x = \alpha \cdot v$, $\alpha$ is a small step size, and $v$ is a normalized gradient direction vector. Here, we denote $v=-\frac{\nabla_{x}J(x,y;\theta)}{\left \| \nabla_{x}J(x,y;\theta) \right \|_2 } $.

    Therefore, the Hessian/vector product can be approximated by the first-order gradient as follows:
    \begin{equation}
        \label{eq:gradientapprox}
        \nabla_{x}^2 J(x,y;\theta)  \cdot v \approx  \frac{\nabla _{x}J(x+\alpha \cdot v,y;\theta)-\nabla_{x}J(x,y;\theta)}{\alpha}.
    \end{equation}
\end{proof}
\subsection{Proof of Corollary 1}
\begin{proof}
    From Eqs.~\eqref{eq:gradient_objective_loss} and ~\eqref{eq:gradientapprox}, the gradient of the loss function $\mathcal{L}(\cdot)$ can be expressed as:
    \begin{equation}
    	\label{eq:gradient_update}\begin{split}
            \nabla_{x^{adv}}\mathcal{L}(x^{adv},y;\theta) &\approx\nabla_{x'}J(x',y;\theta)- \lambda \cdot \nabla_{x'}^2J(x',y;\theta) \cdot \frac{\nabla_{x'}J(x',y;\theta)}{\left \| \nabla_{x'}J(x',y;\theta) \right \|_2 }\\
            & \approx \nabla_{x'}J(x',y;\theta) + \lambda \cdot \frac{\nabla _{x'}J(x'+\alpha \cdot v,y;\theta)-\nabla_{x'}J(x',y;\theta)}{\alpha} \\
            & = (1-\frac{\lambda}{\alpha}) \cdot \nabla_{x'}J(x', y;\theta)+\frac{\lambda}{\alpha} \cdot \nabla_{x'}J(x'+\alpha \cdot v, y;\theta).
            \end{split}
    \end{equation}

    We introduce a balanced coefficient $\delta$ and denote it as $\delta=\frac{\lambda}{\alpha}$.
    Hence, the gradient of the objective function \eqref{eq:objective_loss} at the $t$-th iteration can be approximated as:
    \begin{equation}
    	   \label{eq:precorre2}
            \nabla _{x_t^{adv}}\mathcal{L}(x_t^{adv},y;\theta) \approx (1-\delta)\cdot \nabla_{x_t'}J(x_t',y;\theta) + \delta \cdot \nabla_{x_t'}J({x_t'+\alpha \cdot v},y;\theta).
    \end{equation}
\end{proof}

\section{Visualization of Loss Surfaces}
\textbf{Implementation details.} Given that the adversarial example $x^{adv}$ typically has a large number of dimensions, visualizing the loss function against all dimensions becomes infeasible. To this end, we randomly select two directions, denoted as $r_1$ and $r_2$, from a Gaussian distribution with the same dimension as $x^{adv}$. Next, we calculate the loss change by varying the magnitudes of $k_1$ and $k_2$, representing the scaling factors applied to $r_1$ and $r_2$, respectively, which enables us to visualize the loss function using a two-dimensional plot. This approach provides a slice of the loss function, allowing us to analyze its behavior and understand the impact of perturbations along different directions.

\begin{figure*}
    \centering
    \begin{minipage}{0.12\textwidth}
    \caption*{Raw Image}
        \begin{subfigure}{\textwidth}
        \centering
            \includegraphics[width=\linewidth]{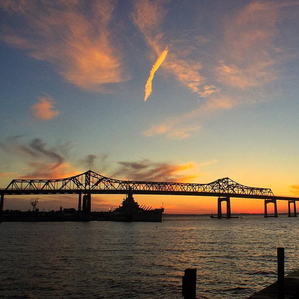}
            \vspace{-0.5cm}
        \end{subfigure}
    \end{minipage}
    \hspace{1mm}
    \begin{minipage}{0.86\textwidth}
        \begin{subfigure}{0.14\textwidth}
            \centering 
            \caption*{MI-FGSM}
            \includegraphics[width=\linewidth]{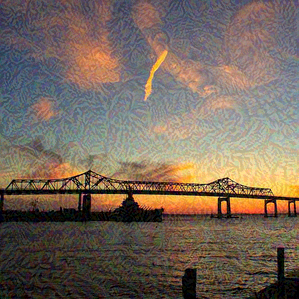}
            \vspace{-0.5cm}
        \end{subfigure}%
        \hspace{3mm}
        \begin{subfigure}{0.14\textwidth} 
            \centering 
            \caption*{NI-FGSM}
            \includegraphics[width=\linewidth]{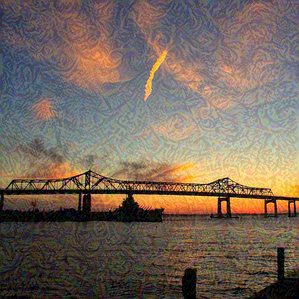}
            \vspace{-0.5cm} 
        \end{subfigure}%
        \hspace{3mm}
        \begin{subfigure}{0.14\textwidth}
            \centering 
            \caption*{VMI-FGSM}
            \includegraphics[width=\linewidth]{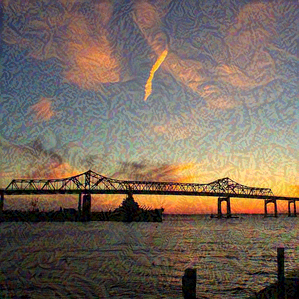}
            \vspace{-0.5cm}
            
        \end{subfigure}
        \hspace{3mm}
        \begin{subfigure}{0.14\textwidth}
            \centering 
            \caption*{EMI-FGSM}
            \includegraphics[width=\linewidth]{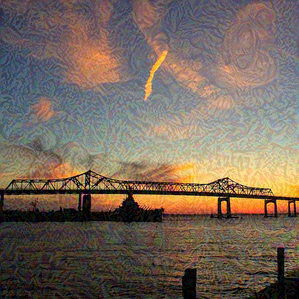}
            \vspace{-0.5cm}
            
        \end{subfigure}%
        \hspace{3mm}
        \begin{subfigure}{0.14\textwidth} 
            \centering 
            \caption*{RAP}
            \includegraphics[width=\linewidth]{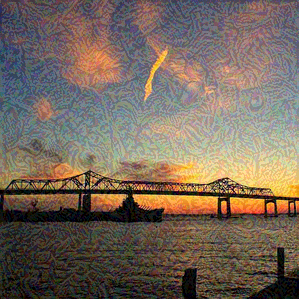}
            \vspace{-0.5cm}
            
        \end{subfigure}%
        \hspace{3mm}
        \begin{subfigure}{0.14\textwidth}
            \centering 
            \caption*{PGN}
            \includegraphics[width=\linewidth]{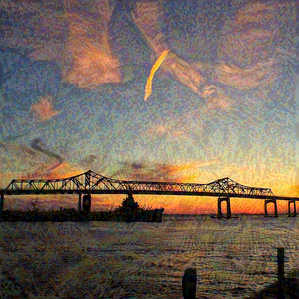}
            \vspace{-0.5cm}
        \end{subfigure}       

        \vspace{3mm}
        \begin{subfigure}{0.14\textwidth}
            \centering 
            \includegraphics[width=\linewidth]{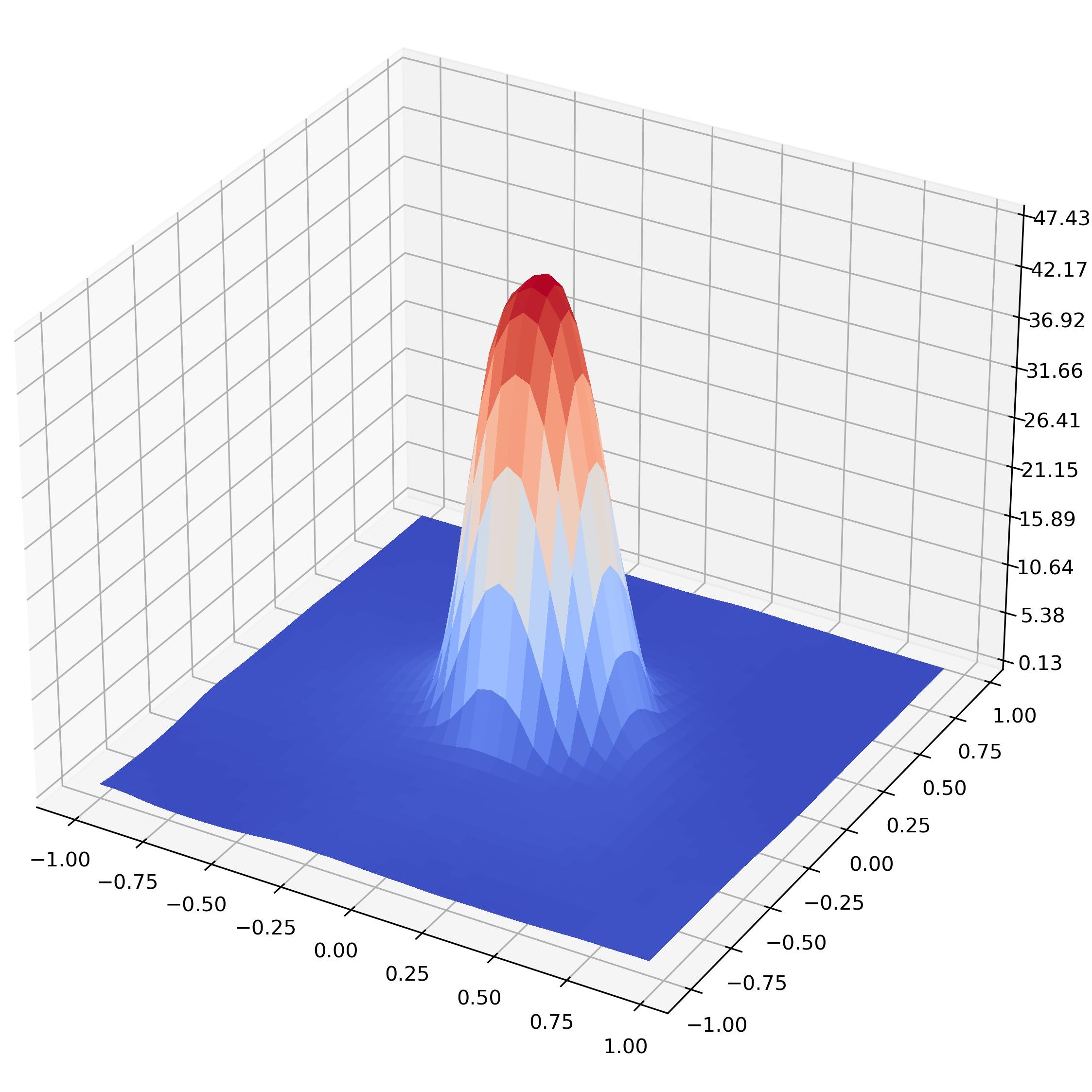}
            \vspace{-0.5cm}
        \end{subfigure}%
        \hspace{3mm}
        \begin{subfigure}{0.14\textwidth} 
            \centering 
            \includegraphics[width=\linewidth]{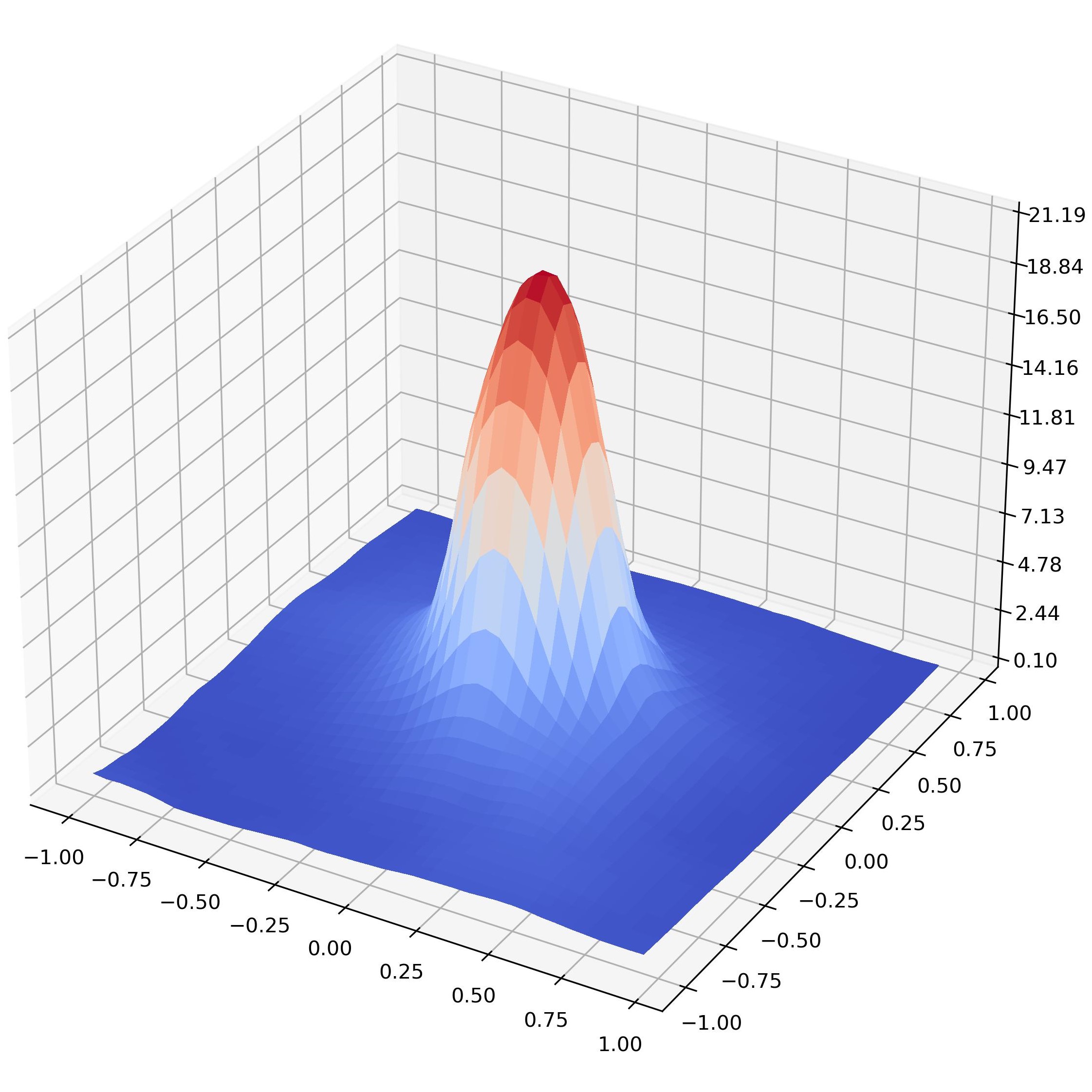}
            \vspace{-0.5cm} 
        \end{subfigure}%
        \hspace{3mm}
        \begin{subfigure}{0.14\textwidth}
            \centering 
            \includegraphics[width=\linewidth]{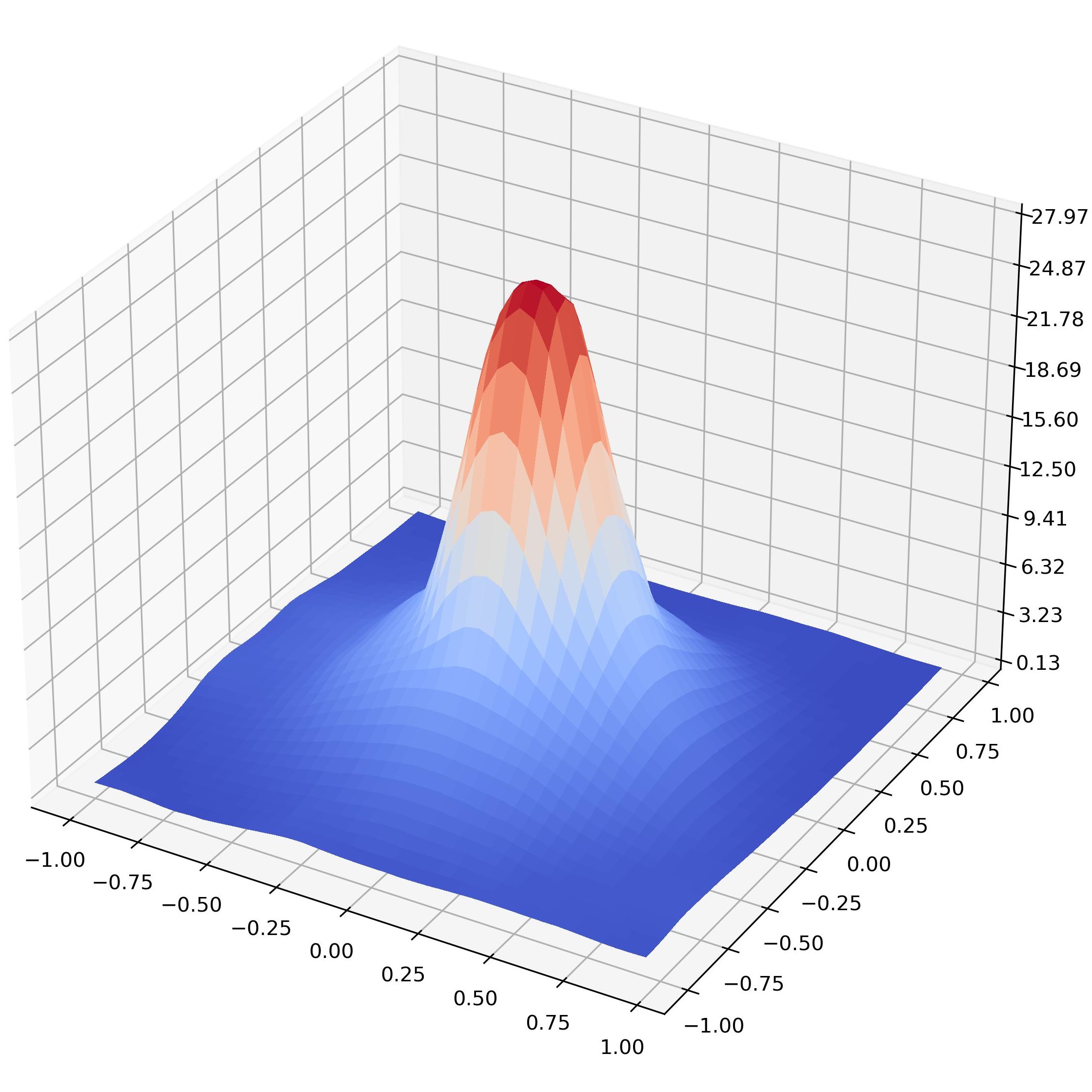}
            \vspace{-0.5cm}
            
        \end{subfigure}
        \hspace{3mm}
        \begin{subfigure}{0.14\textwidth}
            \centering 
            \includegraphics[width=\linewidth]{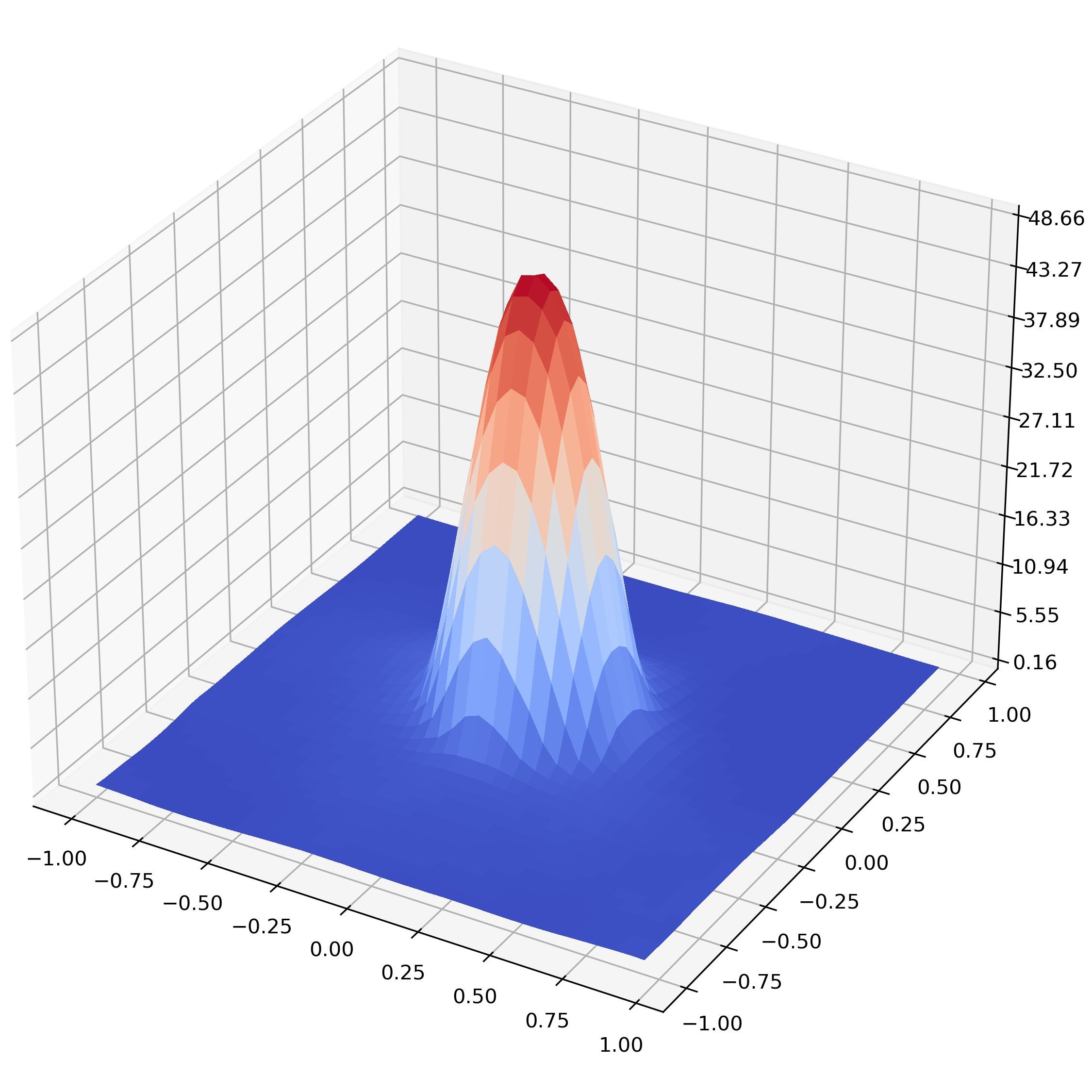}
            \vspace{-0.5cm}
            
        \end{subfigure}%
        \hspace{3mm}
        \begin{subfigure}{0.14\textwidth} 
            \centering 
            \includegraphics[width=\linewidth]{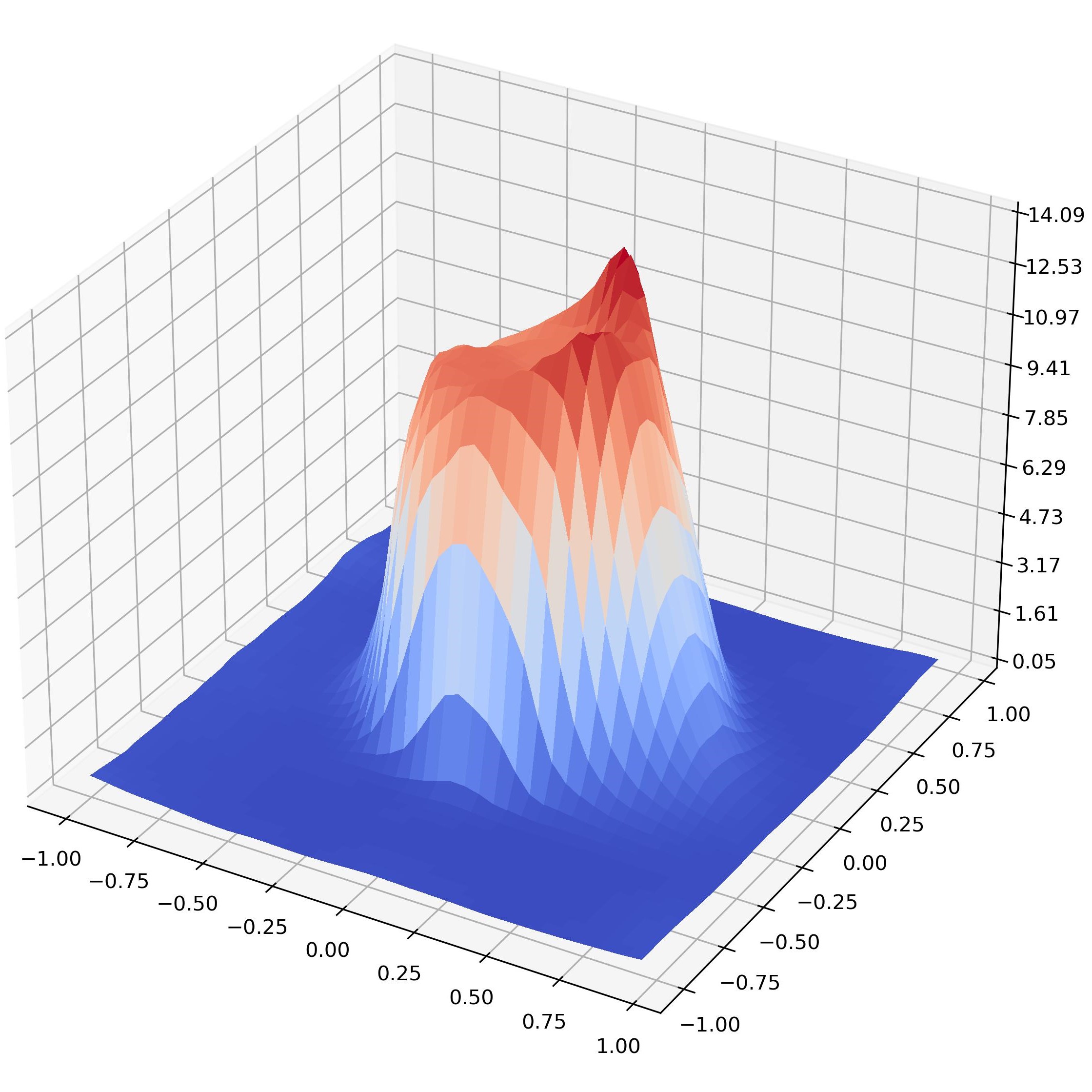}
            \vspace{-0.5cm}
            
        \end{subfigure}%
        \hspace{3mm}
        \begin{subfigure}{0.14\textwidth}
            \centering 
            \includegraphics[width=\linewidth]{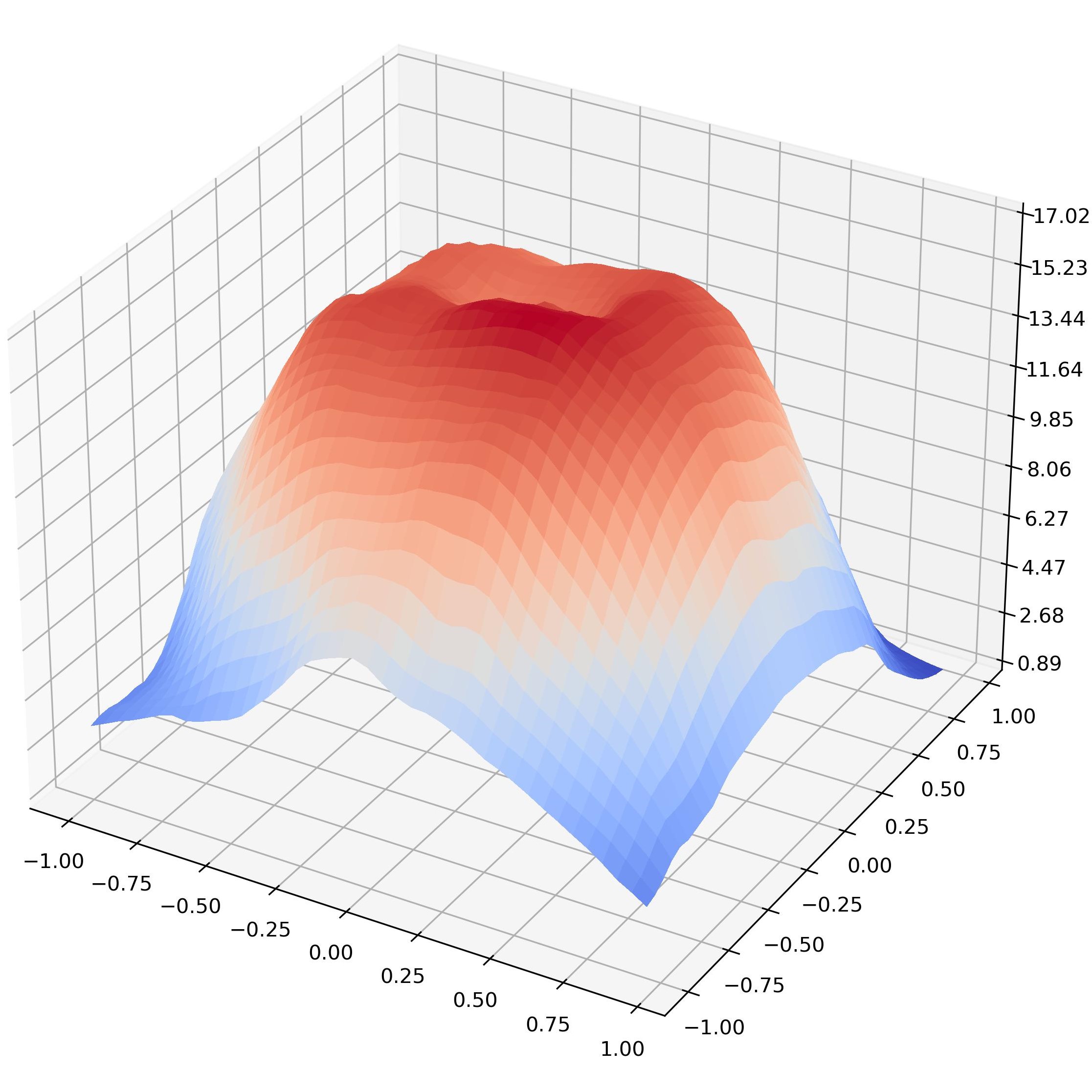}
            \vspace{-0.5cm}
        \end{subfigure}
    \end{minipage}

    \vspace{0.5cm}
    \centering
    \begin{minipage}{0.12\textwidth}
        \begin{subfigure}{\textwidth}
        \centering
            \includegraphics[width=\linewidth]{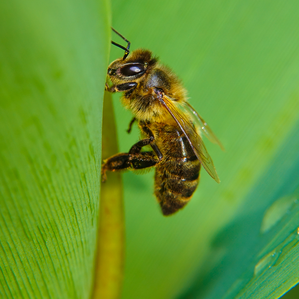}
            \vspace{-0.5cm}
        \end{subfigure}
    \end{minipage}
    \hspace{1mm}
    \begin{minipage}{0.86\textwidth}
        \begin{subfigure}{0.14\textwidth}
            \centering 
            \includegraphics[width=\linewidth]{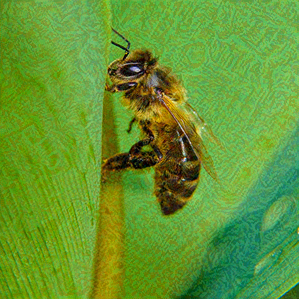}
            \vspace{-0.5cm}
        \end{subfigure}%
        \hspace{3mm}
        \begin{subfigure}{0.14\textwidth} 
            \centering 
            \includegraphics[width=\linewidth]{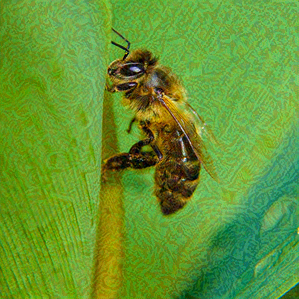}
            \vspace{-0.5cm} 
        \end{subfigure}%
        \hspace{3mm}
        \begin{subfigure}{0.14\textwidth}
            \centering 
            \includegraphics[width=\linewidth]{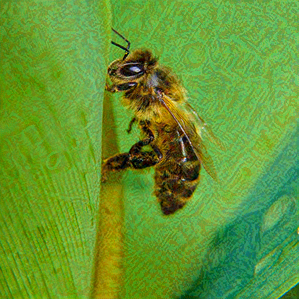}
            \vspace{-0.5cm}
            
        \end{subfigure}
        \hspace{3mm}
        \begin{subfigure}{0.14\textwidth}
            \centering 
            \includegraphics[width=\linewidth]{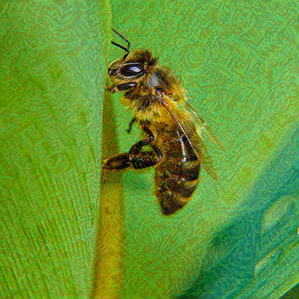}
            \vspace{-0.5cm}
            
        \end{subfigure}%
        \hspace{3mm}
        \begin{subfigure}{0.14\textwidth} 
            \centering 
            \includegraphics[width=\linewidth]{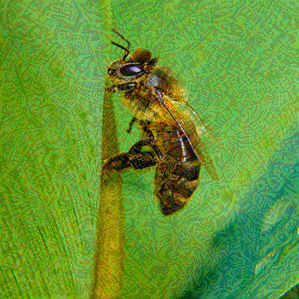}
            \vspace{-0.5cm}
            
        \end{subfigure}%
        \hspace{3mm}
        \begin{subfigure}{0.14\textwidth}
            \centering 
            \includegraphics[width=\linewidth]{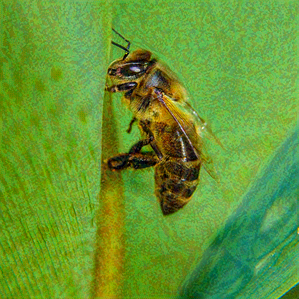}
            \vspace{-0.5cm}
        \end{subfigure}       

        \vspace{3mm}
        \begin{subfigure}{0.14\textwidth}
            \centering 
            \includegraphics[width=\linewidth]{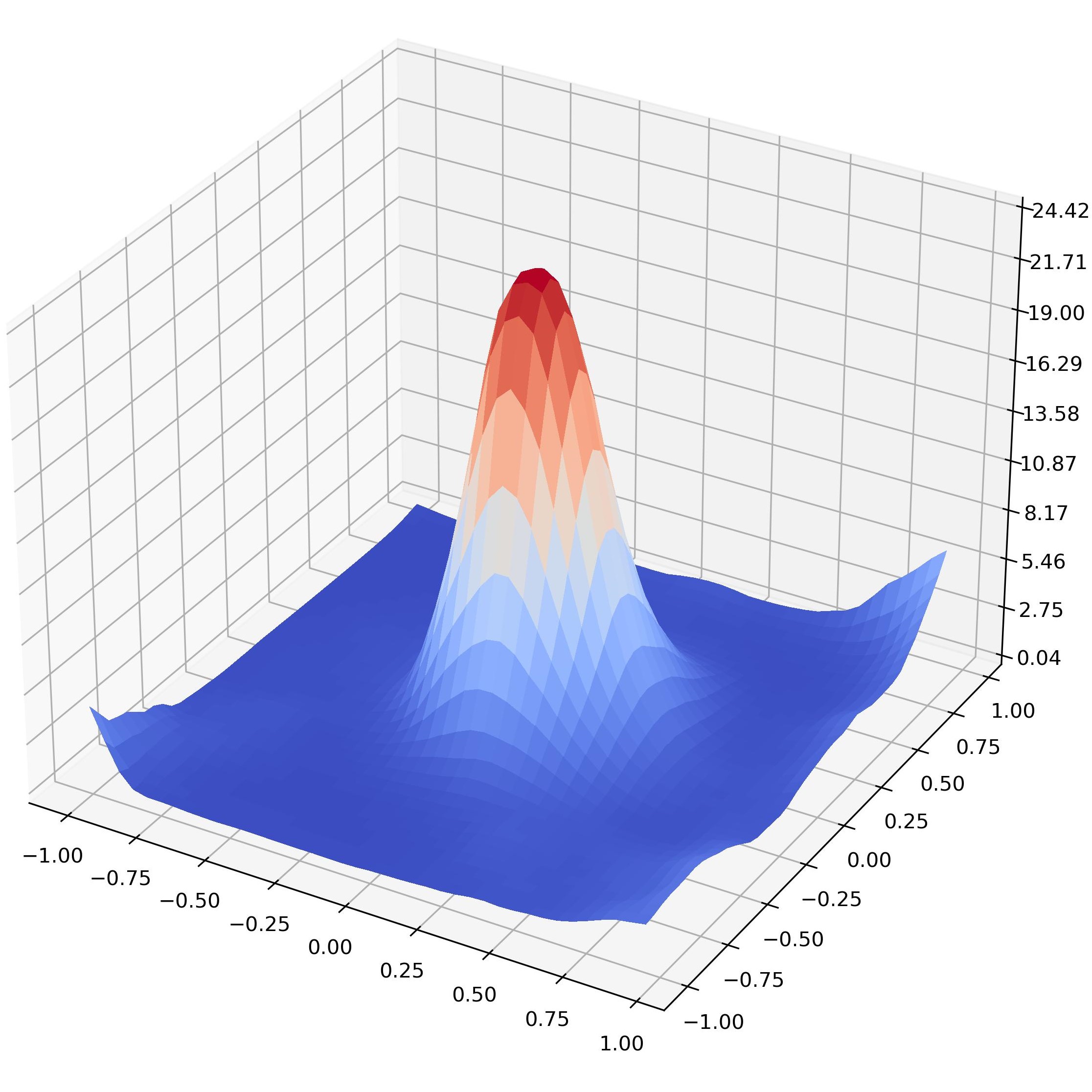}
            \vspace{-0.5cm}
        \end{subfigure}%
        \hspace{3mm}
        \begin{subfigure}{0.14\textwidth} 
            \centering 
            \includegraphics[width=\linewidth]{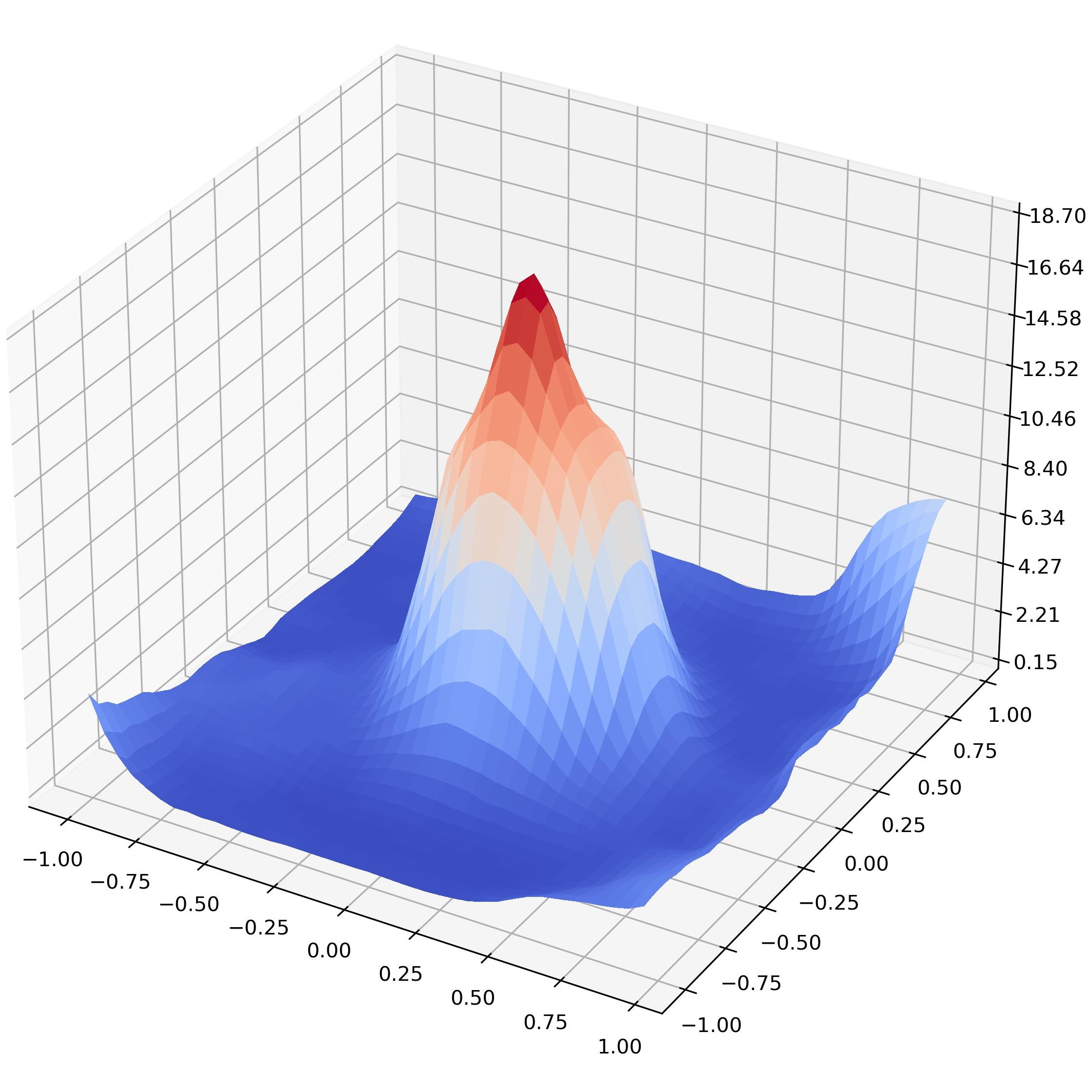}
            \vspace{-0.5cm} 
        \end{subfigure}%
        \hspace{3mm}
        \begin{subfigure}{0.14\textwidth}
            \centering 
            \includegraphics[width=\linewidth]{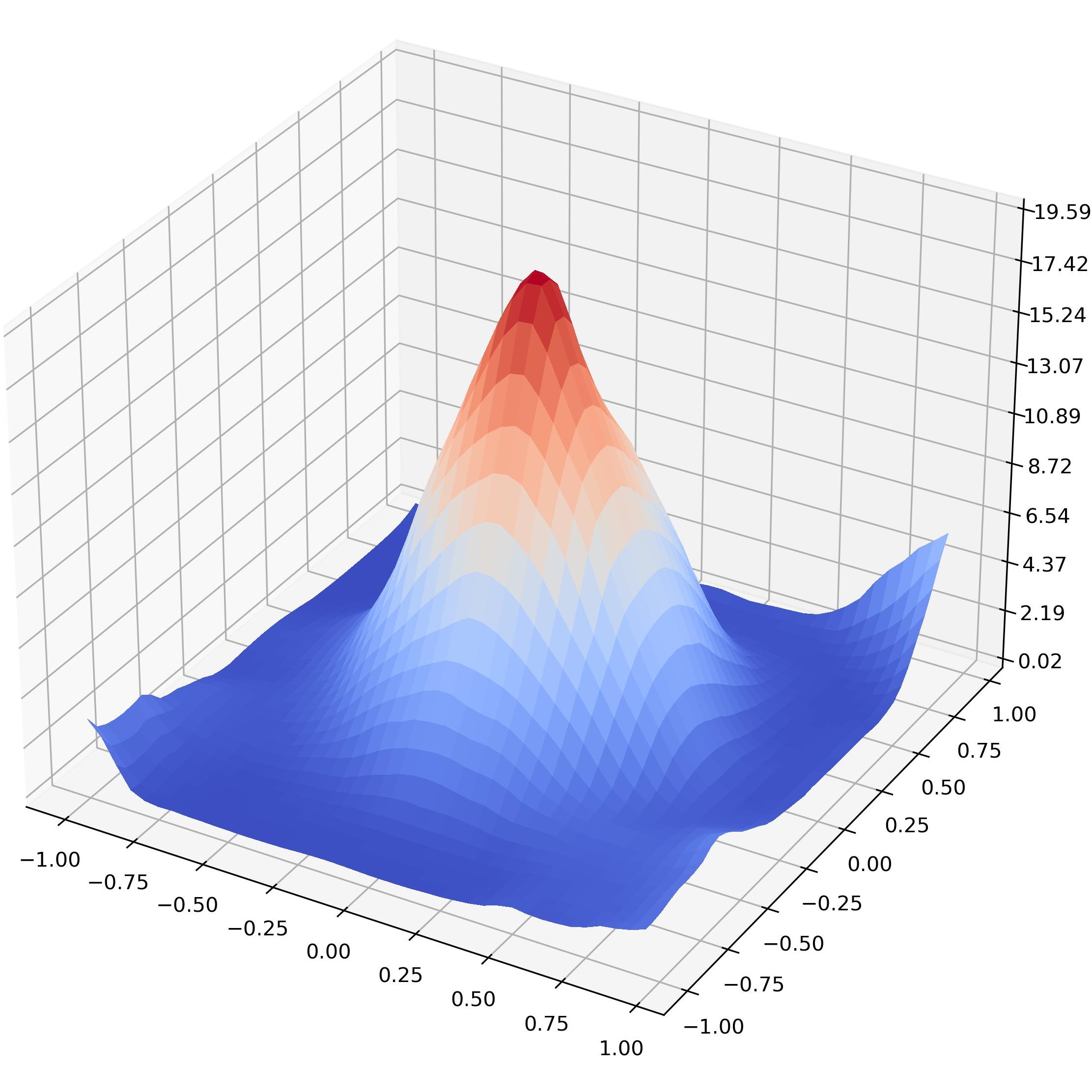}
            \vspace{-0.5cm}
            
        \end{subfigure}
        \hspace{3mm}
        \begin{subfigure}{0.14\textwidth}
            \centering 
            \includegraphics[width=\linewidth]{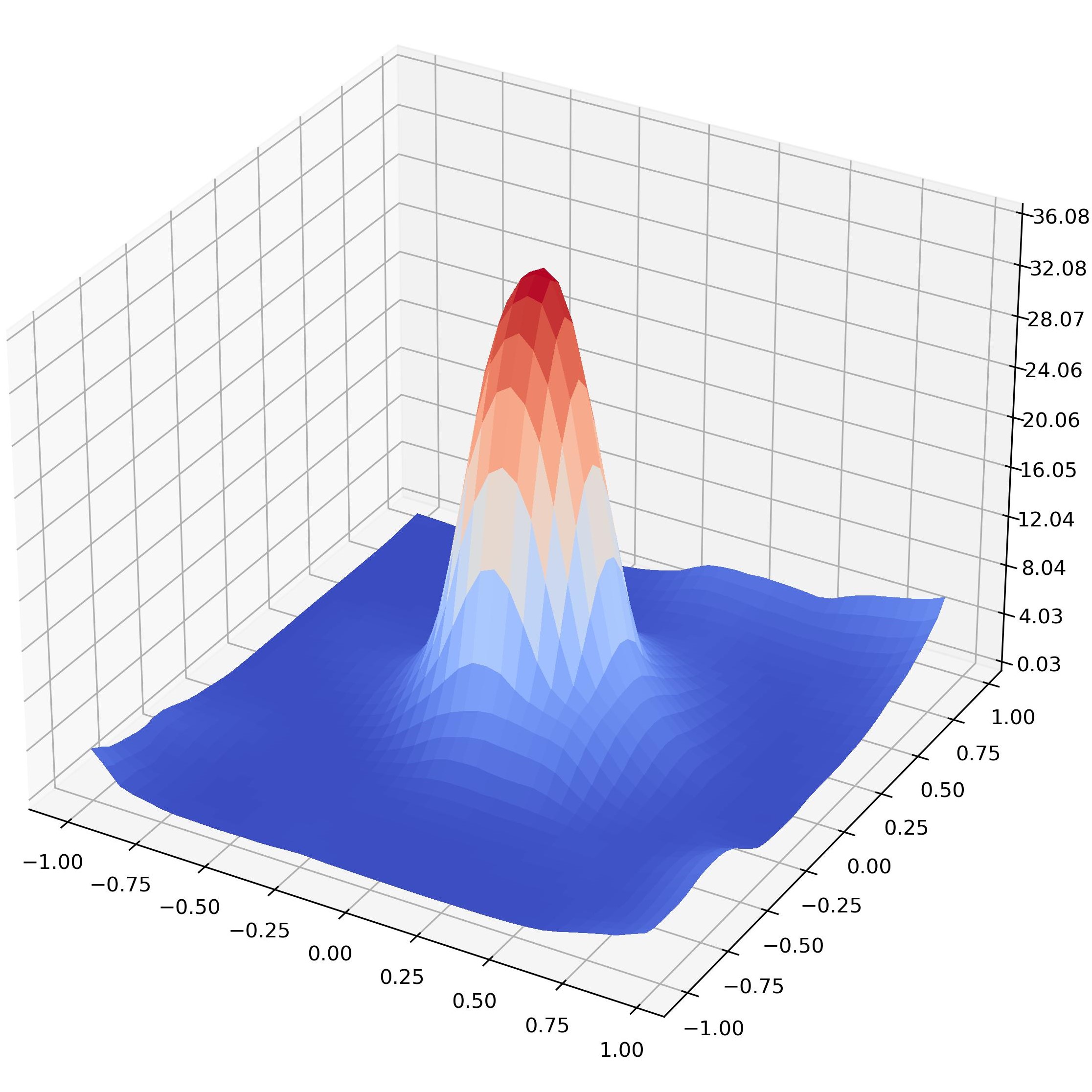}
            \vspace{-0.5cm}
            
        \end{subfigure}%
        \hspace{3mm}
        \begin{subfigure}{0.14\textwidth} 
            \centering 
            \includegraphics[width=\linewidth]{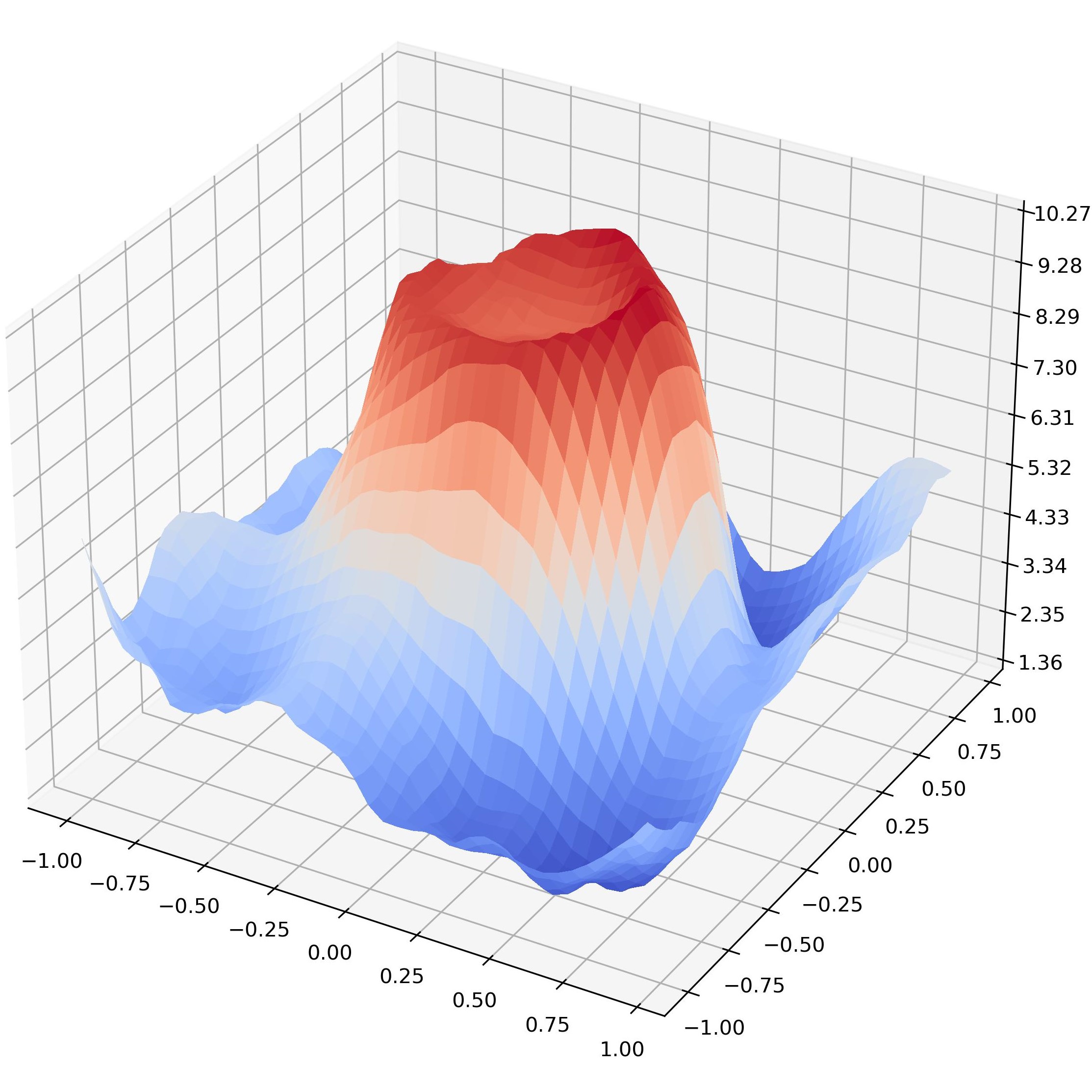}
            \vspace{-0.5cm}
            
        \end{subfigure}%
        \hspace{3mm}
        \begin{subfigure}{0.14\textwidth}
            \centering 
            \includegraphics[width=\linewidth]{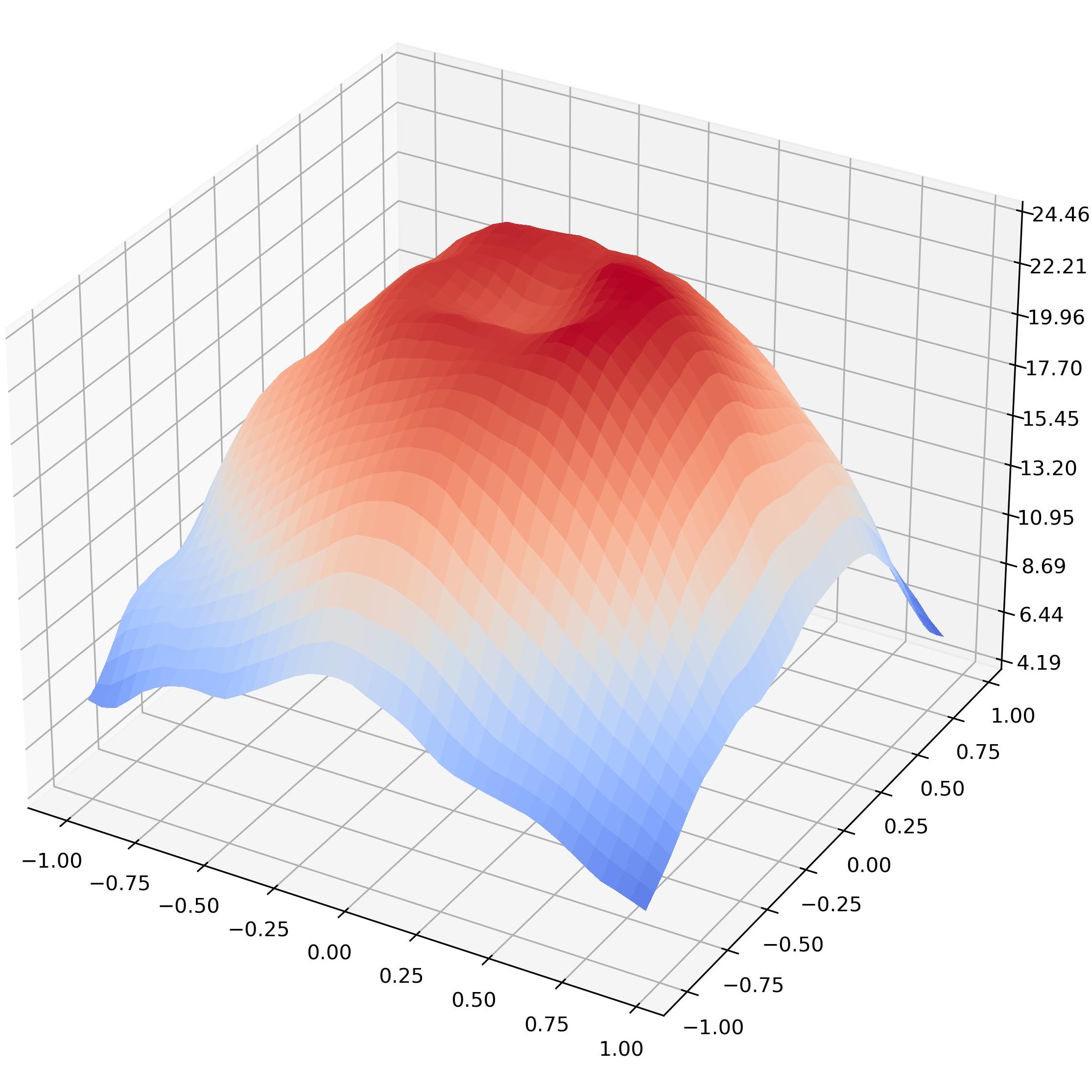}
            \vspace{-0.5cm}
        \end{subfigure}
    \end{minipage}

    \vspace{0.5cm}
    \centering
    \begin{minipage}{0.12\textwidth}
        \begin{subfigure}{\textwidth}
        \centering
            \includegraphics[width=\linewidth]{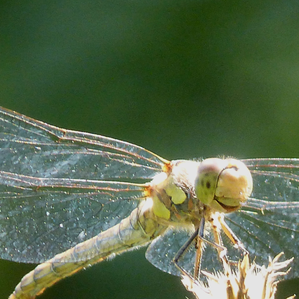}
            \vspace{-0.5cm}
        \end{subfigure}
    \end{minipage}
    \hspace{1mm}
    \begin{minipage}{0.86\textwidth}
        \begin{subfigure}{0.14\textwidth}
            \centering 
            \includegraphics[width=\linewidth]{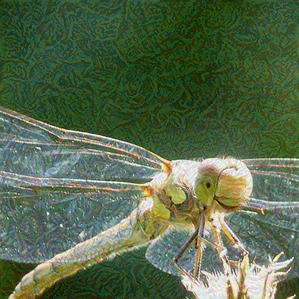}
            \vspace{-0.5cm}
        \end{subfigure}%
        \hspace{3mm}
        \begin{subfigure}{0.14\textwidth} 
            \centering 
            \includegraphics[width=\linewidth]{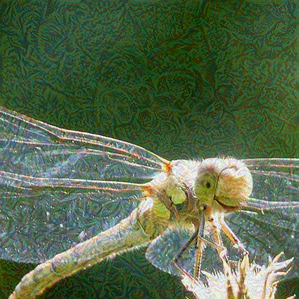}
            \vspace{-0.5cm} 
        \end{subfigure}%
        \hspace{3mm}
        \begin{subfigure}{0.14\textwidth}
            \centering 
            \includegraphics[width=\linewidth]{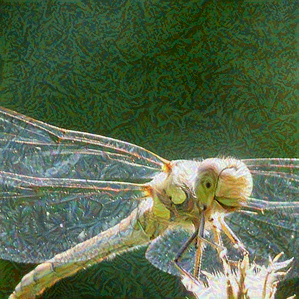}
            \vspace{-0.5cm}
            
        \end{subfigure}
        \hspace{3mm}
        \begin{subfigure}{0.14\textwidth}
            \centering 
            \includegraphics[width=\linewidth]{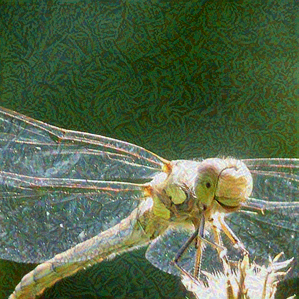}
            \vspace{-0.5cm}
            
        \end{subfigure}%
        \hspace{3mm}
        \begin{subfigure}{0.14\textwidth} 
            \centering 
            \includegraphics[width=\linewidth]{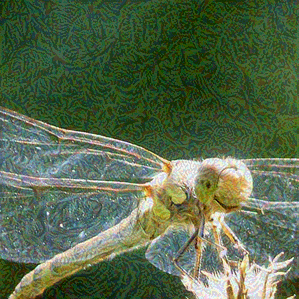}
            \vspace{-0.5cm}
            
        \end{subfigure}%
        \hspace{3mm}
        \begin{subfigure}{0.14\textwidth}
            \centering 
            \includegraphics[width=\linewidth]{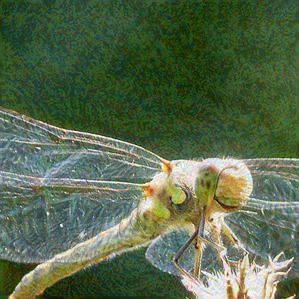}
            \vspace{-0.5cm}
        \end{subfigure}    
        
        \vspace{3mm}
        \begin{subfigure}{0.14\textwidth}
            \centering 
            \includegraphics[width=\linewidth]{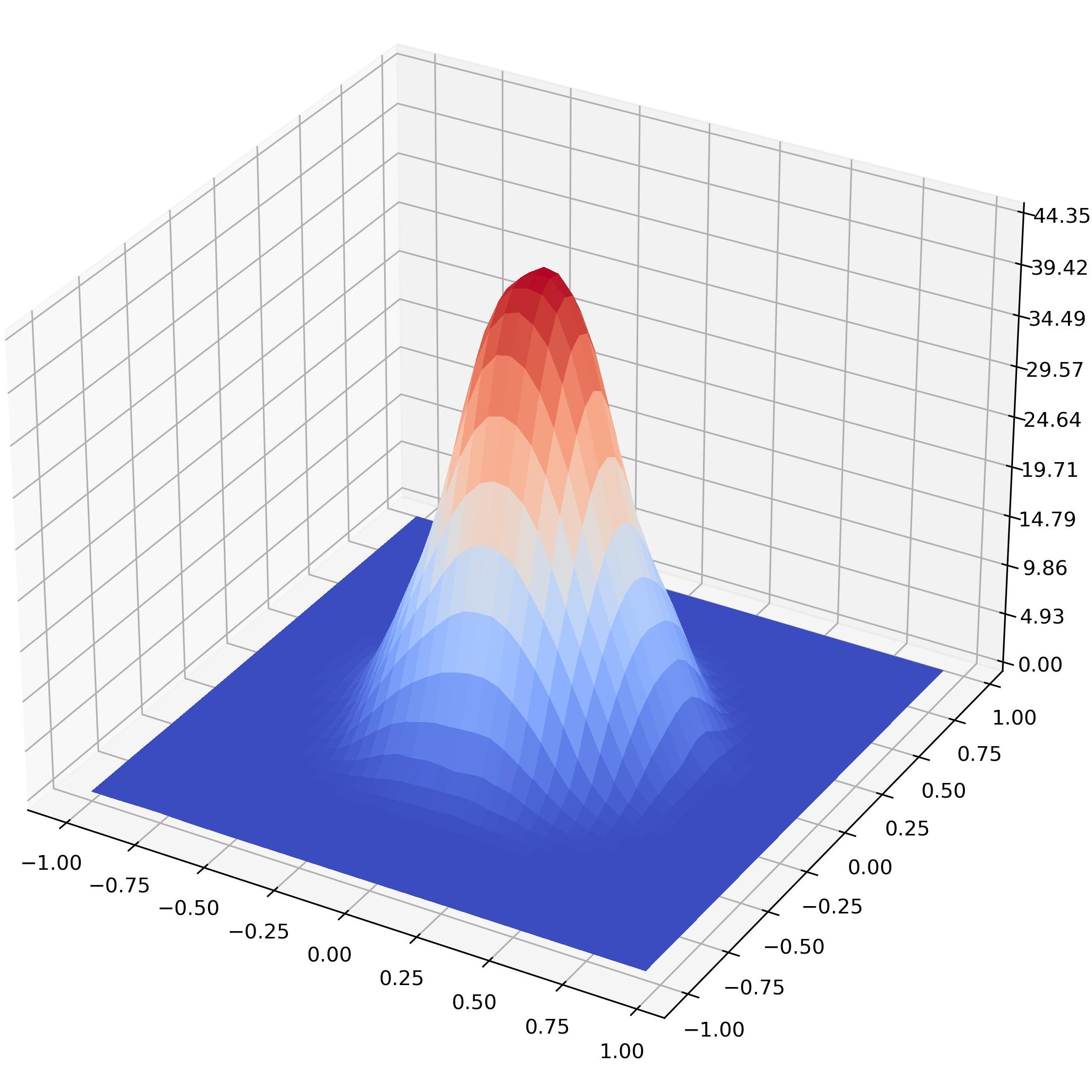}
            \vspace{-0.5cm}
        \end{subfigure}%
        \hspace{3mm}
        \begin{subfigure}{0.14\textwidth} 
            \centering 
            \includegraphics[width=\linewidth]{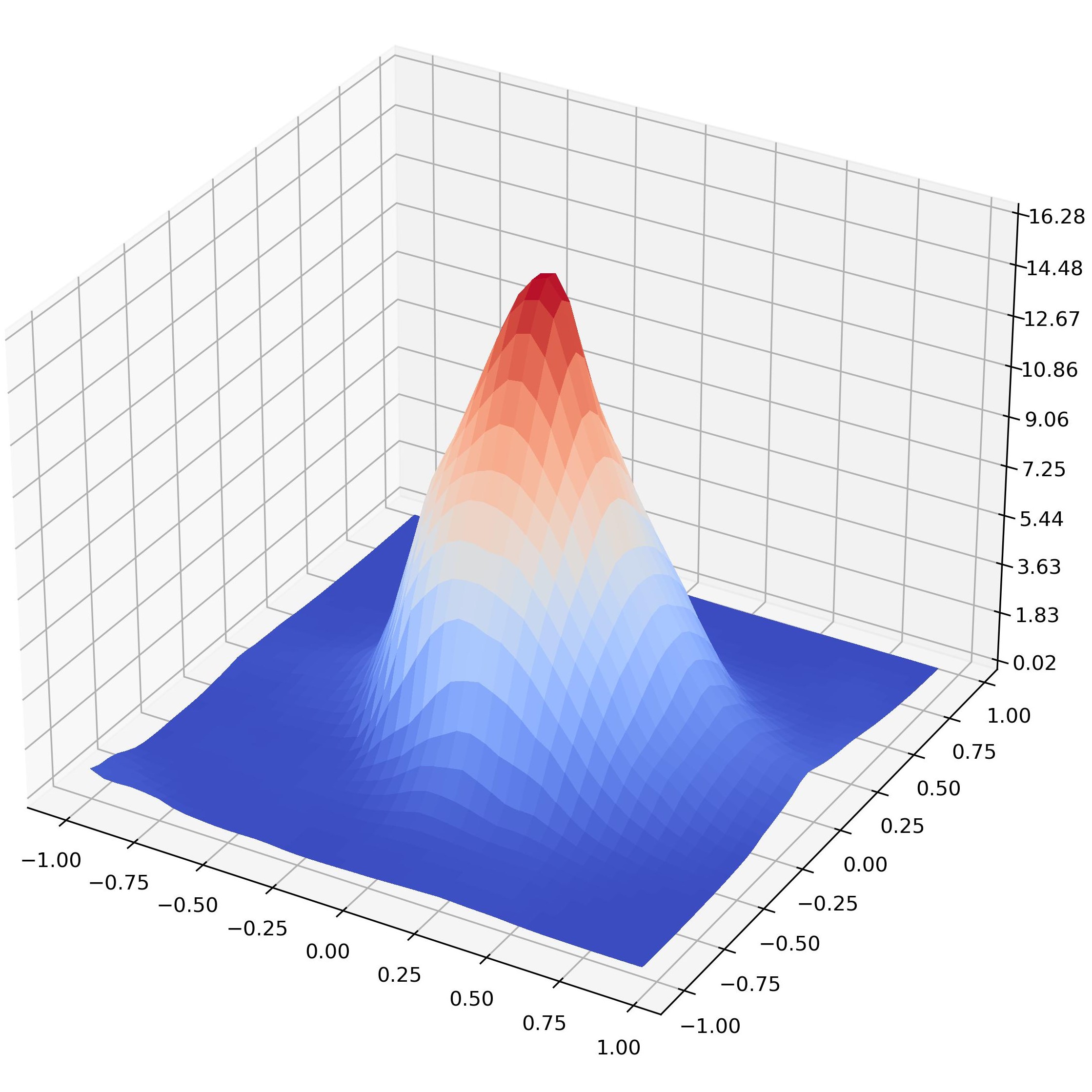}
            \vspace{-0.5cm} 
        \end{subfigure}%
        \hspace{3mm}
        \begin{subfigure}{0.14\textwidth}
            \centering 
            \includegraphics[width=\linewidth]{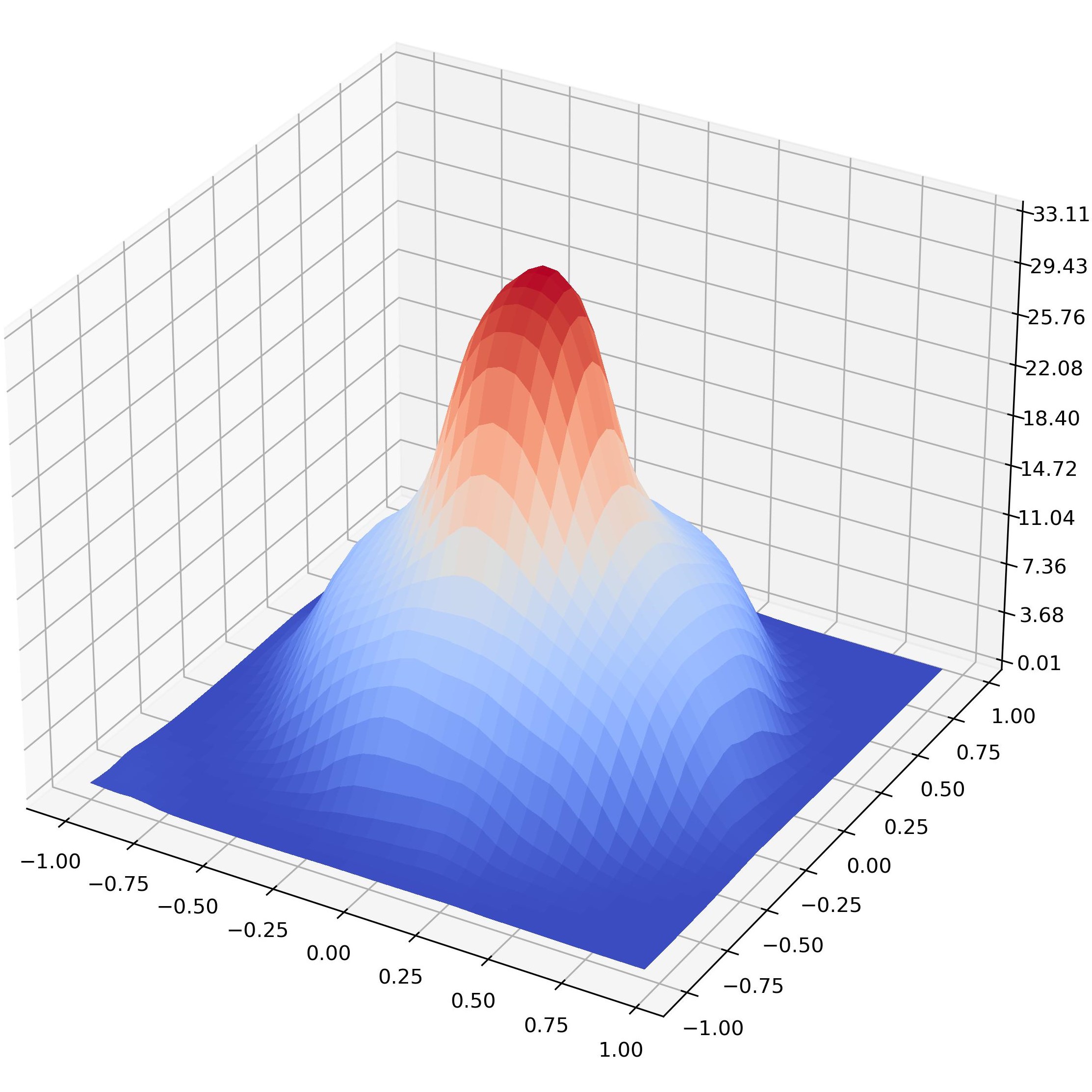}
            \vspace{-0.5cm}
            
        \end{subfigure}
        \hspace{3mm}
        \begin{subfigure}{0.14\textwidth}
            \centering 
            \includegraphics[width=\linewidth]{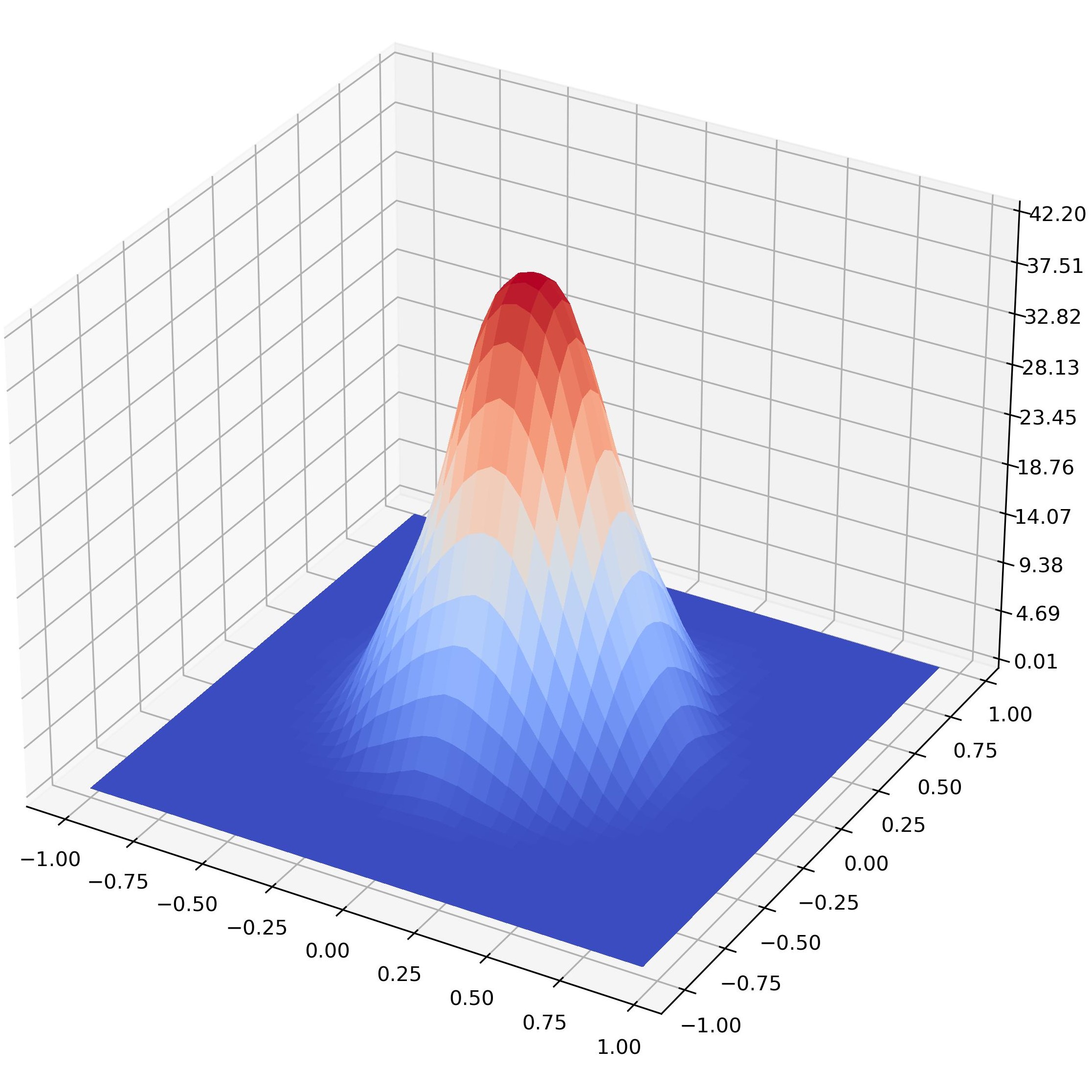}
            \vspace{-0.5cm}
            
        \end{subfigure}%
        \hspace{3mm}
        \begin{subfigure}{0.14\textwidth} 
            \centering 
            \includegraphics[width=\linewidth]{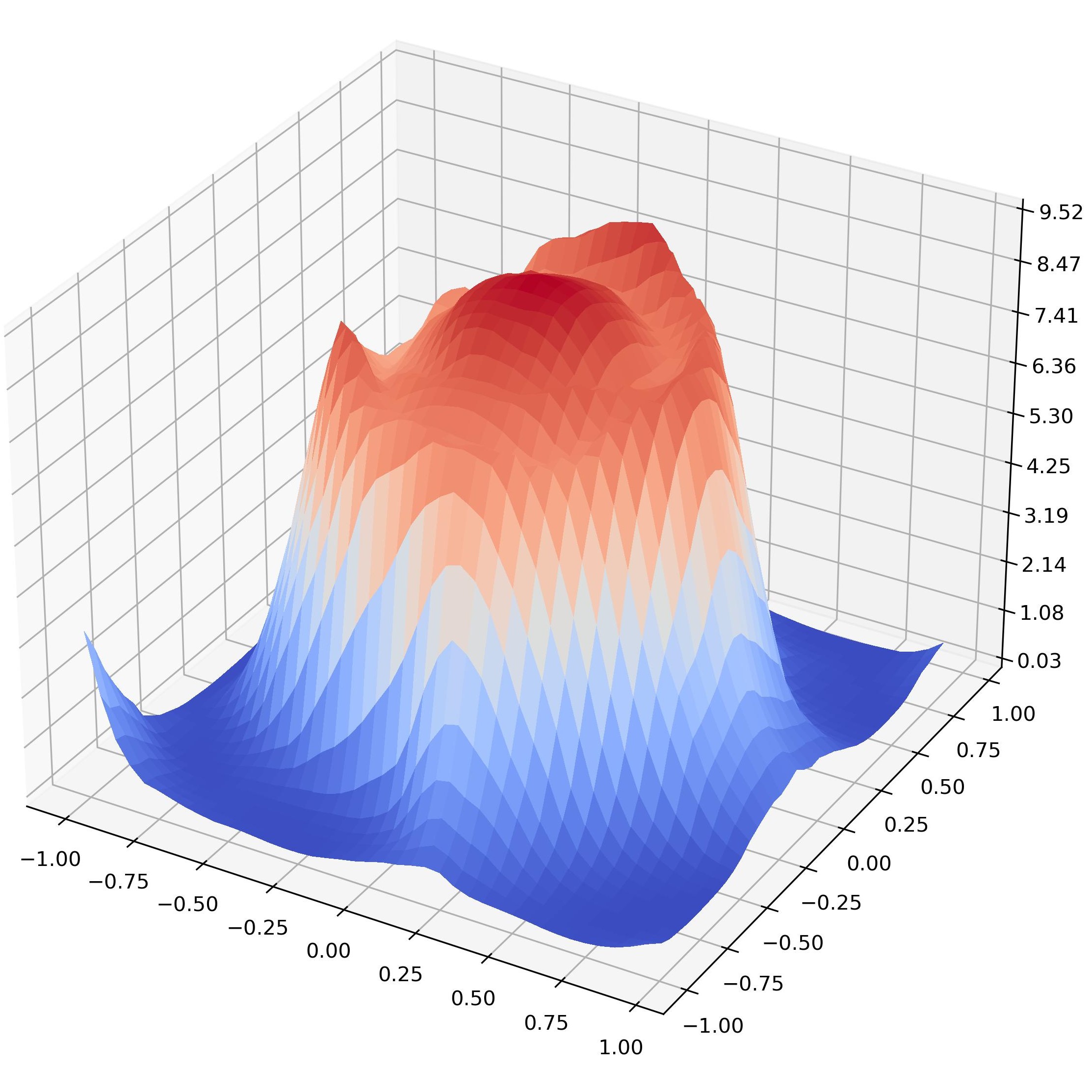}
            \vspace{-0.5cm}
            
        \end{subfigure}%
        \hspace{3mm}
        \begin{subfigure}{0.14\textwidth}
            \centering 
            \includegraphics[width=\linewidth]{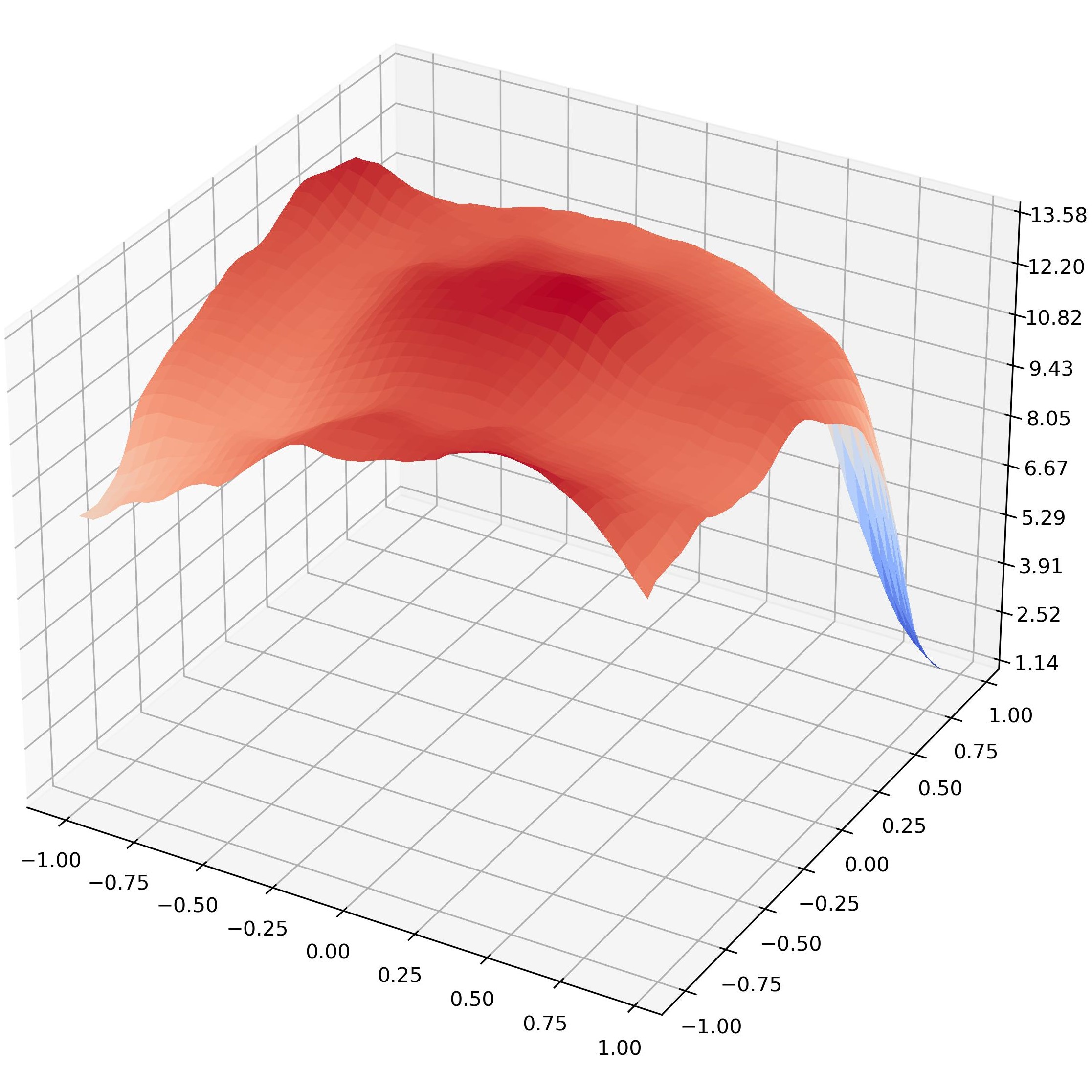}
            \vspace{-0.5cm}
        \end{subfigure}
    \end{minipage}

    \vspace{0.5cm}
    \centering
    \begin{minipage}{0.12\textwidth}
        \begin{subfigure}{\textwidth}
        \centering
            \includegraphics[width=\linewidth]{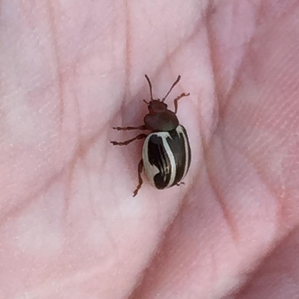}
            \vspace{-0.5cm}
        \end{subfigure}
    \end{minipage}
    \hspace{1mm}
    \begin{minipage}{0.86\textwidth}
        \begin{subfigure}{0.14\textwidth}
            \centering 
            \includegraphics[width=\linewidth]{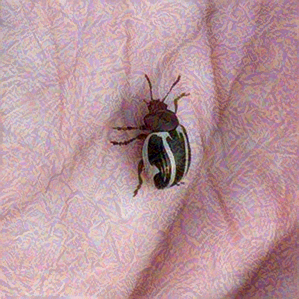}
            \vspace{-0.5cm}
        \end{subfigure}%
        \hspace{3mm}
        \begin{subfigure}{0.14\textwidth} 
            \centering 
            \includegraphics[width=\linewidth]{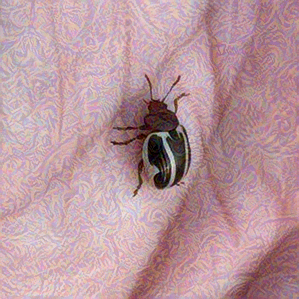}
            \vspace{-0.5cm} 
        \end{subfigure}%
        \hspace{3mm}
        \begin{subfigure}{0.14\textwidth}
            \centering 
            \includegraphics[width=\linewidth]{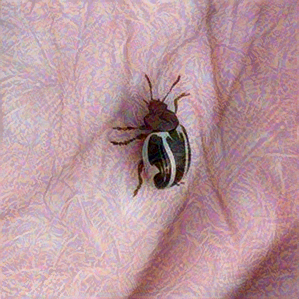}
            \vspace{-0.5cm}
            
        \end{subfigure}
        \hspace{3mm}
        \begin{subfigure}{0.14\textwidth}
            \centering 
            \includegraphics[width=\linewidth]{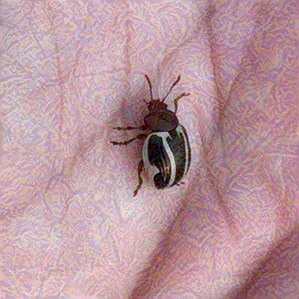}
            \vspace{-0.5cm}
            
        \end{subfigure}%
        \hspace{3mm}
        \begin{subfigure}{0.14\textwidth} 
            \centering 
            \includegraphics[width=\linewidth]{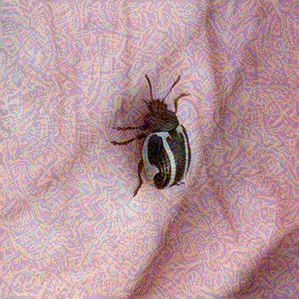}
            \vspace{-0.5cm}
            
        \end{subfigure}%
        \hspace{3mm}
        \begin{subfigure}{0.14\textwidth}
            \centering 
            \includegraphics[width=\linewidth]{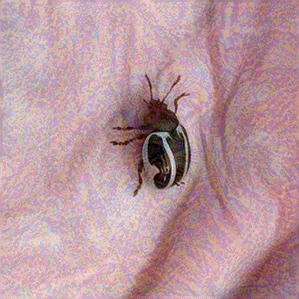}
            \vspace{-0.5cm}
        \end{subfigure}       

        \vspace{3mm}
        \begin{subfigure}{0.14\textwidth}
            \centering 
            \includegraphics[width=\linewidth]{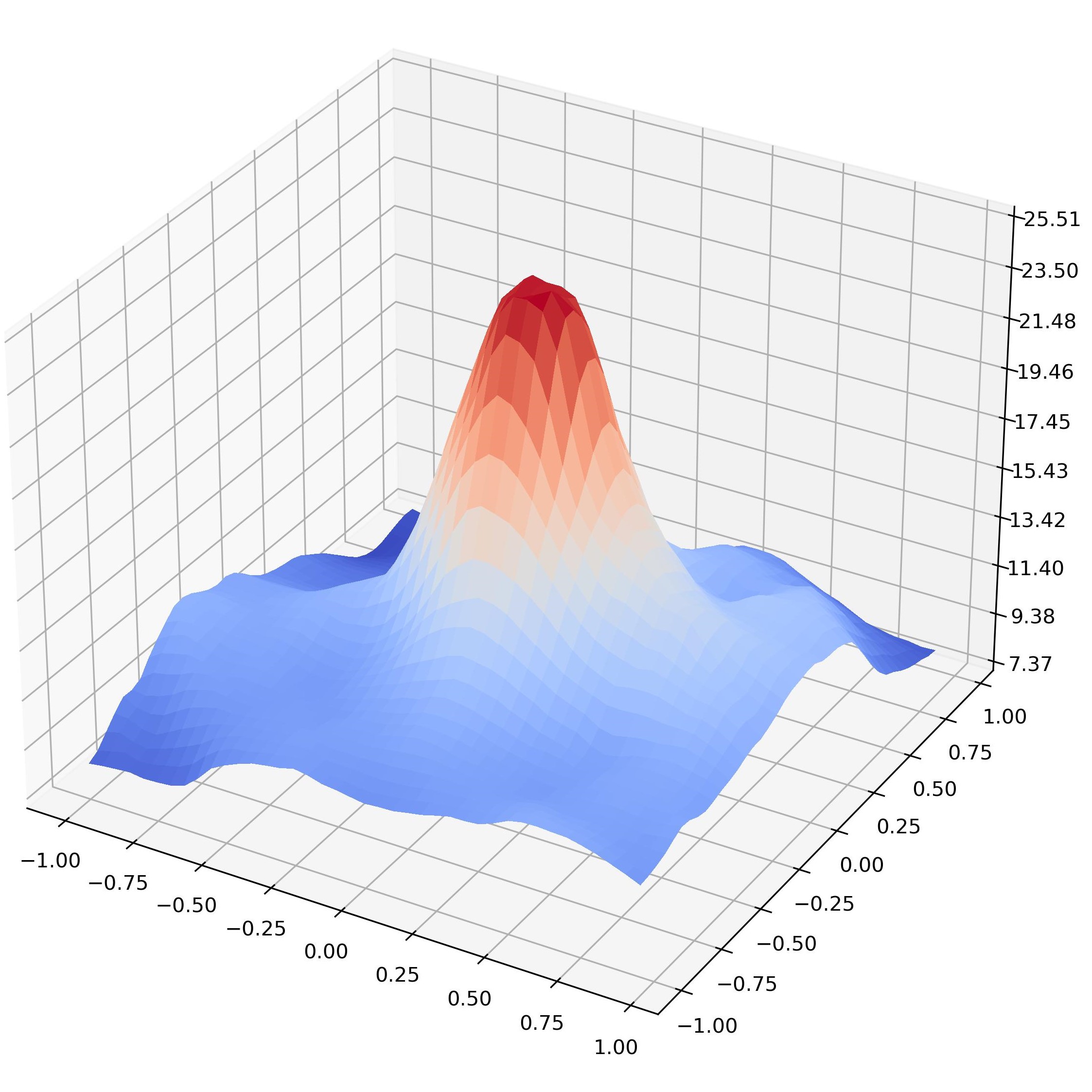}
            \vspace{-0.5cm}
        \end{subfigure}%
        \hspace{3mm}
        \begin{subfigure}{0.14\textwidth} 
            \centering 
            \includegraphics[width=\linewidth]{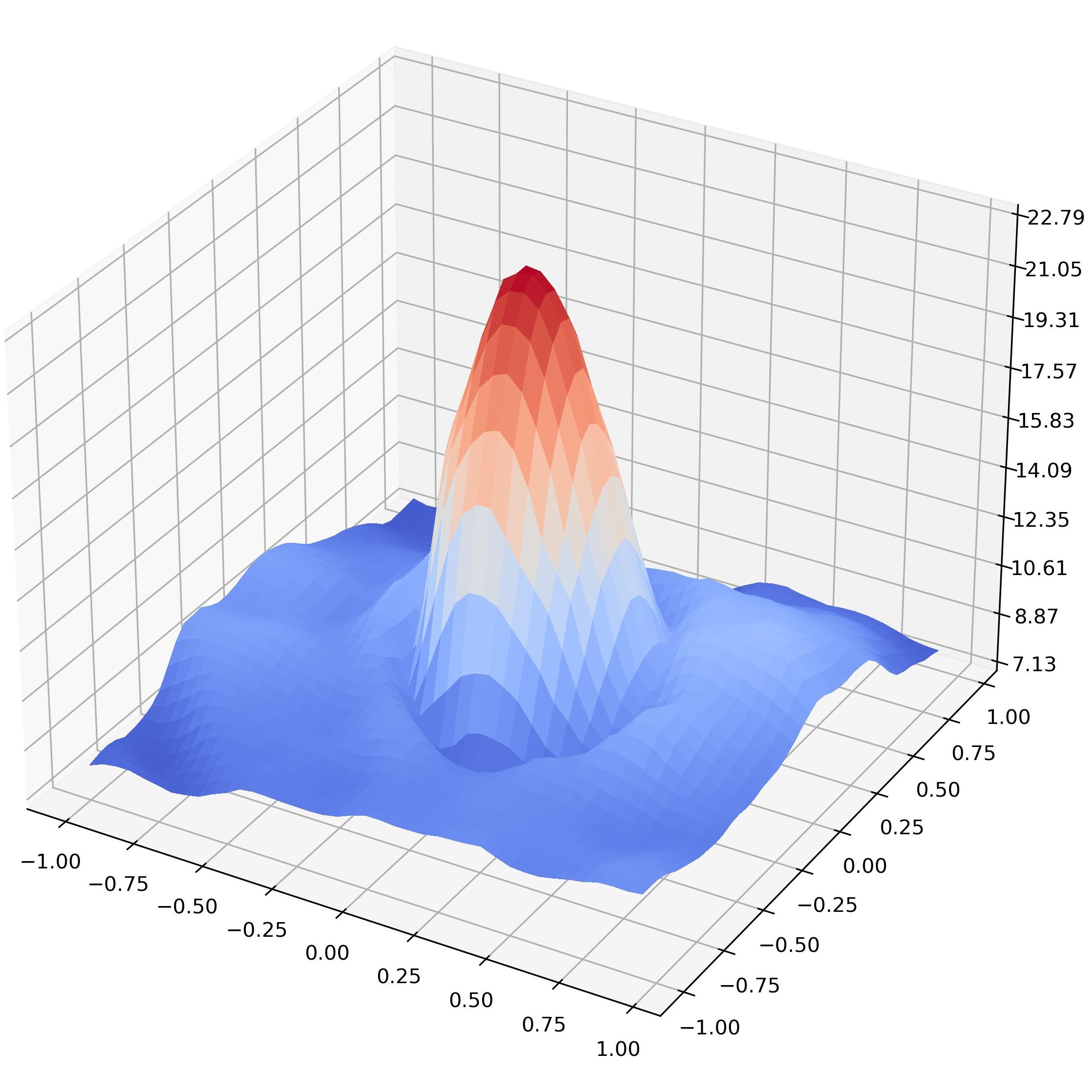}
            \vspace{-0.5cm} 
        \end{subfigure}%
        \hspace{3mm}
        \begin{subfigure}{0.14\textwidth}
            \centering 
            \includegraphics[width=\linewidth]{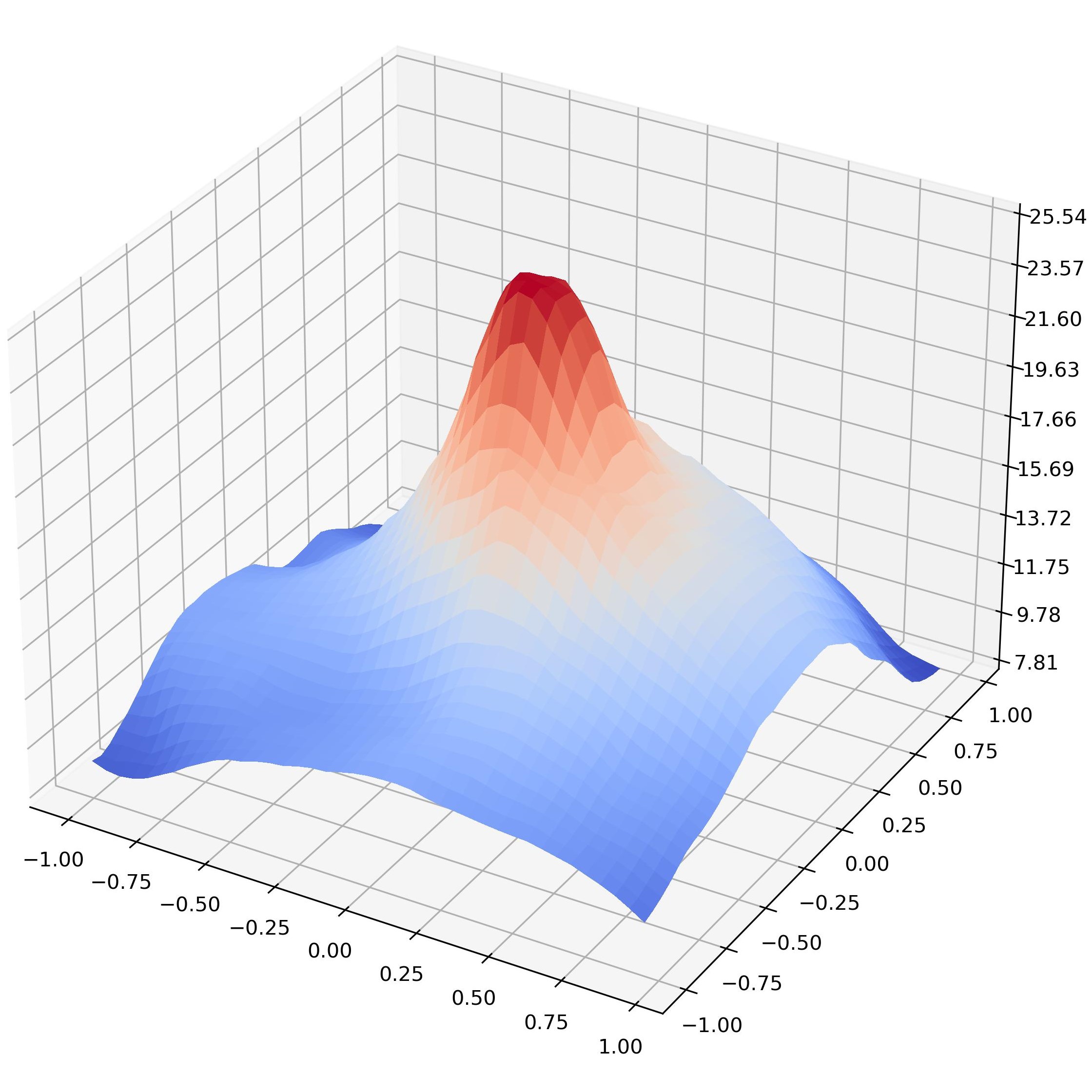}
            \vspace{-0.5cm}
            
        \end{subfigure}
        \hspace{3mm}
        \begin{subfigure}{0.14\textwidth}
            \centering 
            \includegraphics[width=\linewidth]{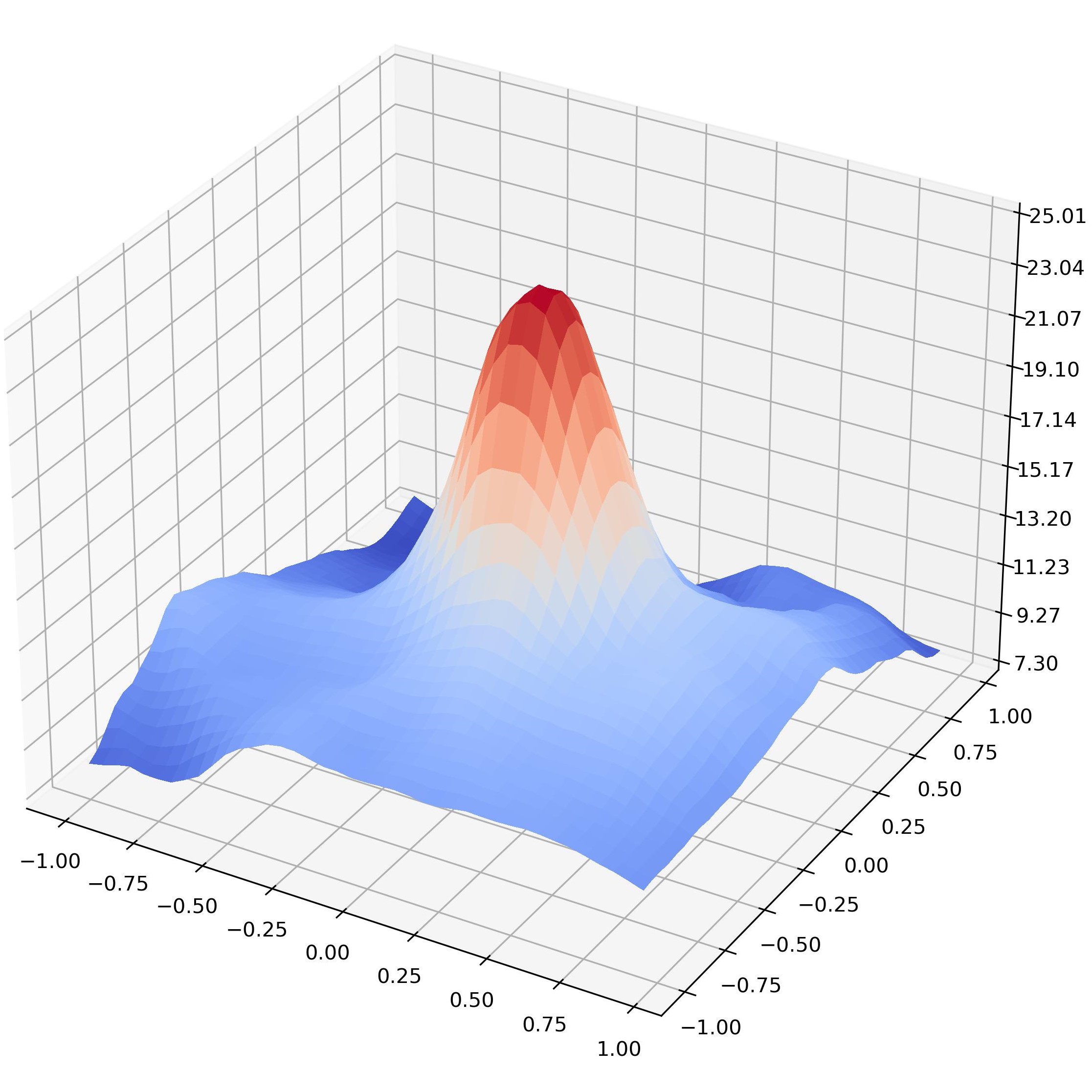}
            \vspace{-0.5cm}
            
        \end{subfigure}%
        \hspace{3mm}
        \begin{subfigure}{0.14\textwidth} 
            \centering 
            \includegraphics[width=\linewidth]{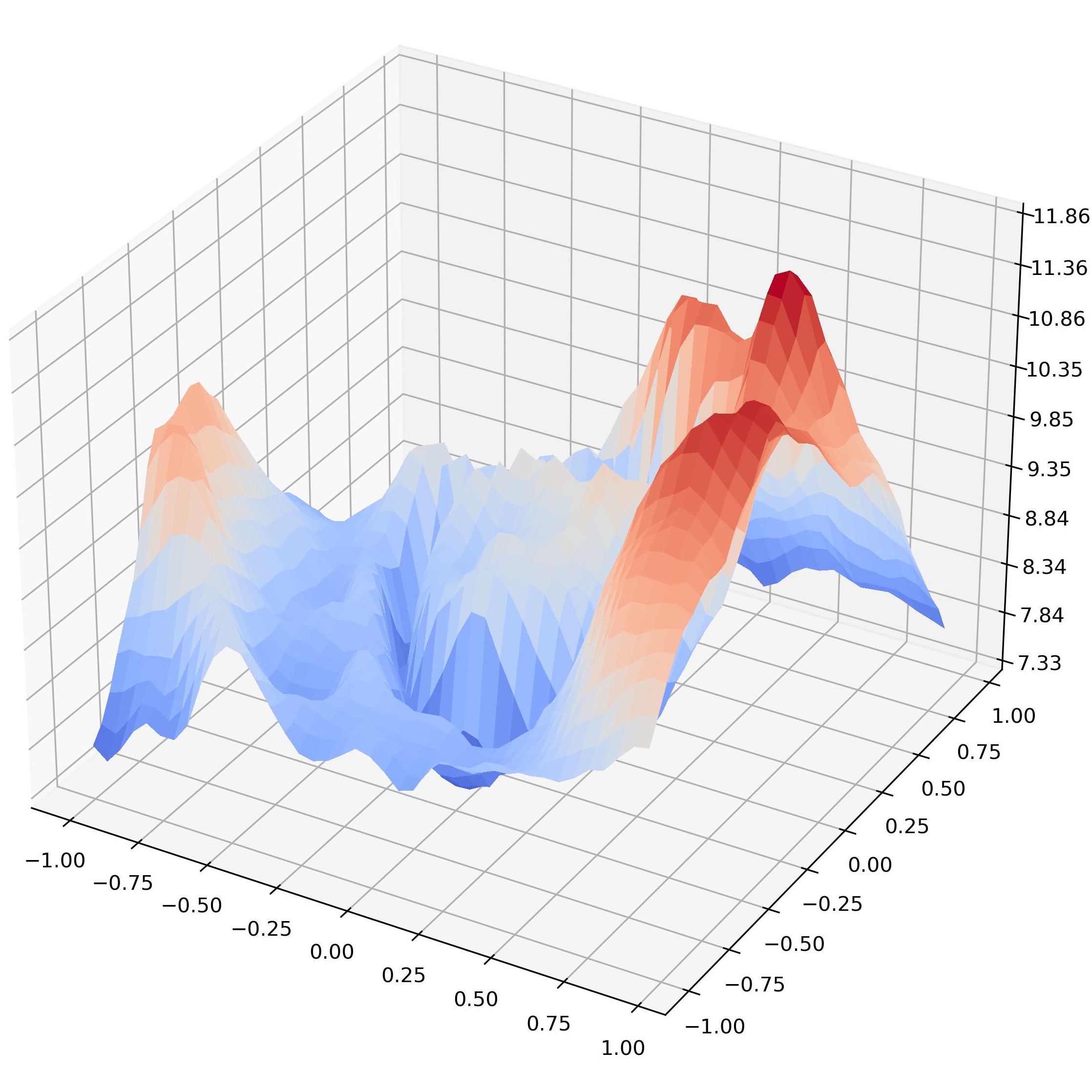}
            \vspace{-0.5cm}
            
        \end{subfigure}%
        \hspace{3mm}
        \begin{subfigure}{0.14\textwidth}
            \centering 
            \includegraphics[width=\linewidth]{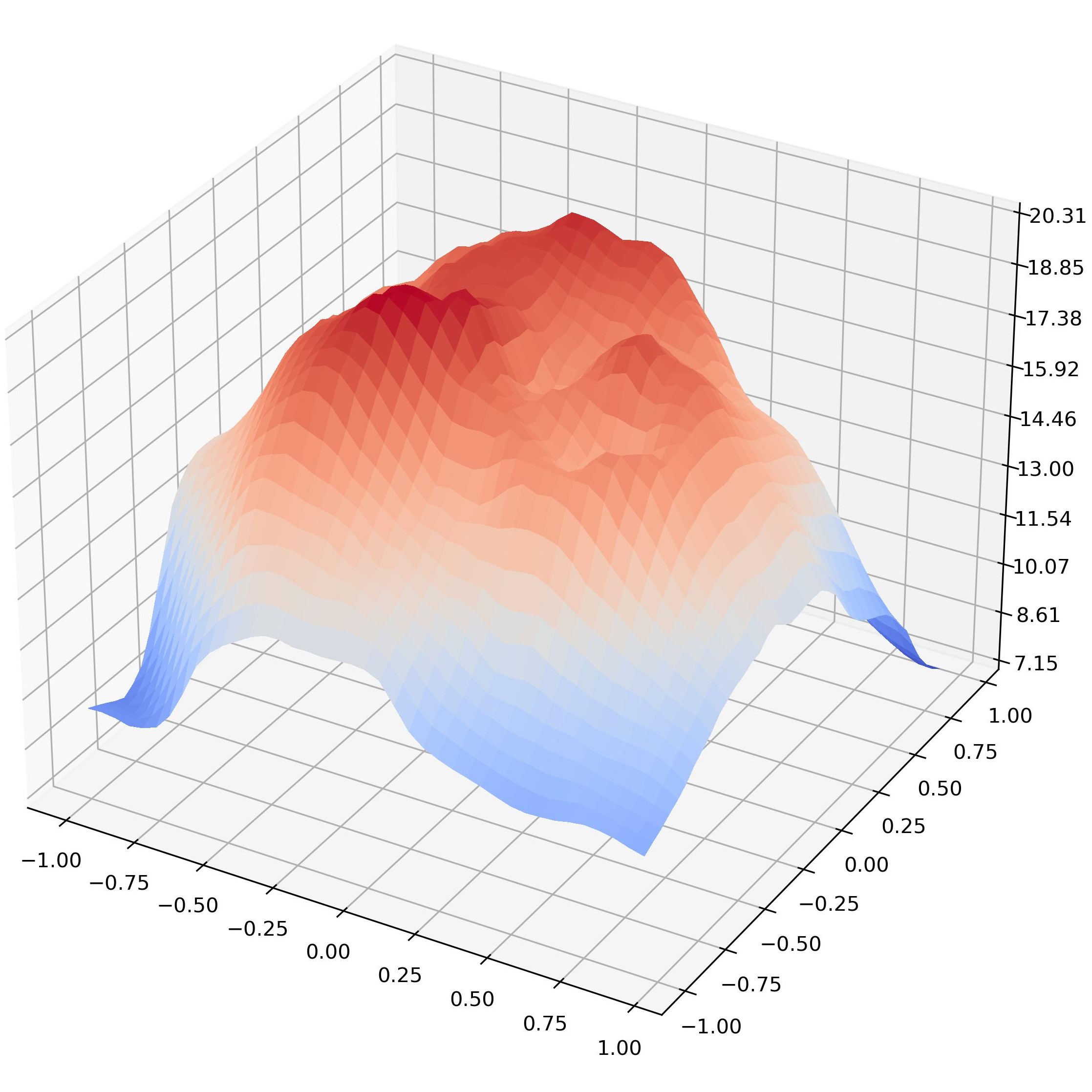}
            \vspace{-0.5cm}
        \end{subfigure}
    \end{minipage}
    \vspace{0.5cm}
    \centering
    \begin{minipage}{0.12\textwidth}
        \begin{subfigure}{\textwidth}
        \centering
            \includegraphics[width=\linewidth]{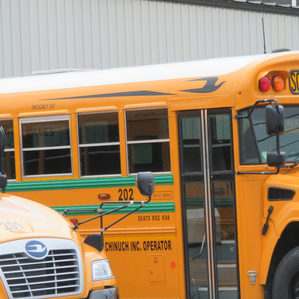}
            \vspace{-0.5cm}
        \end{subfigure}
    \end{minipage}
    \hspace{1mm}
    \begin{minipage}{0.86\textwidth}
        \begin{subfigure}{0.14\textwidth}
            \centering 
            \includegraphics[width=\linewidth]{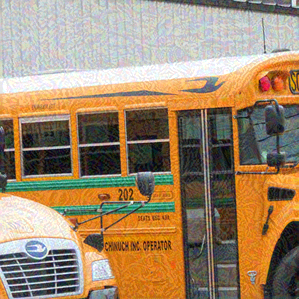}
            \vspace{-0.5cm}
        \end{subfigure}%
        \hspace{3mm}
        \begin{subfigure}{0.14\textwidth} 
            \centering 
            \includegraphics[width=\linewidth]{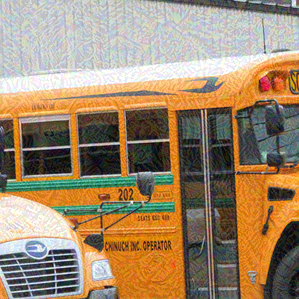}
            \vspace{-0.5cm} 
        \end{subfigure}%
        \hspace{3mm}
        \begin{subfigure}{0.14\textwidth}
            \centering 
            \includegraphics[width=\linewidth]{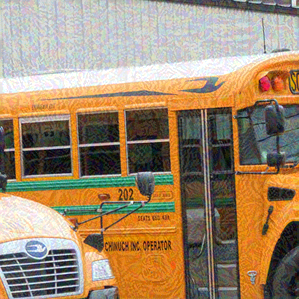}
            \vspace{-0.5cm}
            
        \end{subfigure}
        \hspace{3mm}
        \begin{subfigure}{0.14\textwidth}
            \centering 
            \includegraphics[width=\linewidth]{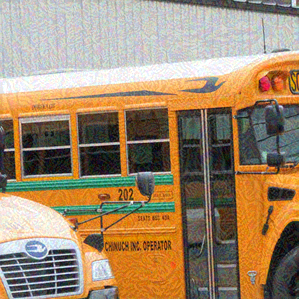}
            \vspace{-0.5cm}
            
        \end{subfigure}%
        \hspace{3mm}
        \begin{subfigure}{0.14\textwidth} 
            \centering 
            \includegraphics[width=\linewidth]{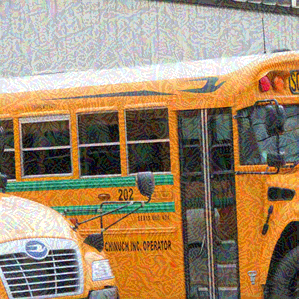}
            \vspace{-0.5cm}
            
        \end{subfigure}%
        \hspace{3mm}
        \begin{subfigure}{0.14\textwidth}
            \centering 
            \includegraphics[width=\linewidth]{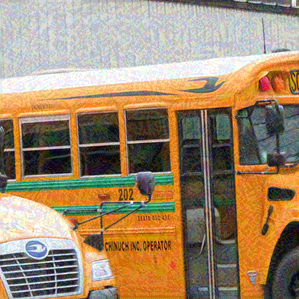}
            \vspace{-0.5cm}
        \end{subfigure}       

        \vspace{3mm}
        \begin{subfigure}{0.14\textwidth}
            \centering 
            \includegraphics[width=\linewidth]{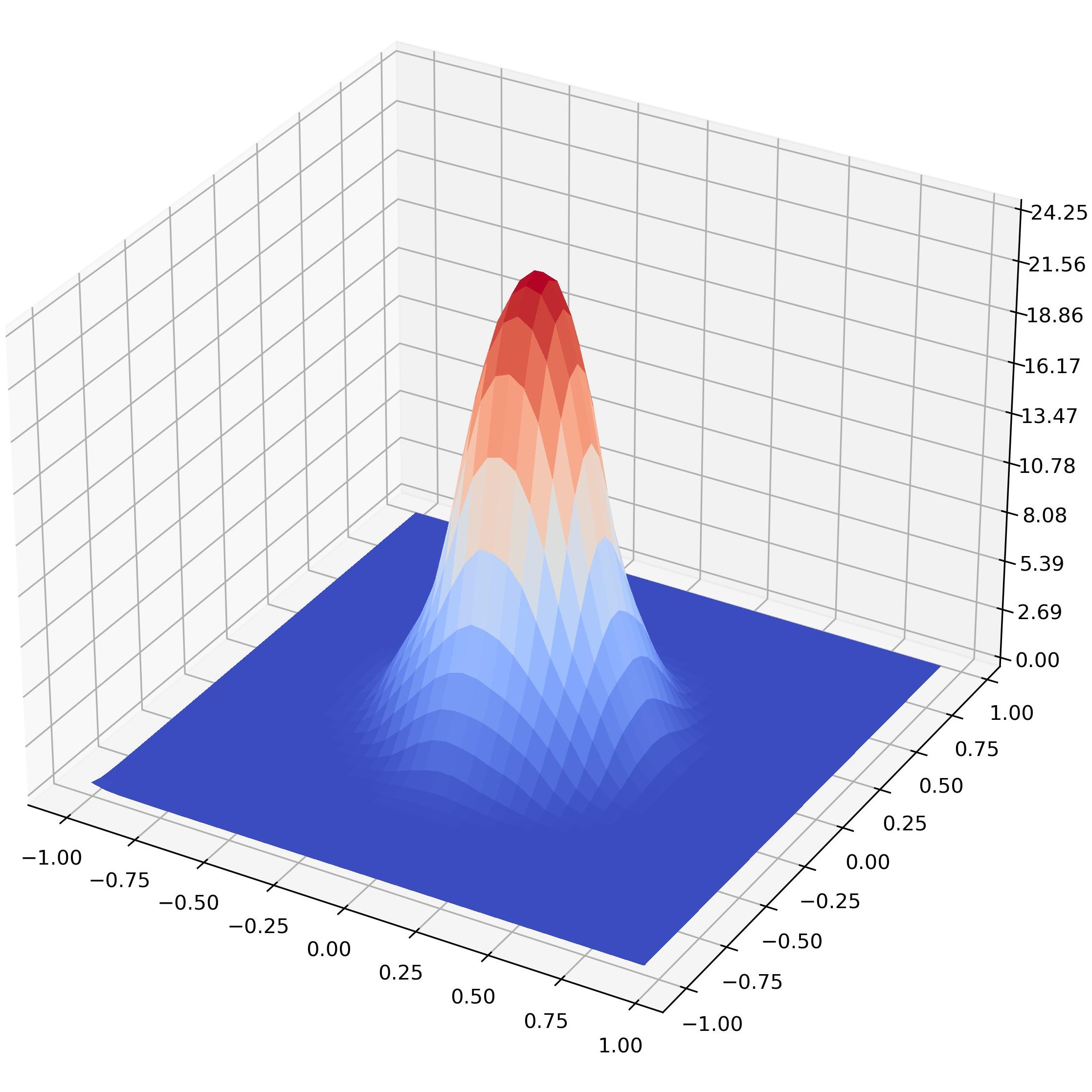}
            \vspace{-0.5cm}
        \end{subfigure}%
        \hspace{3mm}
        \begin{subfigure}{0.14\textwidth} 
            \centering 
            \includegraphics[width=\linewidth]{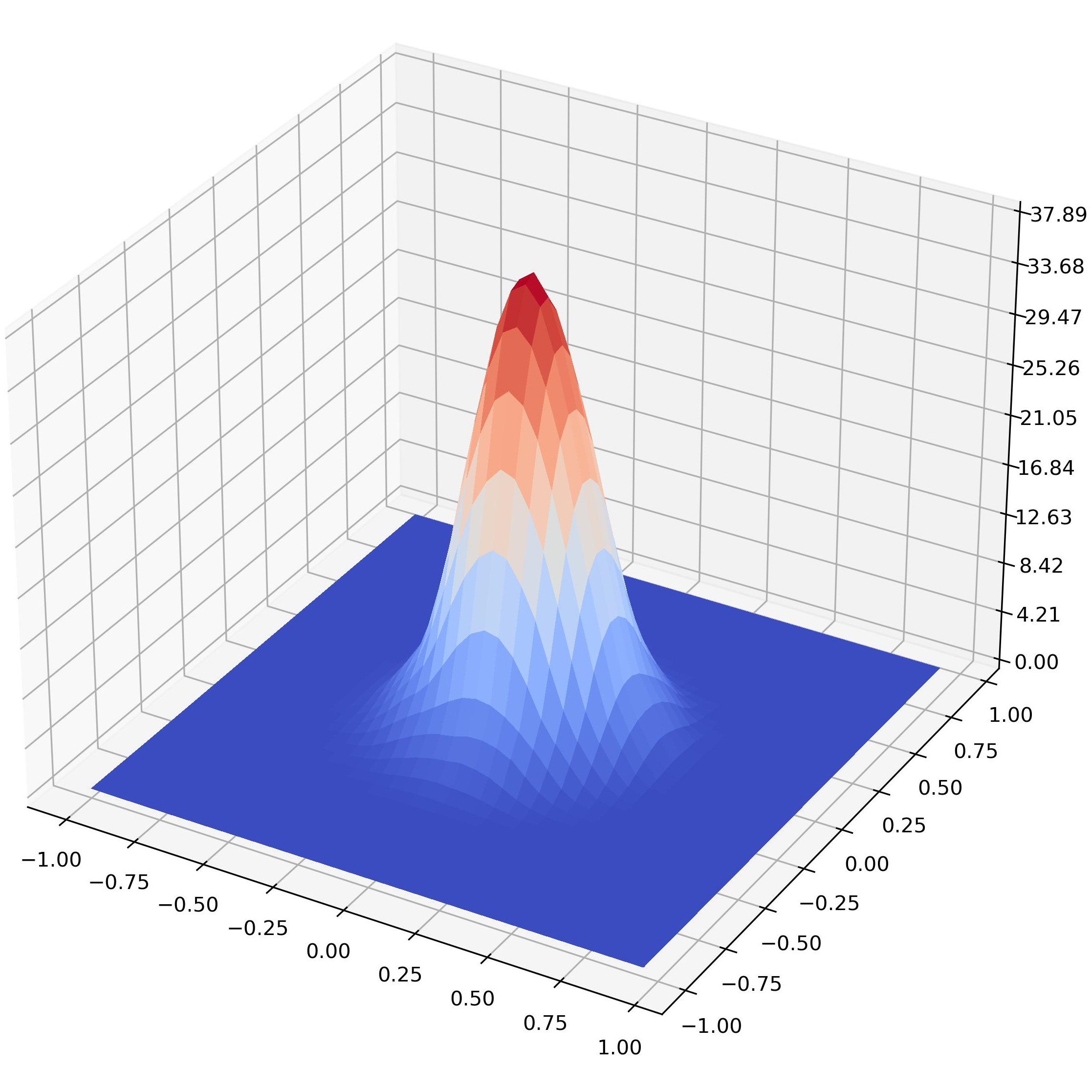}
            \vspace{-0.5cm} 
        \end{subfigure}%
        \hspace{3mm}
        \begin{subfigure}{0.14\textwidth}
            \centering 
            \includegraphics[width=\linewidth]{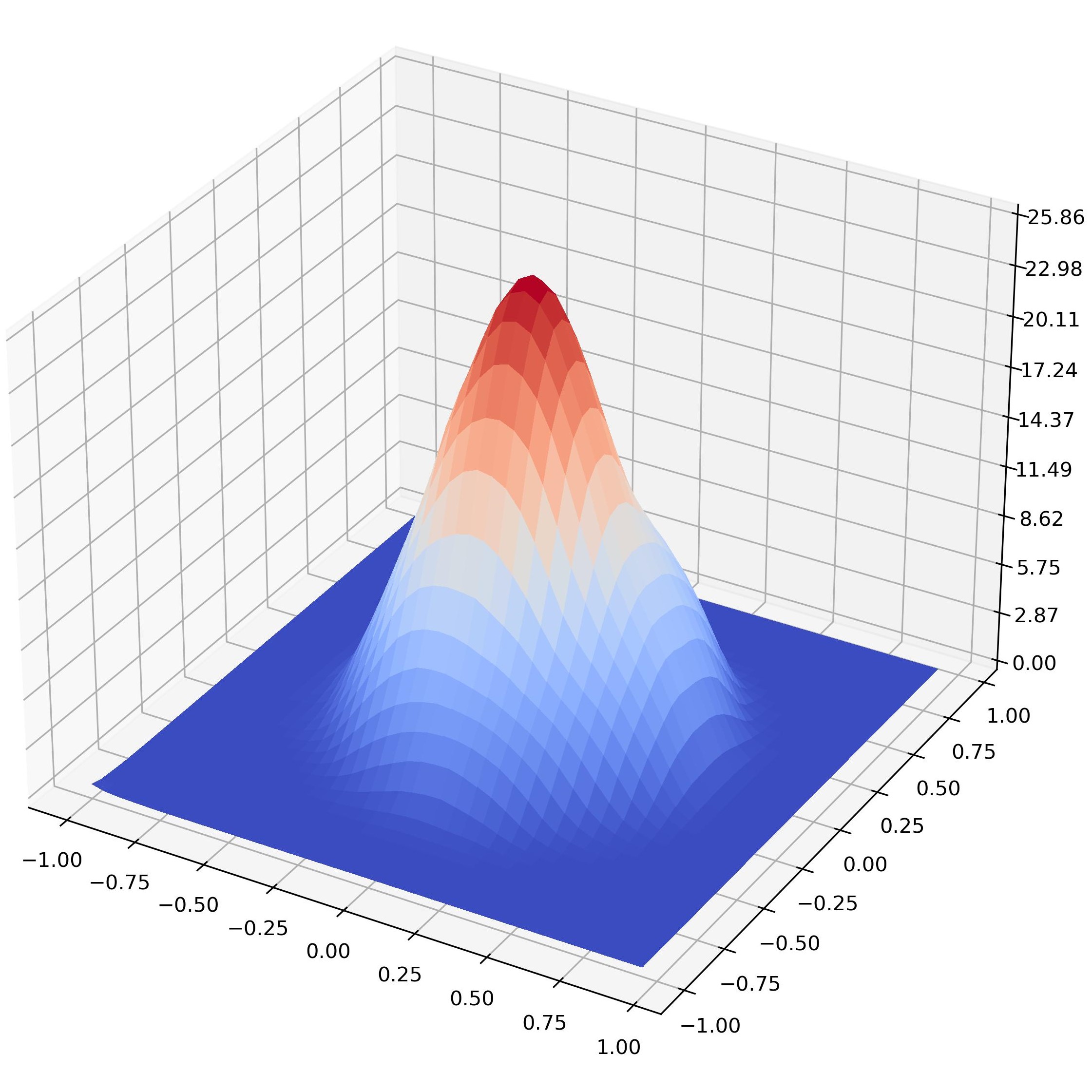}
            \vspace{-0.5cm}
            
        \end{subfigure}
        \hspace{3mm}
        \begin{subfigure}{0.14\textwidth}
            \centering 
            \includegraphics[width=\linewidth]{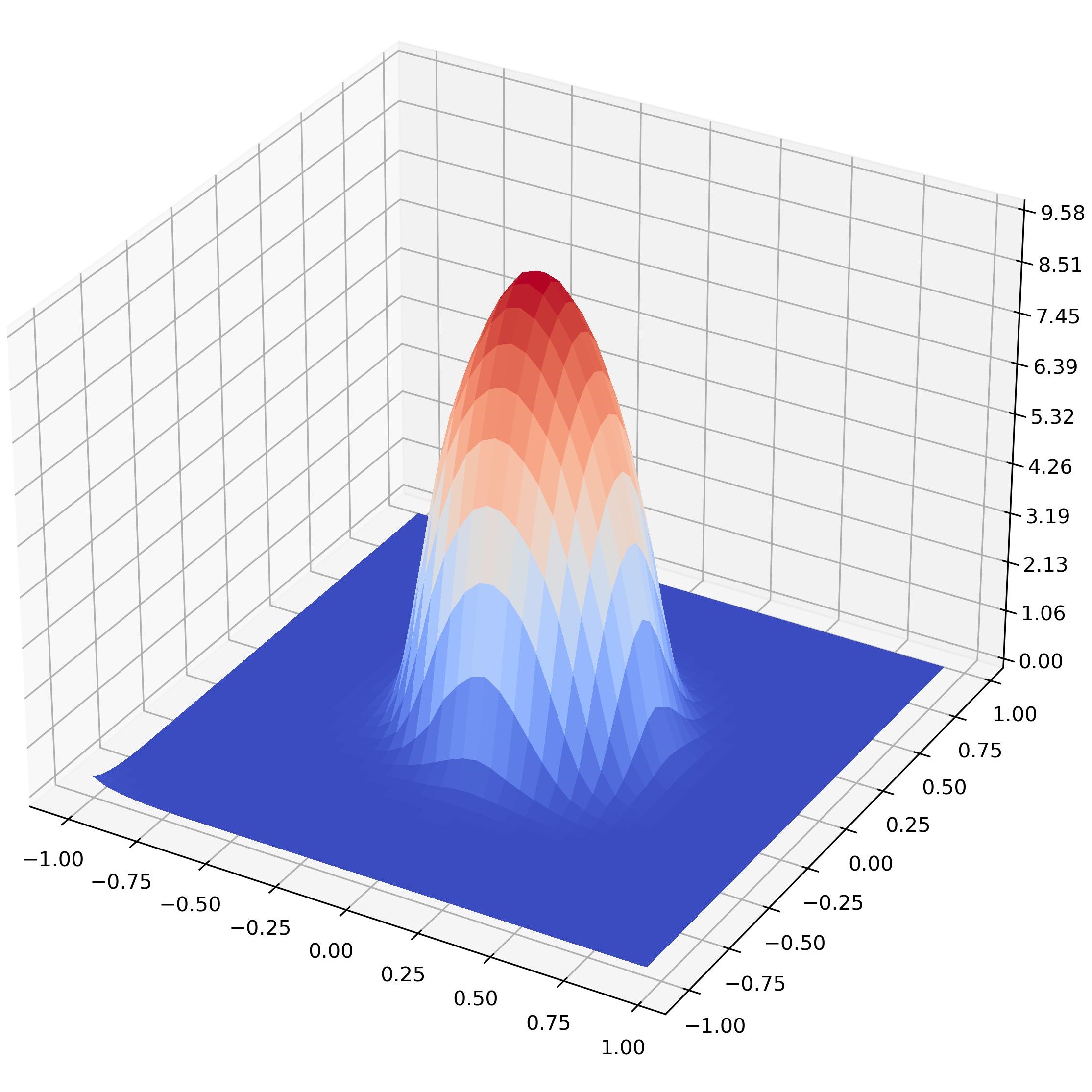}
            \vspace{-0.5cm}
            
        \end{subfigure}%
        \hspace{3mm}
        \begin{subfigure}{0.14\textwidth} 
            \centering 
            \includegraphics[width=\linewidth]{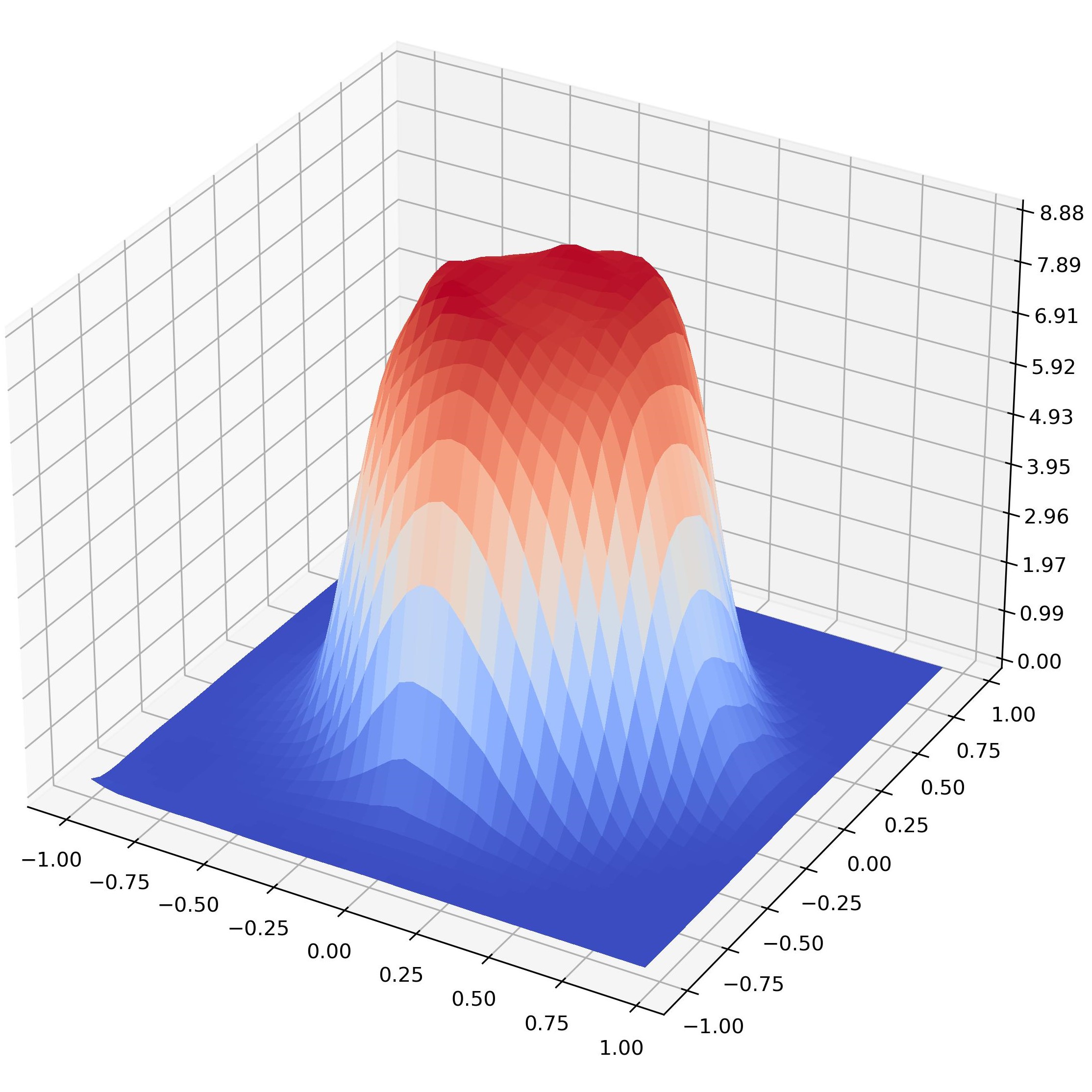}
            \vspace{-0.5cm}
            
        \end{subfigure}%
        \hspace{3mm}
        \begin{subfigure}{0.14\textwidth}
            \centering 
            \includegraphics[width=\linewidth]{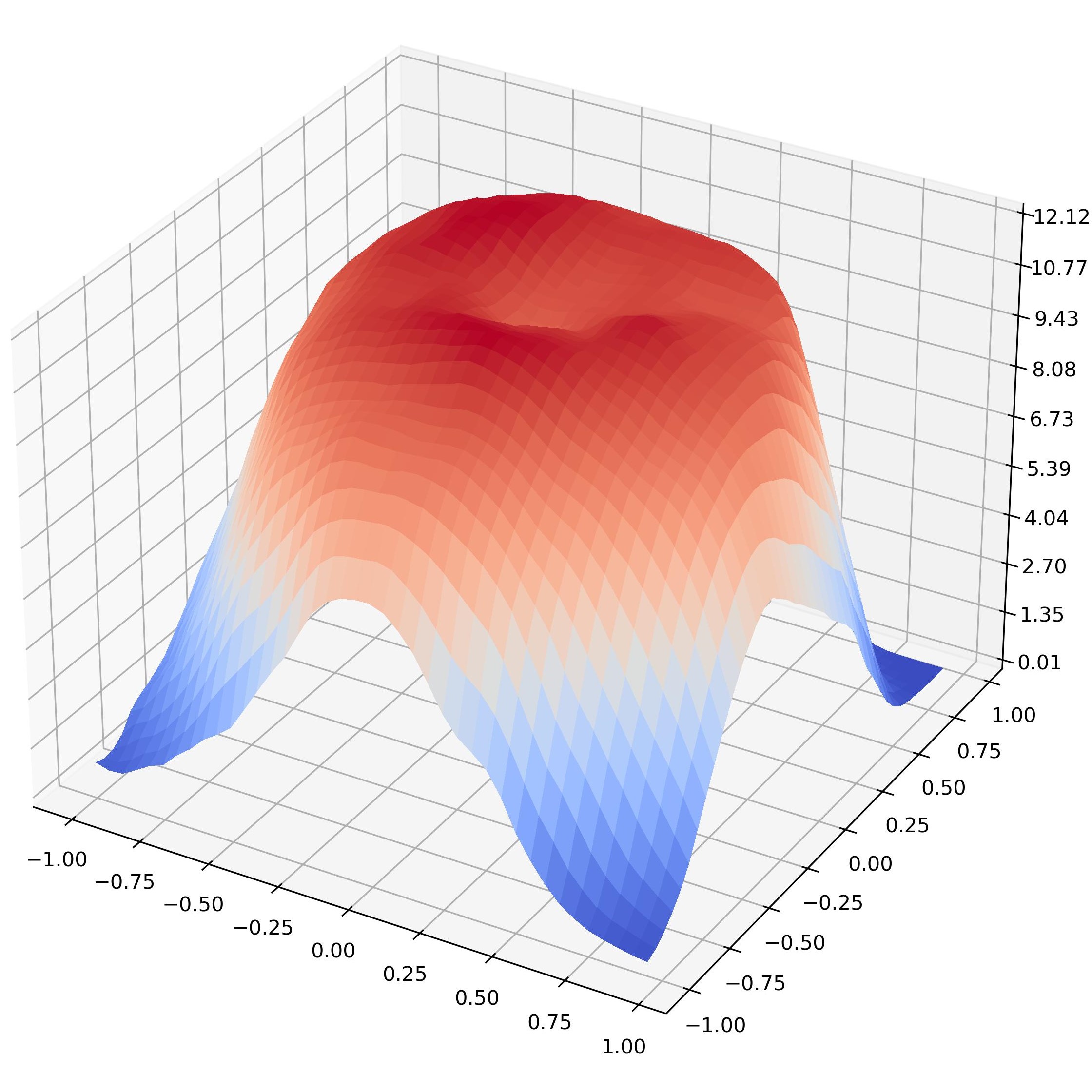}
            \vspace{-0.5cm}
        \end{subfigure}
    \end{minipage}
    
    \caption{Visualization of adversarial examples with their corresponding loss surfaces along two random directions. Here, we randomly sampled five images and generated the adversarial examples on Inc-v3. These selected images include the three images (i.e., the last three ones) that can be successfully transferred using our PGN method but cannot be transferred using other baseline methods. The loss surfaces are also calculated on Inc-v3.}
    \label{fig:surface_maps_appendix}
\end{figure*}

\textbf{Visualization of loss surfaces for more adversarial examples.} We visualize five randomly selected images in the ImageNet-compatible dataset. The adversarial examples are generated by various gradient-based attack methods on Inc-v3. These selected images include the three images that can be successfully transferred using our PGN method but cannot be transferred using other baseline methods. As shown in Fig.~\ref{fig:surface_maps_appendix}, we can observe that our method generates visually similar adversaries as other attacks. Hence, our method demonstrates the capability to guide adversarial examples towards larger and smoother flat regions. This observation substantiates the effectiveness of our PGN method in generating adversarial examples that reside within flat regions, thereby shedding light on the potential role of flat local maxima in generating transferable adversarial examples.

\section{Combined with Other Categories of Attacks}
\subsection{Gradient-based Attacks}
Our PGN attack method can also be combined with various gradient-based attacks. The core of our method involves updating gradients by interpolating the first-order gradients from two samples to approximately minimize the gradient norm. In contrast, conventional gradient-based methods typically utilize a single example for gradient updates. To evaluate the efficacy of our strategy, we incorporate this interpolation approach into previous gradient-based methods, such as I-FGSM (BIM), MI-FGSM, NI-FGSM, VMI-FGSM, EMI-FGSM, and RAP. To simplify the experimental setup, we omitted random sampling and directly substituted the gradient update process of these methods with our proposed strategy.

The experimental results are presented in Table \ref{tab:attack_with_gradient_based}. Notably, when our gradient update strategy is integrated, there is a remarkable improvement in the adversarial transferability of the gradient-based attack methods in the black-box setting. For example, RAP alone achieves an average success rate of $68.80\%$ across the seven models. However, when combined with our PGN method, the average success rate rises to $73.91\%$, exhibiting a significant improvement of $5.11\%$. This outcome underscores the robust scalability of our approach, as it seamlessly integrates with existing methodologies to further amplify the success rate of transfer-based attacks.
\begin{table}[htbp]
  \centering
  \caption{Untargeted attack success rates (\%) of our PGN method, when it is integrated with I-FGSM (BIM), MI-FGSM, NI-FGSM, VMI-FGSM, EMI-FGSM, and RAP, respectively. The adversarial examples are generated on Inc-v3. * indicates the white-box model.}
  \vspace{1mm}
  \setlength{\tabcolsep}{3.0pt}
    \begin{tabular}{|c|ccccccc|c|}
    \hline
    Attack & Inc-v3 & Inc-v4 & IncRes-v2 & Res-101 & Inc-v3$_{ens3}$ & Inc-v3$_{ens4}$ & IncRes-v2$_{ens}$ & Avg.  \\
    \hline
    \hline
    BIM   & \textbf{100.0}*   & 28.1  & 20.6  & 26.7  & 11.9  & 11.9  & 5.0   & 29.17 \\
    \cellcolor[rgb]{0.9, 0.9, 0.9}PGN-BIM & \cellcolor[rgb]{0.9, 0.9, 0.9}\textbf{100.0}*   & \cellcolor[rgb]{0.9, 0.9, 0.9}\textbf{33.1}  & \cellcolor[rgb]{0.9, 0.9, 0.9}\textbf{25.3}  & \cellcolor[rgb]{0.9, 0.9, 0.9}\textbf{30.4}  & \cellcolor[rgb]{0.9, 0.9, 0.9}\textbf{13.7}  & \cellcolor[rgb]{0.9, 0.9, 0.9}\textbf{13.8}    & \cellcolor[rgb]{0.9, 0.9, 0.9}\textbf{6.1}   & \cellcolor[rgb]{0.9, 0.9, 0.9}\textbf{31.77} \\
    \hline
    MI    & \textbf{100.0}*   & 50.8  & 46.3  & 48.9  & 23.3  & 22.2  & 11.7  & 43.31 \\
    \cellcolor[rgb]{0.9, 0.9, 0.9}PGN-MI & \cellcolor[rgb]{0.9, 0.9, 0.9}\textbf{100.0}*   & \cellcolor[rgb]{0.9, 0.9, 0.9}\textbf{56.4}  & \cellcolor[rgb]{0.9, 0.9, 0.9}\textbf{53.3}  & \cellcolor[rgb]{0.9, 0.9, 0.9}\textbf{55.0}  & \cellcolor[rgb]{0.9, 0.9, 0.9}\textbf{24.7}  & \cellcolor[rgb]{0.9, 0.9, 0.9}\textbf{24.7}  & \cellcolor[rgb]{0.9, 0.9, 0.9}\textbf{11.8}  & \cellcolor[rgb]{0.9, 0.9, 0.9}\textbf{46.54} \\
    \hline
    NI    & \textbf{100.0}*   & 62.3  & 59.8   & 57.8  & 22.4 & 22.3  & 11.8  & 48.06 \\
    \cellcolor[rgb]{0.9, 0.9, 0.9}PGN-NI & \cellcolor[rgb]{0.9, 0.9, 0.9}\textbf{100.0}*   & \cellcolor[rgb]{0.9, 0.9, 0.9}\textbf{68.6} & \cellcolor[rgb]{0.9, 0.9, 0.9}\textbf{65.8}  & \cellcolor[rgb]{0.9, 0.9, 0.9}\textbf{61.9} & \cellcolor[rgb]{0.9, 0.9, 0.9}\textbf{25.7}  & \cellcolor[rgb]{0.9, 0.9, 0.9}\textbf{25.8}  & \cellcolor[rgb]{0.9, 0.9, 0.9}\textbf{12.8}  & \cellcolor[rgb]{0.9, 0.9, 0.9}\textbf{51.51} \\
    \hline
    VMI   & \textbf{100.0}*   & 75.2  & 70.2  & 66.0   & 41.7  & 40.9  & 24.8  & 59.83 \\
    \cellcolor[rgb]{0.9, 0.9, 0.9}PGN-VMI & \cellcolor[rgb]{0.9, 0.9, 0.9}\textbf{100.0}*   & \cellcolor[rgb]{0.9, 0.9, 0.9}\textbf{79.8}  & \cellcolor[rgb]{0.9, 0.9, 0.9}\textbf{75.9}  & \cellcolor[rgb]{0.9, 0.9, 0.9}\textbf{69.8}  & \cellcolor[rgb]{0.9, 0.9, 0.9}\textbf{45.1}  & \cellcolor[rgb]{0.9, 0.9, 0.9}\textbf{46.1}   & \cellcolor[rgb]{0.9, 0.9, 0.9}\textbf{27.8}  & \cellcolor[rgb]{0.9, 0.9, 0.9}\textbf{63.50} \\
    \hline
    EMI   & \textbf{100.0}*   & 81.6  & 77.0  & 72.0  & 32.8  & 32.2  & 17.5  & 59.01 \\
    \cellcolor[rgb]{0.9, 0.9, 0.9}PGN-EMI & \cellcolor[rgb]{0.9, 0.9, 0.9}\textbf{100.0}*   & \cellcolor[rgb]{0.9, 0.9, 0.9}\textbf{83.6}  & \cellcolor[rgb]{0.9, 0.9, 0.9}\textbf{81.7}  & \cellcolor[rgb]{0.9, 0.9, 0.9}\textbf{75.9}  & \cellcolor[rgb]{0.9, 0.9, 0.9}\textbf{35.2}  & \cellcolor[rgb]{0.9, 0.9, 0.9}\textbf{34.4}  & \cellcolor[rgb]{0.9, 0.9, 0.9}\textbf{17.5}  & \cellcolor[rgb]{0.9, 0.9, 0.9}\textbf{61.19} \\
    \hline
    RAP   & 99.9  & 84.6  & 79.3  & 76.6  & 57.2  & 51.3  & 32.7  & 68.80 \\
    \cellcolor[rgb]{0.9, 0.9, 0.9}PGN-RAP & \cellcolor[rgb]{0.9, 0.9, 0.9}\textbf{100.0}*  & \cellcolor[rgb]{0.9, 0.9, 0.9}\textbf{90.3}  & \cellcolor[rgb]{0.9, 0.9, 0.9}\textbf{86.4}  & \cellcolor[rgb]{0.9, 0.9, 0.9}\textbf{85.7}  & \cellcolor[rgb]{0.9, 0.9, 0.9}\textbf{59.5}  & \cellcolor[rgb]{0.9, 0.9, 0.9}\textbf{56.3}  & \cellcolor[rgb]{0.9, 0.9, 0.9}\textbf{39.3}  & \cellcolor[rgb]{0.9, 0.9, 0.9}\textbf{73.91} \\
    \hline
    \end{tabular}%
  \label{tab:attack_with_gradient_based}%
\end{table}%

\subsection{Skip Gradient Method}
Moreover, our method can also be combined with the Skip Gradient Method (SGM) \citep{wu2020skip}. By using MI-FGSM as the backbone method, we first compare our PGN method with SGM on the ResNet-18 model. Since SGM modifies the backpropagation and is compatible with our PGN, we also integrate PGN into SGM to evaluate its generality to other attacks. As shown in Table \ref{tab:attack_sgm}, our PGN consistently exhibits better attack performance than SGM, which further shows its superiority in boosting adversarial transferability. Besides, PGN-SGM outperforms both PGN and SGM, showing its remarkable compatibility with various attacks.
\begin{table}[h]
  \centering
  \caption{Untargeted attack success rates (\%) of our PGN method with SGM method in the single model setting. The adversarial examples are crafted on Res-18.}
  \vspace{1mm}
  \setlength{\tabcolsep}{2.50pt}
    \begin{tabular}{|c|ccccccc|c|}
    \hline
    Source:Res-18 & Res-101 & Res-152 & Inc-v3 & Inc-v4 & IncRes-v2 & Inc-v3$_{ens3}$ & Inc-v3$_{ens4}$ & Avg. \\
    \hline
    MI    & 82.8  & 73.3  & 54.5  & 48.7  & 33.9  & 17.4  & 18.1  & 46.96 \\
    SGM   & 89.1  & 82.5  & 66.0  & 58.8  & 45.4  & 20.8  & 20.8  & 54.77 \\
    \cellcolor[rgb]{0.9, 0.9, 0.9}PGN   & \cellcolor[rgb]{0.9, 0.9, 0.9}\textbf{95.9} & \cellcolor[rgb]{0.9, 0.9, 0.9}\textbf{92.9} & \cellcolor[rgb]{0.9, 0.9, 0.9}\textbf{77.9} & \cellcolor[rgb]{0.9, 0.9, 0.9}\textbf{75.9} & \cellcolor[rgb]{0.9, 0.9, 0.9}\textbf{58.7} & \cellcolor[rgb]{0.9, 0.9, 0.9}\textbf{33.7} & \cellcolor[rgb]{0.9, 0.9, 0.9}\textbf{36.3} & \cellcolor[rgb]{0.9, 0.9, 0.9}\textbf{67.33} \\
    \hline
    \cellcolor[rgb]{0.9, 0.9, 0.9}PGN-SGM & \cellcolor[rgb]{0.9, 0.9, 0.9}\textbf{96.7} & \cellcolor[rgb]{0.9, 0.9, 0.9}\textbf{93.8} & \cellcolor[rgb]{0.9, 0.9, 0.9}\textbf{80.1} & \cellcolor[rgb]{0.9, 0.9, 0.9}\textbf{76.3} & \cellcolor[rgb]{0.9, 0.9, 0.9}\textbf{62.8} & \cellcolor[rgb]{0.9, 0.9, 0.9}\textbf{33.8} & \cellcolor[rgb]{0.9, 0.9, 0.9}\textbf{37.0} & \cellcolor[rgb]{0.9, 0.9, 0.9}\textbf{68.64} \\
    \hline
    \end{tabular}%
  \label{tab:attack_sgm}%
\end{table}%


\section{Ablation Studies for More Hyper-parameters}
\subsection{The Number of Sampled Examples, $N$}
\begin{figure}[h]
    \centering
    \begin{subfigure}{0.50\linewidth}
        \centering
        \begin{tikzpicture}[clip]
        \begin{axis}[
        	xlabel={\tiny The number of sampled examples $N$}, 
        	ylabel={\tiny Attack success rates (\%)},,
        	grid=both,
        	minor grid style={gray!25, dashed},
        	major grid style={gray!25, dashed},
            scale only axis,
        	width=0.99\linewidth,
            ylabel style={font=\tiny, yshift=-5pt},
            xlabel style={font=\tiny, yshift=5pt},
            tick label style={font=\tiny},
            legend style={font=\tiny, fill opacity=0.8,legend columns=2},
            legend pos=south west
        ]
            \addplot[line width=1pt,solid,mark=*,color=cyan, mark options={mark size=1pt}] %
            	table[x=N,y=Incv4,col sep=comma]{images/Parameters/Search_N.csv};
             \addlegendentry{Inc-v4};
            
            \addplot[line width=1pt,solid,mark=pentagon,color=purple, mark options={mark size=1pt}] %
            	table[x=N,y=IncRes_v2,col sep=comma]{images/Parameters/Search_N.csv};
             \addlegendentry{IncRes-v2};
             
            \addplot[line width=1pt,solid,mark=triangle,color=brown, mark options={mark size=1pt}] %
            	table[x=N,y=Res101,col sep=comma]{images/Parameters/Search_N.csv};
             \addlegendentry{Res-101};
             
            \addplot[line width=1pt,solid,mark=square,color=orange, mark options={mark size=1pt}] %
            	table[x=N,y=Res152,col sep=comma]{images/Parameters/Search_N.csv};
             \addlegendentry{Res-152};


            \addplot[line width=1pt,solid,mark=oplus,color=red, mark options={mark size=1pt}] %
            	table[x=N,y=Incv3_ens4,col sep=comma]{images/Parameters/Search_N.csv};
             \addlegendentry{Inc-v3$_{ens4}$};

            \addplot[line width=1pt,solid,mark=diamond,color=teal, mark options={mark size=1pt}] %
            	table[x=N,y=IncRes_v2_ens,col sep=comma]{images/Parameters/Search_N.csv};
             \addlegendentry{IncRes-v2$_{ens}$};
        \end{axis}
        \end{tikzpicture}
        \label{fig:parameters_studies:N}
    \end{subfigure}

    \caption{Untargeted attack success rates (\%) on six black-box models with the different number of sampled samples $N$. The adversarial examples are generated by PGN on Inc-v3.}
    \label{fig:parameters_studies_n}
\end{figure}
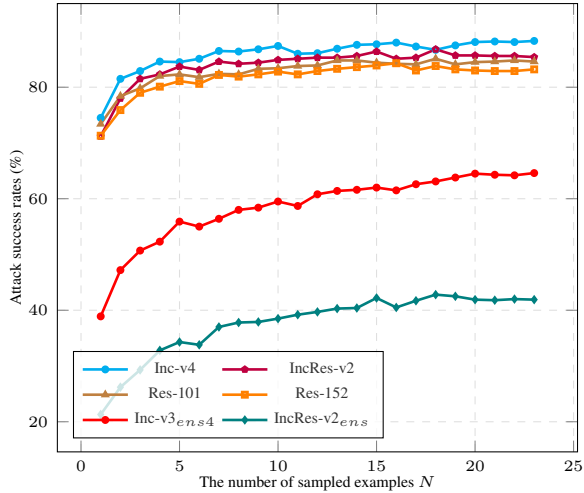

In this study, we employ random sampling of multiple examples and calculate the average gradients of these examples to mitigate the variance resulting from random sampling during the iterative process. To investigate the influence of the number of sampled examples, denoted as $N$, we conduct ablation experiments to analyze this parameter. As illustrated in Figure \ref{fig:parameters_studies_n}, when $N=1$, our method demonstrates the lowest level of transferability. However, as we increase the value of $N$, the transferability exhibits rapid improvement until $N=12$, after which it gradually converges for normally trained models. Notably, when $N>12$, a slight performance improvement can still be achieved by increasing the number of sampled examples in our PGN method. To strike a balance between transferability and computational overhead, we set $N=20$ in our work. This observation further substantiates that sampling random examples from the vicinity of the adversarial example effectively facilitates neighborhood exploration. Consequently, it stabilizes the gradient update process and encourages the discovery of flatter regions by the adversarial example.

\subsection{Finite Difference Step-size $\alpha$}
In this work, we simply set the finite difference step-size to $\alpha = 1.6/255 \approx 6.27$e-3 to avoid redundant hyperparameters. In order to study the effect of the smaller step-size $\alpha$ on the finite difference method, we performed more experiments. We investigate the transferability of adversarial examples when the alpha is set to 1e-4 and 1e-5, and the experimental results are reported in Table \ref{tab:ablation_fd_step}. The results show that smaller step-sizes cannot improve adversarial transferability. This is because the smaller the step-size, the closer the two neighboring gradients are, leading our method to degrade to MI-FGSM.

With the current choice of $\alpha$, our method also approximates the actual Hessian/vector product. We compared the cosine similarity of perturbations generated by the Hessian/vector product and the finite difference method. The average cosine similarity of 1,000 images is about $0.9035$. Note that existing adversarial attacks typically rely on the sign of the gradient, rather than requiring an exact gradient value. Thus, we can approximate the second-order Hessian matrix by using the finite difference method to accelerate the attack process.

\begin{table}[h]
  \centering
  \caption{Untargeted attack success rate (\%) when using smaller step sizes in the finite-difference method. * indicates the white-box model.}
    \vspace{1mm}
    \setlength{\tabcolsep}{3.0pt}
    \begin{tabular}{|c|cccc|}
    \hline
    Size & Inc-v3   & Inc-v4  & IncRes-v2 & Res-101  \\
    \hline
    \hline
    $6.27e-3$ & 100.0* & 91.6 & 90.6 & 89.1\\
    \hline
    $1e-4$ & 100.0* & 86.4 & 84.6 & 82.6 \\
    \hline
    $1e-5$ & 100.0* & 85.3 & 85.0 & 82.4 \\
    \hline
    \end{tabular}%
  \label{tab:ablation_fd_step}%
\end{table}%

\section{Differences between Our method and Related Works}
There are two previous works similar to our proposed attack method (i.e., PGN), where one is Reverse Adversarial Perturbation (RAP)~\citep{qin2022boosting}, which is an adversarial attack method. And the other is Gradient Norm Penalty (GNP)~\citep{zhao2022penalizing}, which was proposed to improve generalization. To demonstrate the novelty and significance of the proposed method, we give a detailed discussion about the differences and similarities compared to these two works.

\subsection{Differences with RAP}
Reverse adversarial perturbation (RAP) is a closely related work that encourages adversarial examples to be located at a region with low loss values. However, there are still many differences between our method and RAP.

\textbf{Empirical verification.} Both RAP and our PGN aim to achieve flat local optima for more transferable adversarial examples. Moreover,this is the first work to provide empirical verification that adversarial examples in flat local optima are more transferable.

\textbf{Methodology.} RAP injects worst-case perturbations into the optimization process to maximize the loss and achieve flat local optima. In contrast, PGN introduces a gradient norm into the loss to achieve a flat local optimum. Hence, the proposed method differs significantly from RAP.

\textbf{Efficiency.} Our method is more computationally efficient than RAP. In RAP, the outer loop is $400$, and the inner loop is $8$. In contrast, our PGN method employs finite differences to approximate the second-order Hessian matrix and samples multiple points to obtain a more stable gradient, which requires only $10$ outer loops and $20$ inner loops for the computation. Here we also compared the time consumed for each batch of images between our method and RAP. The experiment was conducted on RTX 2080 Ti with a CUDA environment. We chose Inc-v3 as the source model and set the batch size to 10. Experimental results show that RAP will consume $207.19 s$ in each batch size, while our method only consumes $29.85 s$, which validates that our PGN is more computationally efficient.

\subsection{Compared with Gradient Norm Penalty}
Zhao \etal ~\citep{zhao2022penalizing} proposed a gradient norm penalty method in deep learning to improve generalization, which is similar to our method. However, there are also some differences between these two works, which are summarized as follows:

\textbf{Different goals.} Zhao \etal primarily focus on computing flat local minima to improve generalization during model training. In contrast, our main objective is to investigate the potential impact of flat local minima on the transferability of adversarial attacks. Previous research in the field of adversarial attacks has shown limited attention to the relationship between flat local optima and adversarial transferability. Consequently, our work significantly contributes a new insight into the domain of adversarial attacks.

\textbf{Objective function.} A distinguishing feature of our proposed method lies in its emphasis on the gradient information surrounding the adversarial example. Specifically, we approximate the second-order Hessian matrix using the finite difference method, which plays a crucial role in our approach. The work presented by Zhao \etal also supports and provides theoretical foundations for the feasibility of our method within the realm of adversarial attacks.

In summary, our main motivation in this work is to explore whether flat local optimum can improve adversarial transferability. To the best of our knowledge, it is also the first work that penalized the gradient norm and the finite difference method support our motivation in the field of adversarial attacks.

\section{Additional Experimental Results}
\subsection{Evaluated on Transformer-based Models}
To further validate the effectiveness of our PGN method, we evaluated the performance on the Transformer-based models, \ie  ViT \citep{dosovitskiy2020vit}, PiT \citep{heo2021rethinking}, Visformer \citep{chen2021visformer}, and Swin \citep{liu2021swin}. The adversarial examples are generated on Inc-v3 and the attack success rates are summarized in Table \ref{tab:attack_vit}. It can be seen that our PGN can consistently outperform the baselines on these Transformers, showing its high effectiveness and generality to various architectures.
\begin{table}[h]
  \centering
  \caption{Untargeted attack success rate (\%) of adversarial examples in Transformer-based models. The adversarial examples are generated on Inc-v3.}
    \vspace{1mm}
    \setlength{\tabcolsep}{5.0pt}
    \begin{tabular}{|c|cccc|c|}
    \hline
    Attack & ViT   & PiT   & Visformer & Swin  & Avg. \\
    \hline
    \hline
    MI    & 18.9  & 18.1  & 23.7  & 22.2  & 20.73 \\
    NI    & 20.0  & 19.9  & 25.5  & 25.5  & 22.73 \\
    VMI   & 28.4  & 34.0  & 41.1  & 40.8  & 36.08 \\
    EMI   & 26.0  & 27.3  & 36.9  & 36.7  & 31.73 \\
    RAP   & 35.2  & 41.8  & 50.6  & 50.2  & 44.45 \\
    \cellcolor[rgb]{0.9, 0.9, 0.9}PGN   & \cellcolor[rgb]{0.9, 0.9, 0.9}\textbf{44.7} & \cellcolor[rgb]{0.9, 0.9, 0.9}\textbf{53.2} & \cellcolor[rgb]{0.9, 0.9, 0.9}\textbf{65.5} & \cellcolor[rgb]{0.9, 0.9, 0.9}\textbf{64.5} & \cellcolor[rgb]{0.9, 0.9, 0.9}\textbf{56.98} \\
    \hline
    \end{tabular}%
  \label{tab:attack_vit}%
\end{table}%

\subsection{Attack Defense Models}
\label{sec:defense}
In this subsection, besides normally trained models and adversarially trained models, we further validate the effectiveness of our methods on other defenses, including Bit-Red \citep{xu2018feature}, ComDefend \citep{jia2019comdefend}, JPEG \citep{guo2018countering}, HGD \citep{liao2018defense}, R\&P \citep{xie2018mitigating}, NIPs-r3 \citep{naseer2020self}, FD \citep{liu2019feature}, NPR \citep{naseer2020self}, and RS \citep{cohen2019certified}. The adversarial examples are generated on an ensemble of Inc-v3, Inc-v4, and IncRes-v2, and the weight for each model is $1/3$.

The experimental results are reported in Table \ref{tab:attack_defense_models}. In the context of ensemble models, it is evident that our algorithm can considerably enhance existing attack methods. For instance, VMI, EMI, and RAP achieve average success rates of $54.91\%$, $61.59\%$, and $69.94\%$, respectively, against the six defense models. In contrast, our proposed PGN method achieves an average success rate of $77.08\%$, surpassing them by $22.17\%$, $15.49\%$, and $7.14\%$, respectively. This notable improvement demonstrates the remarkable effectiveness of our proposed method against adversarially trained models as well as other defense models. Consequently, it poses a more substantial threat to advanced defense models. These findings further validate that the discovery of adversarial examples within flat regions can significantly enhance the transferability of adversarial attacks.
\begin{table*}[htbp]
  \centering
  \caption{Untargeted attack success rates (\%) on six defense models. The adversarial examples are crafted on the ensemble models, \textit{i.e.} Inc-v3, Inc-v4 and IncRes-v2.}
    \vspace{1mm}
    \setlength{\tabcolsep}{5.0pt}
    \begin{tabular}{|c|ccccccccc|c|}
    \hline
    Attack & ComDefend & JPEG  & NIPs-r3 & FD & R\&P  & HGD & Bit-Red  & NPR & RS & AVG. \\
    \hline
    \hline
    MI & 54.9 & 49.5 & 29.9  & 51.7 & 22.2 & 24.8 & 23.8 & 36.5 & 30.3 & 39.96 \\
    NI  & 56.9 & 50.8 & 29.3  & 53.6 & 23.1 & 22.3 & 23.9 & 38.1 & 80.7 & 36.52\\
    VMI  & 72.4 & 71.5 & 58.1  & 67.8 & 50.6 & 54.3 & 39.0 & 44.9 & 35.6 & 54.91\\
    EMI  & 78.2 & 74.1 & 69.1  & 74.8 & 60.1 & 64.8 & 47.6 & 46.8 & 38.8 & 61.59\\
    RAP  & 89.5 & 88.1 & 81.0  & 79.6 & 73.1 & 72.3 & 54.6 & 49.6 & 41.7 & 69.94\\
    \cellcolor[rgb]{0.9, 0.9, 0.9}PGN & \cellcolor[rgb]{0.9, 0.9, 0.9}\textbf{93.7} & \cellcolor[rgb]{0.9, 0.9, 0.9}\textbf{91.3} & \cellcolor[rgb]{0.9, 0.9, 0.9}\textbf{88.3} & \cellcolor[rgb]{0.9, 0.9, 0.9}\textbf{85.7} & \cellcolor[rgb]{0.9, 0.9, 0.9}\textbf{83.6} & \cellcolor[rgb]{0.9, 0.9, 0.9}\textbf{82.5} & \cellcolor[rgb]{0.9, 0.9, 0.9}\textbf{72.1} & \cellcolor[rgb]{0.9, 0.9, 0.9}\textbf{51.3} & \cellcolor[rgb]{0.9, 0.9, 0.9}\textbf{45.2} & \cellcolor[rgb]{0.9, 0.9, 0.9}\textbf{77.08}\\
    \hline
    \end{tabular}%
  \label{tab:attack_defense_models}%
\end{table*}%


\subsection{Attack Success Rates on CIFAR-10}
To further illustrate the effectiveness of our PGN method on different datasets, we conduct experiments on CIFAR-10 \citep{krizhevsky2009learning}. we set the hyperparameters as follows: maximum perturbation $\epsilon=8/255$, number of iterations $T=10$, and step size $\alpha=1/255$. We compare our PGN method with various gradient-based attacks, including MI-FGSM, NI-FGSM, VMI-FGSM, EMI-FGSM, and RAP. The adversarial examples are generated on the VGG-16, ResNet-50, and DenseNet-121 models, respectively. The results in Table \ref{tab:Attack_on_cifar10} clearly show that our PGN method can enhance the attack transferability on the CIFAR-10 dataset. This verifies our motivation that adversarial examples located in flat local regions tend to exhibit better transferability across diverse models. Moreover, our attack method shows superior performance when applied to other datasets, reinforcing its versatility and effectiveness.

\begin{table}[htbp]
  \centering
    \caption{Untargeted attack success rates (\%) on the CIFAR-10 dataset for the attack methods in the single model setting. The adversarial examples are crafted on VGG-16, ResNet-50 (Res-50), and DenseNet-121, respectively.}
    \vspace{2mm}
  \begin{subtable}{1.0\linewidth}
  \centering
  \setlength{\tabcolsep}{2.5pt}
    \begin{tabular}{|c|cccccccc|}
    \hline
    Attack & MobileNet & VGG-19 & GoogLeNet & Inc-v3 & DenseNet-121 & DenseNet-169 & Res-34 & Res-50 \\
    \hline
    \hline
    MI    & 52.18 & 57.56 & 47.29 & 52.74 & 40.96 & 42.40  & 41.72 & 41.93 \\
    NI    & 56.13 & 61.36 & 49.19 & 54.87 & 37.61 & 39.74 & 38.74 & 38.90 \\
    VMI   & 66.14 & 68.05 & 60.89 & 65.63 & 55.62 & 57.21 & 55.35 & 56.46 \\
    EMI   & 70.69 & 74.36 & 66.78 & 70.56 & 59.98 & 63.04 & 60.47 & 61.83 \\
    RAP   & 77.98 & 78.43 & 73.41 & 77.86 & 68.30 & 69.74 & 65.48 & 66.27 \\
    \cellcolor[rgb]{0.9, 0.9, 0.9}PGN   & \cellcolor[rgb]{0.9, 0.9, 0.9}\textbf{85.97} & \cellcolor[rgb]{0.9, 0.9, 0.9}\textbf{86.73} & \cellcolor[rgb]{0.9, 0.9, 0.9}\textbf{82.82} & \cellcolor[rgb]{0.9, 0.9, 0.9}\textbf{85.59} & \cellcolor[rgb]{0.9, 0.9, 0.9}\textbf{72.48} & \cellcolor[rgb]{0.9, 0.9, 0.9}\textbf{74.66} & \cellcolor[rgb]{0.9, 0.9, 0.9}\textbf{71.62} & \cellcolor[rgb]{0.9, 0.9, 0.9}\textbf{72.95} \\
    \hline
    \end{tabular}%
    \subcaption{Untargeted attack success rates (\%) for the adversarial examples crafted on VGG-16.}
  \end{subtable}
  
  \begin{subtable}{1.0\linewidth}
  \centering
  \setlength{\tabcolsep}{2.5pt}
    \begin{tabular}{|c|cccccccc|}
    \hline
    Attack & MobileNet & VGG16 & VGG19 & GoogLeNet & Inc-v3 & DenseNet-121 & DenseNet-169 & Res-34 \\
    \hline
    \hline
    MI    & 70.42 & 67.37 & 65.8  & 63.06 & 69.02 & 72.39 & 73.34 & 67.78 \\
    NI    & 71.97 & 65.57 & 63.76 & 63.28 & 69.13 & 71.03 & 72.78 & 65.02 \\
    VMI   & 77.60 & 74.27 & 73.26 & 71.11 & 75.99 & 76.80 & 77.68 & 73.54 \\
    EMI   & 80.11 & 78.66 & 77.43 & 76.34 & 78.12 & 79.68 & 80.12 & 77.24 \\
    RAP   & 86.92 & 84.24 & 83.46 & 80.68 & 81.75 & 83.54 & 84.98 & 81.36 \\
    \cellcolor[rgb]{0.9, 0.9, 0.9}PGN   & \cellcolor[rgb]{0.9, 0.9, 0.9}\textbf{90.88} & \cellcolor[rgb]{0.9, 0.9, 0.9}\textbf{88.68} & \cellcolor[rgb]{0.9, 0.9, 0.9}\textbf{88.07} & \cellcolor[rgb]{0.9, 0.9, 0.9}\textbf{85.79} & \cellcolor[rgb]{0.9, 0.9, 0.9}\textbf{89.53} & \cellcolor[rgb]{0.9, 0.9, 0.9}\textbf{89.93} & \cellcolor[rgb]{0.9, 0.9, 0.9}\textbf{90.91} & \cellcolor[rgb]{0.9, 0.9, 0.9}\textbf{87.19} \\
    \hline
    \end{tabular}%
    \subcaption{Untargeted attack success rates (\%) for the adversarial examples crafted on Res-50.}
    \end{subtable}%
    
    \begin{subtable}{1.0\linewidth}
      \centering
      \setlength{\tabcolsep}{4.0pt}
        \begin{tabular}{|c|cccccccc|}
        \hline
        Attack & MobileNet & VGG16 & VGG19 & GoogLeNet & Inc-v3 & DenseNet-169 & Res-34 & Res-50 \\
        \hline
        \hline
        MI    & 63.47 & 60.64 & 60.08 & 57.39 & 63.35 & 71.09 & 61.99 & 67.37 \\
        NI    & 66.86 & 61.80 & 60.85 & 60.30  & 66.54 & 75.92 & 61.57 & 69.25 \\
        VMI   & 71.28 & 68.49 & 68.01 & 65.40  & 70.62 & 75.51 & 68.42 & 72.50 \\
        EMI   & 74.36 & 75.66 & 73.54 & 70.41 & 75.23 & 78.94 & 73.64 & 77.45 \\
        RAP   & 79.98 & 80.22 & 78.68 & 76.39 & 80.55 & 84.25 & 78.65 & 80.03 \\
        \cellcolor[rgb]{0.9, 0.9, 0.9}PGN   & \cellcolor[rgb]{0.9, 0.9, 0.9}\textbf{86.73} & \cellcolor[rgb]{0.9, 0.9, 0.9}\textbf{85.12} & \cellcolor[rgb]{0.9, 0.9, 0.9}\textbf{84.66} & \cellcolor[rgb]{0.9, 0.9, 0.9}\textbf{81.82} & \cellcolor[rgb]{0.9, 0.9, 0.9}\textbf{86.26} & \cellcolor[rgb]{0.9, 0.9, 0.9}\textbf{88.35} & \cellcolor[rgb]{0.9, 0.9, 0.9}\textbf{83.30} & \cellcolor[rgb]{0.9, 0.9, 0.9}\textbf{86.21} \\
        \hline
        \end{tabular}%
        \subcaption{Untargeted attack success rates (\%) for the adversarial examples crafted on DenseNet-121.}
    \end{subtable}%

    \label{tab:Attack_on_cifar10}
\end{table}%

\end{document}